\documentclass[journal, twoside]{IEEEtran}

\usepackage{xcolor}
\newcommand{\revised}[1]{\textcolor{black}{#1}}
\usepackage[utf8]{inputenc}   
\usepackage[T1]{fontenc}      
\usepackage{enumitem}

\usepackage{amsmath,amsfonts, amssymb}
\usepackage{siunitx}  
\usepackage{cuted}  

\usepackage{bm}
\usepackage{algorithm}
\usepackage{algpseudocode}
\algrenewcommand\algorithmicrequire{\textbf{Input:}}
\algrenewcommand\algorithmicensure{\textbf{Output:}}

\usepackage[caption=false]{subfig}
\usepackage{array}
\usepackage{textcomp}
\usepackage{stfloats}
\usepackage{url}
\usepackage{verbatim}
\usepackage[pdftex]{graphicx}
\graphicspath{./figs/}

\usepackage{booktabs}
\usepackage{tabularx}

\usepackage{cite}
\usepackage{balance}

\makeatletter
\let\NAT@parse\undefined
\makeatother

\usepackage[colorlinks,linkcolor=blue,anchorcolor=black,citecolor=blue,urlcolor=blue,hyperfootnotes=true]{hyperref}   
\usepackage[all]{hypcap} 

\begin{document}
\bstctlcite{IEEEtran:BSTcontrol}  

\title{Effector-Centric NMPC of Tiltable-Multirotors for Offset-Free Omnidirectional Aerial Manipulation}

\author{Jinjie Li, \textit{Graduate Student Member, IEEE}, Yicheng Chen, \textit{Graduate Student Member, IEEE}, Johannes Kübel, Haokun Liu, \textit{Graduate Student Member, IEEE}, Junichiro Sugihara, Moju Zhao, \textit{Member, IEEE}
\thanks{Received 25 September 2025; revised 21 March 2026 and 26 June 2026; accepted 1 August 2026. This work was supported in part by JSPS KAKENHI under Grant 23H03472 and in part by CSC. \textit{(Corresponding author: Moju Zhao.)}}
\thanks{All authors are with the DRAGON Lab, Department of Mechanical Engineering, The University of Tokyo, Tokyo 113-8656, Japan (e-mail: jinjie-li@dragon.t.u-tokyo.ac.jp; chou@dragon.t.u-tokyo.ac.jp).}
}

\markboth{IEEE~Transactions~on~Robotics,~Vol.~xx, No.~x, Aug.~2026}%
{Li~\MakeLowercase{\textit{et al.}}:~Effector-Centric~NMPC~of~Tiltable-Multirotors~for~Offset-Free~Omnidirectional~Aerial~Manipulation}


\IEEEaftertitletext{\vspace{-1.5\baselineskip}}  
\maketitle


\begin{abstract}

Aerial manipulation extends robotic operations to previously inaccessible aerial environments. Unlike arm-equipped aerial systems, tiltable-multirotors can directly generate six-degree-of-freedom wrenches through their flight bases, enabling both efficient movement and omnidirectional operation by tilting the thrust direction.

This work presents a design analysis and a wrench-based control framework for tiltable-multirotors in aerial manipulation. We show that a four-rotor tiltable configuration provides a balance between interference-free propeller sizing and hovering efficiency across different attitudes, and its null-space redundancy is crucial for traversing singular configurations under physical constraints. We further show that an upward end-effector placement yields a favorable trade-off between geometric clearance and available wrench. To address disturbances, we propose a dual strategy consisting of a modified integral term for model error and an acceleration-based estimator for external wrenches. Building on these insights, we develop an effector-centric nonlinear model predictive control (NMPC) framework that integrates design choices, singularity handling, and disturbance compensation into a unified formulation.


The proposed framework runs fully onboard at 100 Hz on a custom-built tiltable-quadrotor. Real-world experiments, including a 90$^\circ$ step cartwheel rotation, whiteboard pushing, and \revised{continuous 360$^\circ$} valve turning,
\revised{demonstrate the feasibility of wrench-based omnidirectional manipulation with singularity traversal on a one-DoF-per-arm tiltable-quadrotor.}

\end{abstract}




\section{Introduction}

\begin{figure}[t]
    \centerline{\includegraphics[trim=5mm 5mm 0 0,clip,width=3.5in]{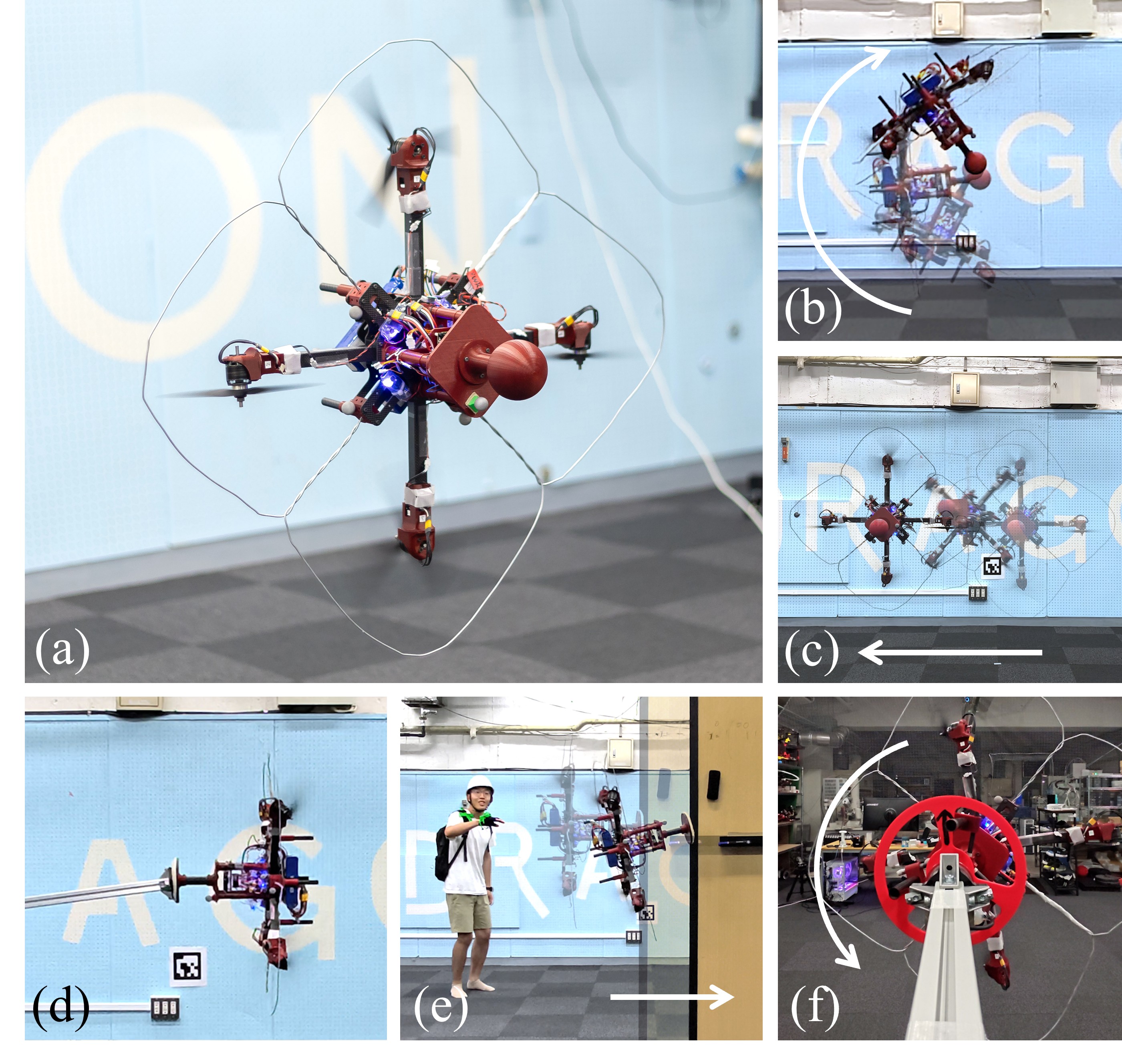}} 
    \vspace*{-1mm}
    \caption{\revised{Representative experiments included in this article}: (a)–(c) demonstrate effector-centric omnidirectional flight; (d) presents external wrench estimation and compensation, which enables the whiteboard-pushing task in (e); and (f) shows continuous turning of a vertically installed valve.}
    \label{fig:highlight}
\end{figure}

\IEEEPARstart{T}{he} focus of robotics has been shifting from locomotion to manipulation \cite{billard_trends_2019}. In line with this trend, aerial manipulation has attracted increasing attention in recent years. Traditional manipulation originated in industrial applications, where serial or parallel structures \cite{tsai_robot_1999} are used to provide six-degree-of-freedom (6-DoF) motion and the required wrench, with task-specific end-effectors mounted for different operations. Inspired by this paradigm, many aerial platforms have also been equipped with robotic manipulators \cite{ollero_past_2022, zhong_prototype_2024}, including serial arms \cite{nava_direct_2020, he_flying_2025, li_fixed-time_2025, lee_autonomous_2025} and parallel arms \cite{tzoumanikas_aerial_2020, bodie_dynamic_2021, cuniato_design_2023}. However, in theory, the motion and wrench required at the end-effector can be generated directly by the flight base itself, \revised{potentially reducing} structural and control complexity. Ryll et al. \cite{ryll_6d_2019} termed this paradigm the \textit{flying end-effector}, which consists of 1) an aerial base capable of generating a stable 6-DoF wrench, 2) an estimator for external wrenches, and 3) a controller to \revised{support} stable flight across tasks.
In this work, we adopt this paradigm and achieve omnidirectional manipulation on a tiltable-quadrotor using a nonlinear model predictive control (NMPC) framework, as shown in Fig.~\ref{fig:highlight}.

\setlength{\textfloatsep}{8pt plus 1.0pt minus 2.0pt}

\IEEEpubidadjcol  

Although Ryll’s work successfully demonstrated sliding-surface and peg-in-hole tasks, a key limitation of their system was the lack of \textit{omnidirectional manipulation}.
Adopting a hexrotor with fixed tilting angles, they achieved full actuation but generated a non-uniform wrench distribution across $SO(3)$, restricting manipulation primarily to the robot's underside.
Note that the concept of \textit{omnidirectional manipulation} here is stricter than \textit{omnidirectionality} as defined in \cite{hamandi_design_2021}. Specifically, \revised{this definition requires} the zero-wrench point to lie inside the admissible wrench space, and the end-effector mounted on the flying base should also be orientable in \revised{arbitrary directions} for wrench application, which requires the generated force to exceed gravity in the corresponding orientations. Under this definition, \revised{relatively few reported systems} have achieved omnidirectional manipulation.

\revised{An early omnidirectional aerial manipulator is} the octorotor proposed by Brescianini et al. \cite{brescianini_design_2016}, which employed eight bidirectional propellers fixed at different angles. A similar concept is ODAR, developed by Park et al. \cite{park_odar_2018}. Yiğit et al. later introduced a macro-mini aerial manipulator \cite{yigit_dynamic_2023}, in which an omnidirectional flying platform is tethered with winches to save energy. However, such \textit{fixed-rotor} designs inherently suffer from counteracting internal forces.
To address these limitations, researchers explored tilting structures, which allow the thrust directions to be reoriented and thereby reduce energy consumption.
One direction is the \textit{multirotor} style, in which each arm carries a tiltable rotor with one or two DoFs \cite{yang_new_2024}. A representative example is Voliro by Kamel et al. \cite{kamel_voliro_2018, allenspach_design_2020, bodie_active_2021}, which features six symmetrically arranged rotors that tilt radially.
Another direction is the \textit{multilink} style, in which each link carries a tiltable rotor module to form articulated aerial robots. A representative example is DRAGON by Zhao et al. \cite{zhao_design_2018, zhao_versatile_2022}, which consists of four links, each equipped with a 2-DoF rotor module.
The \textit{multirotor} style \revised{generally} has lower mechanical complexity than the \textit{multilink} style while retaining the capability of omnidirectional manipulation, and is therefore the primary domain of this work. 
Focusing on robots with one single-axis tiltable rotor per arm, we are curious whether \revised{more compact} platforms, such as tiltable-trirotors or tiltable-quadrotors, are sufficient for omnidirectional operation. This raises the following question:

\begin{enumerate}
\item \textit{How does the rotor number influence the design trade-offs and omnidirectional performance of tiltable-multirotors?}
\end{enumerate}

The review by Hamandi et al. \cite{hamandi_design_2021} provides a valuable overview of different tiltable-rotor platforms, but focuses primarily on taxonomy rather than concrete design trade-offs or flight performance. In this work, we quantitatively analyze how rotor number affects the \revised{available} propeller size that avoids physical and aerodynamic interference between adjacent rotors, showing that fewer rotors offer \revised{a larger design margin} for propeller sizing. We further evaluate hovering efficiency across different attitudes, showing that more rotors lead to a more even thrust distribution. Balancing these factors, the \revised{symmetric} tiltable-quadrotor emerges as a \revised{practical compromise for the analyzed one-DoF-per-arm tiltable-multirotors}.

Omnidirectional capability comes from full- or over-actuation, which requires at least six independent actuators \cite{ollero_past_2022}, as in tiltable-trirotors.
However, in practice, omnidirectional flight requires traversing singular poses, in which an arm is directed upwards while hovering. This process is further constrained by motor-cable routing, limiting the tilting range to one side. Our analysis shows that the fully actuated tilt-trirotor \revised{struggles} to satisfy these practical requirements, although it is theoretically feasible. By contrast, the tiltable-quadrotor and platforms with more rotors are overactuated, allowing null-space redundancy to be used to bypass singularities within motor-cable limits.
While Bodie et al. \cite{bodie_towards_2020} achieved singularity-included flight on a tiltable-hexrotor, \revised{reported tiltable-quadrotor experiments have not yet shown the same capability}, possibly because fewer control inputs generally imply greater control difficulty. 
In this work, we achieve not only $360^\circ$ roll and pitch rotations but also flight through singular configurations using a tiltable-quadrotor, \revised{extending experimental omnidirectional manipulation to a four-rotor platform within the comparison in Table~\ref{tab:comparison_related_works}}.


\begin{table}[t]
\centering
\caption{Comparison with representative works on tiltable-quadrotors}
\label{tab:comparison_related_works}
\resizebox{\columnwidth}{!}{
\begin{tabular}{lccccc}
\toprule
\textbf{Ref.} & \textbf{Control} & \shortstack{\textbf{Servo}\\\textbf{Dyn.}} & \shortstack{\textbf{Sing.}\\\textbf{Handling}} & \shortstack{\textbf{Wrench}\\\textbf{Est.}} & \shortstack{\textbf{EE-}\\\textbf{Centric}} \\
\midrule
\cite{ryll_novel_2015} & Feedback Linearization & No & No & No & No \\
\cite{oosedo_flight_2015} & Switching PID & No & No & No & No \\
\cite{invernizzi_full_2018} & IQTO & No & No & No & No \\
\cite{lee_autonomous_2024} & Switching PID & No & No & Yes & No \\
\cite{li_servo_2024} & NMPC & Yes & No & No & No \\
\cite{mellet_design_2025} & Impedance & No & No & Yes & No \\
\cite{guan_thrust-coefficient-coupled_2026} & PID + Online Param. Id. & No & No & Yes & No \\
\textbf{Ours} & \textbf{NMPC} & \textbf{Yes} & \textbf{Yes} & \textbf{Yes} & \textbf{Yes} \\
\bottomrule
\end{tabular}
}
\end{table}

\begin{enumerate}
\setcounter{enumi}{1}
\item \textit{How does the mounting angle of a rigid end-effector affect geometric properties and wrench capability?}
\end{enumerate}

Once omnidirectional flight is achieved, the next question is how the end-effector should be mounted on the airframe for different tasks.
\revised{This design choice has received limited systematic attention.} Existing platforms \revised{provide} partial insights: Ryll et al. \cite{ryll_6d_2019} placed the rigid end-effector downward to exploit the underside workspace, most overactuated aerial robots mounted it laterally for horizontal tasks \cite{gonzalez-morgado_controlled_2025, mellet_design_2025, bodie_active_2021}, and Hui et al. \cite{hui_passive_2024} mounted a contact-based sensor on top of the robot to push against the ceiling.
In this work, we analyze this problem from three perspectives: (i) geometric properties such as \revised{required operation distance} and \revised{allowable rod radius}, (ii) the \revised{available wrench envelope} in the body frame, and (iii) the \revised{available wrench envelope} in the world frame, including a representative task requiring coupled wrench generation.
Our results show that vertical mounting yields a larger wrench envelope than horizontal mounting under the considered design criteria, making it \revised{well suited} for tasks requiring large wrench output. The wrench envelope in the world frame also provides a practical reference for task-oriented deployment.



\begin{enumerate}
\setcounter{enumi}{2}
\item \textit{How can model error and external wrench be handled separately in a common NMPC framework?}
\end{enumerate}

Another essential topic is disturbance handling, where disturbances generally consist of \textit{model error} and \textit{external wrench}.
These two components have different origins: the former arises from the robot itself, such as manufacturing errors, whereas the latter stems from the environment.

Most trajectory-tracking works on tiltable-multirotors treat them as a single disturbance term and compensate for it using observers \cite{mckinnon_estimating_2020,wang_neural_2023,yang_new_2024} or integral action \cite{allenspach_design_2020, invernizzi_comparison_2021, li_servo_2024}. However, in aerial manipulation, independent estimation of the external wrench is required for environmental interaction \cite{ryll_6d_2019,bodie_active_2021}.
In this work, we explicitly separate the two components: \textit{model error} is compensated using a modified integral term, and \textit{external wrench} is estimated via an acceleration-based method.
Unlike the integral terms \cite{allenspach_design_2020, invernizzi_comparison_2021, li_servo_2024} that rely on reference signals, 
the proposed integral term is based on the NMPC predictions. This design avoids unnecessary accumulation when the reference is far from the current state and \revised{helps reduce overshoot in the tracking response}.
To incorporate these methods into NMPC, \textit{model errors} are treated as constant parameters and \textit{external wrenches} are treated as additional states. \revised{Our main contribution is this role-separated integration within the NMPC framework, together with a frequency-response analysis of the dual strategy.}

\begin{enumerate}
\setcounter{enumi}{3}
\item \textit{How can NMPC be formulated to track effector-centric trajectories and unify the above components?}
\end{enumerate}

Previous NMPC studies on tiltable-multirotors typically use the center of gravity (CoG) as the reference frame \cite{brunner_trajectory_2020,bicego_nonlinear_2020,li_servo_2024}.
Although effective for flight, this formulation requires additional coordinate transformations in the planner to generate trajectories applicable to end-effectors (EEs).
In this work, we instead embed this transformation into the controller, mainly to simplify the construction of task trajectories.
A related idea was proposed by He et al. \cite{he_flying_2025} for aerial platforms equipped with a robotic arm in the form of an EE-centric whole-body MPC controller. Following this idea, we develop an effector-centric NMPC framework for tiltable-multirotors with a rigid end-effector. Specifically, the cost function is defined in the EE frame for task execution, while the internal dynamics are maintained in the CoG frame for computational simplicity.
Moreover, NMPC serves as the unifying core of the overall framework: 
null-space redundancy is exploited to generate singularity-free references; the mounting-angle analysis informs the pose of the EE; \textit{model errors} are incorporated as constant parameters; and external wrenches are introduced as additional states with their own dynamics.

\setlength{\textfloatsep}{8pt plus 1.0pt minus 2.0pt}
\begin{figure}[t]
    \centerline{\includegraphics[trim=8cm 5.0cm 7.6cm 4.2cm,clip,width=3.5in]{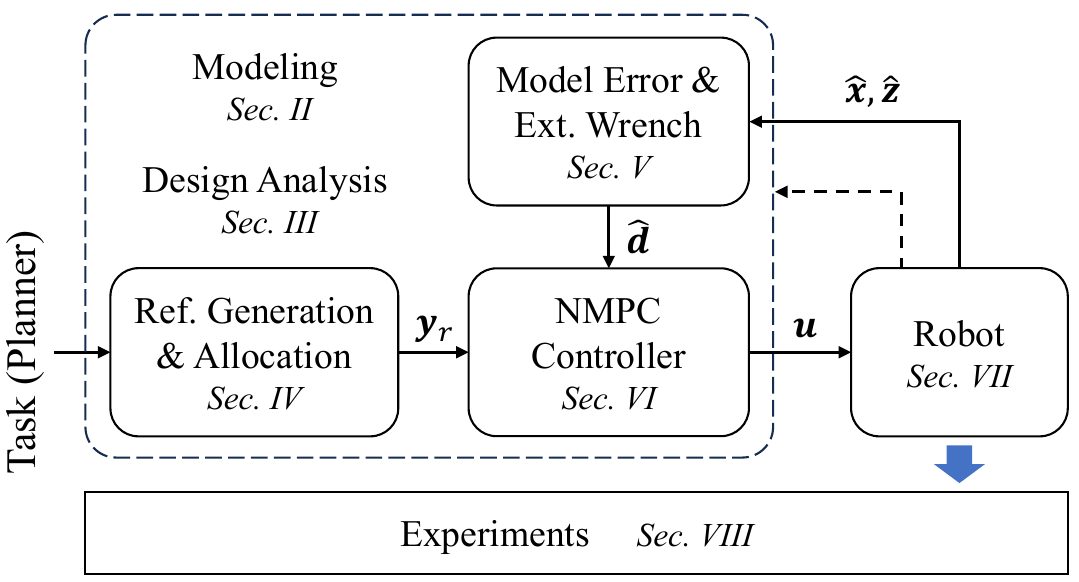}} 
    \vspace*{-1mm}
    \caption{
    Simplified block diagram of the proposed effector-centric NMPC framework.
    \revised{The task planner generates predefined trajectories or teleoperation pose commands.}
    Secs.~\ref{sec:modeling} and \ref{sec:design} provide the basis for the main algorithms in Secs. \ref{sec:ref}--\ref{sec:control}, whose results are implemented in Sec.~\ref{sec:robot}. \revised{Note that the allocation is performed before the controller to provide the reference.}
    The entire framework is validated in Sec.~\ref{sec:exp}.}
    \label{fig:overall_structure}
\end{figure}

To summarize, the main contributions of this article are:

\begin{enumerate}
\item A quantitative design analysis of rotor number and end-effector mounting angle for tiltable-multirotors, showing that the tiltable-quadrotor provides a \revised{practical compromise under the analyzed metrics} and that vertical mounting is \revised{well suited} for tasks requiring large wrench.

\item A method to handle singularities for tiltable-quadrotors that exploits null-space redundancy, demonstrating the importance of overactuation for singularity traversal under structural limits.
\item \revised{An NMPC-integrated disturbance-handling strategy with separate roles for model-error compensation and external-wrench estimation,} enabling both offset-free tracking and environmental interaction.
\item An effector-centric NMPC framework that tracks an end-effector reference rather than a CoG reference and integrates the proposed design and control components into a unified architecture.

\item Extensive real-world experiments that validate the overall framework, including $\pm180^\circ$ roll and pitch EE-centric rotations, singularity-included cartwheel rotations, open-loop and closed-loop wrench estimation, whiteboard pushing, and continuous \revised{$360^\circ$} turning of a vertical valve.

\end{enumerate}

\revised{The main value of the proposed framework lies in its integrated NMPC control architecture and experimental validation, rather than in formal stability or convergence analysis.}
To the best of our knowledge, \revised{this is the first experimental demonstration of wrench-based omnidirectional aerial manipulation with singularity traversal on a one-DoF-per-arm tiltable-quadrotor.}






The block diagram of the proposed framework, together with its correspondence to the sections of this article, is shown in Fig.~\ref{fig:overall_structure}. Section~\ref{sec:modeling} introduces the mathematical model of tiltable-multirotors, followed by the design analysis in Section~\ref{sec:design}.
Based on this model, Section~\ref{sec:ref} presents the reference generation method, and Section~\ref{sec:error} details the handling of model error and external wrench. These components are then integrated into the effector-centric NMPC framework presented in Section~\ref{sec:control}. Section~\ref{sec:robot} describes the robot platform, and Section~\ref{sec:exp} reports the experimental results. Finally, Section~\ref{sec:conclusion} concludes the article.


\section{Modeling} \label{sec:modeling}

\subsection{Notation and Coordinate Systems}

We denote scalars by nonbold symbols $x, X \in \mathbb{R}$, vectors by bold lowercase $\boldsymbol{x} \in \mathbb{R}^n$, and matrices by bold uppercase $\boldsymbol{X} \in \mathbb{R}^{n \times m}$.
We use $(\cdot)_r$ for reference, $(\cdot)_c$ for commands, and
$\hat{\cdot}$ for estimated values. A vector in $\left\{\mathcal{W}\right\}$ is denoted by $^{W}\boldsymbol{p}$, and the rotation from $\left\{\mathcal{B}\right\}$ to $\left\{\mathcal{W}\right\}$ is denoted by $^{W}_{B}\boldsymbol{R}$ (rotation matrix) or $^{W}_{B}\boldsymbol{q}=\left[q_w, q_x, q_y, q_z\right]^\top$ (attitude quaternion). The rotation matrix converted from a quaternion $\boldsymbol{q}$ is denoted by $^{W}_{B}\boldsymbol{R}(\boldsymbol{q})$. For quaternions, we use $\mathcal{V}(\cdot)$ for their vector part $\mathcal{V}(\boldsymbol{q}) := \left[q_x, q_y, q_z\right]^\top$, $\circ$ for multiplication, and $(\cdot)^*$ for the conjugation. Note that $\boldsymbol{q}^*=\boldsymbol{q}^{-1}$ for a unit quaternion.

\begin{figure}[t]
    \centerline{\includegraphics[trim=7.5cm 5cm 7.5cm 5cm,clip,width=3.4in]{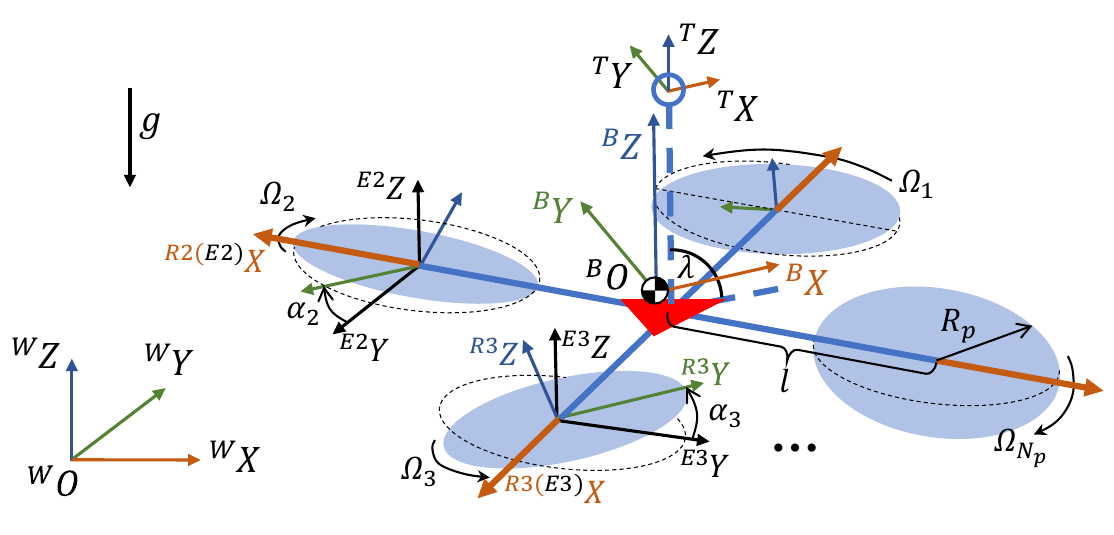}} 
    \vspace*{-3mm}
    \caption{Diagram of a tiltable-multirotor with the ENU  world frame $\left\{\mathcal{W}\right\}$ (X East, Y North, Z Up), the FLU body frame $\left\{\mathcal{B}\right\}$ (X Forward, Y Left, Z Up), and the tool frame $\left\{\mathcal{T}\right\}$ for end-effectors. The $i$th rotor $\left\{\mathcal{R}_i\right\}$ rotates around the X-axis (red) of $\left\{\mathcal{E}_i\right\}$.}
    \label{fig:coordinate}
\end{figure}

As depicted in Fig.~\ref{fig:coordinate}, for a robot with $N_p$ rotors, the coordinate systems include the world frame $\left\{\mathcal{W}\right\}$, the body frame $\left\{\mathcal{B}\right\}$ whose origin is at the center of gravity (CoG), the $i$th arm-end frame $\left\{\mathcal{E}_i\right\}$ and rotor frame $\left\{\mathcal{R}_i\right\}$ $(i=1,..., N_p)$, as well as the tool frame $\left\{\mathcal{T}\right\}$ for manipulation.
The origin of $\left\{\mathcal{E}_i\right\}$ is located at the tip of the $i$th arm, with its X-axis pointing outward and its Z-axis pointing upward. The frame $\left\{\mathcal{R}_i\right\}$ is obtained from $\left\{\mathcal{E}_i\right\}$ by a rotation about the X-axis, with rotation angle denoted by $\alpha_i$. The relative pose of $\left\{\mathcal{T}\right\}$ w.r.t. $\left\{\mathcal{B}\right\}$ is defined in Sec.~\ref{sec:design}.


\setlength{\textfloatsep}{8pt plus 1.0pt minus 2.0pt}

\subsection{Model for a Tiltable-Multirotor}

The model of a tiltable-multirotor consists of five components: rotor model, servo model, actuator wrench model, external wrench model, and rigid-body model.

\subsubsection{Rotor Model}

The rotor model describes the relationship between the rotation speed $\Omega_i$ and the resulting thrust $f_i$ and reaction torque $\tau_i$, often expressed in the following quadratic form:
\begin{equation}
\label{eq:rotor_dynamics}
    f_i={k}_{t} \, \Omega_i^2, \quad \tau_i = - b_{i} \, {k}_{q} \, \Omega_i^2,
\end{equation}
where ${k}_{t}$ and ${k}_{q}$ are coefficients identified from motor tests, and $b_{i}\in\{-1,1\}$ indicates the rotation direction. Motors require time to reach the desired speed. However, this transient process is less critical than that of servos and can be neglected \cite{li_servo_2024}. We find in practice that this neglect also reduces attitude error. The thrust command is denoted as $\boldsymbol{f}_c=\left[f_{1},f_{2},\cdots,f_{N_p}\right]^\top$.


\subsubsection{Servo Model}

Similar to rotors, servos also require time to reach the desired angle. Considering most low-cost servos only support angle control, we model the transient process using the following first-order system:
\begin{equation}
    \label{eq:servo_model}
    \dot{\boldsymbol{\alpha}} = \frac{1}{t_{\rm servo}} \left( \boldsymbol{\alpha}_c - \boldsymbol{\alpha} \right), \enspace {\rm with} \ \alpha_i \in \left[\alpha_{i,\min}, \alpha_{i,\max}\right]
\end{equation}
where $\boldsymbol{\alpha} = \left[\alpha_1, \alpha_2, \ldots, \alpha_{N_p} \right]^\top$ contains the states of all servo angles, $\boldsymbol{\alpha}_c = \left[\alpha_{1c}, \alpha_{2c}, \ldots, \alpha_{N_pc} \right]^\top$ denotes the servo angle commands, and $t_{\rm servo}$ is the identified time constant.

\subsubsection{Actuator Wrench}

To evaluate the effect of actuators on the entire robot, their influence must be expressed in the body frame $\left\{\mathcal{B}\right\}$. The rotor wrench is first expressed in its rotor frame $\left\{\mathcal{R}_i\right\}$ as
\begin{equation} \label{eq:input_wrench}
        {^{R_i}\boldsymbol{f}_i} = \left[0, 0, f_i \right]^\top, \quad {^{R_i}\boldsymbol{\tau}_i} = \left[0, 0, \tau_i \right]^\top.
\end{equation}
Based on the definition of tiltable-multirotors, the tilt angle $\alpha_i$ determines the rotation from $\left\{\mathcal{R}_i\right\}$ to $\left\{\mathcal{E}_i\right\}$ about the X-axis:
\begin{equation}
    {^{E_i}_{R_i}\boldsymbol{R}} = \boldsymbol{R}_X\left(\alpha_i\right).
\end{equation}
The gyroscopic effect of tilting rotating propellers is considered negligible \cite{bicego_nonlinear_2020} and is therefore neglected. The net force and torque generated by the propellers are then derived as
\begin{subequations}
    \vspace{-3mm}
\label{eq:resultant_wrench}
\begin{align}
    {^B\boldsymbol{f}_u} = &\sum_{i=1}^{N_p}{^{B}_{R_i}\boldsymbol{R}} \ {^{R_i}\boldsymbol{f}_i}, \ \text{where} \ {^{B}_{R_i}\boldsymbol{R}} = {^{B}_{E_i}\boldsymbol{R}} \ {^{E_i}_{R_i}\boldsymbol{R}} \label{eq:fu} \\
    {^B\boldsymbol{\tau}_u} = &\sum_{i=1}^{N_p}\left({^{B}_{R_i}\boldsymbol{R}} \ {^{R_i}\boldsymbol{\tau}_i} + ^{B}\boldsymbol{p}_{E_i} \times \left({^{B}_{R_i}\boldsymbol{R}} \ {^{R_i}\boldsymbol{f}_i} \right) \right), \label{eq:tau_u}
\end{align} \label{eq:fu_and_tauu}
\end{subequations}
where the position $^{B}\boldsymbol{p}_{E_i}$ and attitude ${^{B}_{E_i}\boldsymbol{R}}$ from $\left\{\mathcal{E}_i\right\}$ to $\left\{\mathcal{B}\right\}$ can be obtained from geometric properties.

\subsubsection{External Wrench}

The external wrench may arise from various sources, including wind gusts, collisions, and manipulation.
\revised{With a rigid end-effector mounted on the robot, we represent the combined environmental effects as an equivalent resultant wrench at the tool frame $\left\{\mathcal{T}\right\}$, assuming that end-effector interaction is dominant.} This wrench is denoted by ${^T\boldsymbol{f}_{de}}, {^T\boldsymbol{\tau}_{de}}$, and the corresponding forms in other frames are
\begin{subequations} \label{eq:effector_wrench}
    \begin{align}
        {^{W}\boldsymbol{f}_{de}} &= {^{W}_{B}\boldsymbol{R}} \ {^{B}_{T}\boldsymbol{R}} \ {^T\boldsymbol{f}_{de}}, \\
        {^{B}\boldsymbol{\tau}_{de}} &= {^{B}_{T}\boldsymbol{R}} \ {^T\boldsymbol{\tau}_{de}} + {^{B}\boldsymbol{p}_{T_o}} \times {^{B}_{T}\boldsymbol{R}} \ {^T\boldsymbol{f}_{de}},
    \end{align}
\end{subequations}
where ${^{B}\boldsymbol{p}_{T_o}}, {^{B}_{T}\boldsymbol{R}}$ are from geometric properties.

The external wrench is assumed constant over each NMPC prediction horizon and modeled as
\begin{equation}  \label{eq:ext_wrench_mdl}
    {^W\dot{\boldsymbol{f}}_{de}} = \boldsymbol{0}, \quad {^B\dot{\boldsymbol{\tau}}_{de}} = \boldsymbol{0}.
\end{equation}

\subsubsection{Rigid-Body Model}

The motion of the tiltable-multirotor is affected by actuator wrenches, gravity, and disturbances. The disturbances consist of model error and external wrench. With position ${^W\boldsymbol{p}}$, velocity $^W\boldsymbol{v}$, quaternion ${^W_B\boldsymbol{q}}$, and angular velocity ${^B \boldsymbol{\omega}}$ (the angular velocity of $\left\{\mathcal{B}\right\}$ with respect to $\left\{\mathcal{W}\right\}$, expressed in $\left\{\mathcal{B}\right\}$) as states, a standard six-DoF rigid-body dynamic model can be formulated \cite{li_nonlinear_2023} as
\begin{subequations}
\label{eq:rigid_body}
\begin{align}
{^W\dot{\boldsymbol{p}}} &= {^W\boldsymbol{v}}, \label{eq:rigid_body_1} \\[5pt]
{^W\dot{\boldsymbol{v}}} &= \left({^{W}_{B}\boldsymbol{R}(\boldsymbol{q})} \ {^B\boldsymbol{f}_u} + {^W\boldsymbol{f}_{de}} + {^W\boldsymbol{f}_{dm}}\right) / {m} + {^W\boldsymbol{{g}}}, \label{eq:rigid_body_2}\\[5pt]
{^W_B\dot{\boldsymbol{q}}} &= \frac{1}{2} \ {^W_B\boldsymbol{q}} \circ \mathcal{H}(^B\boldsymbol{\omega}), \label{eq:rigid_body_3}\\[5pt]
{^B\dot{\boldsymbol{\omega}}} &=\boldsymbol{{I}}^{-1}  \left(-{^B \boldsymbol{\omega}} \times\left({\boldsymbol{{I}}}  {^B \boldsymbol{\omega}}\right) + {^B\boldsymbol{\tau}_u} + {^B\boldsymbol{\tau}_{de}} + {^B\boldsymbol{\tau}_{dm}}\right), \label{eq:rigid_body_4}
\end{align}
\end{subequations}
where $\circ$ denotes quaternion multiplication; $\mathcal{H}(\cdot)$ denotes homogenization of a 3D vector, $\mathcal{H}(\boldsymbol{p}) := [0, \boldsymbol{p}]^\top$; ${^B\boldsymbol{f}_u}$ and ${^B\boldsymbol{\tau}_u}$ are given in (\ref{eq:resultant_wrench}); ${^W\boldsymbol{f}_{de}}$ and ${^B\boldsymbol{\tau}_{de}}$ are the external wrench defined in (\ref{eq:effector_wrench}); ${^W\boldsymbol{f}_{dm}}$ and ${^B\boldsymbol{\tau}_{dm}}$ represent the wrench due to model error;
$m$, ${^W\boldsymbol{{g}}}=[0,0,{-g}]^\top$, and $\boldsymbol{I}=\texttt{diag} ({I}_{xx}, {I}_{yy}, {I}_{zz})$ denote the mass, gravity vector, and inertia matrix, respectively.

The model established above is used in the subsequent sections on design analysis, reference generation, wrench estimation, and control.


\section{Design Analysis} \label{sec:design}

Although the previous section established the robot model, the rotor number $N_p$ and the end-effector mounting angle $\lambda$ remain to be determined. In this section, we first analyze how the rotor number affects inter-rotor interference and hovering efficiency, and show that a quadrotor configuration provides a \revised{practical compromise under the analyzed design metrics}. Subsequently, we analyze how the mounting angle affects geometric properties and wrench capability, and show that vertical mounting is \revised{well suited} for tasks requiring large wrench output.

\subsection{Analysis of Rotor Number in Tiltable-Multirotors} \label{sec:design_rotor_num}

\subsubsection{Interference}

\setlength{\textfloatsep}{8pt plus 1.0pt minus 2.0pt}
\begin{figure}[t]
    \centering
    \subfloat[Illustration of rotor interference in tiltable multirotors]{
    \includegraphics[trim=7.5cm 6cm 7.5cm 6cm,clip,width=3.4in]{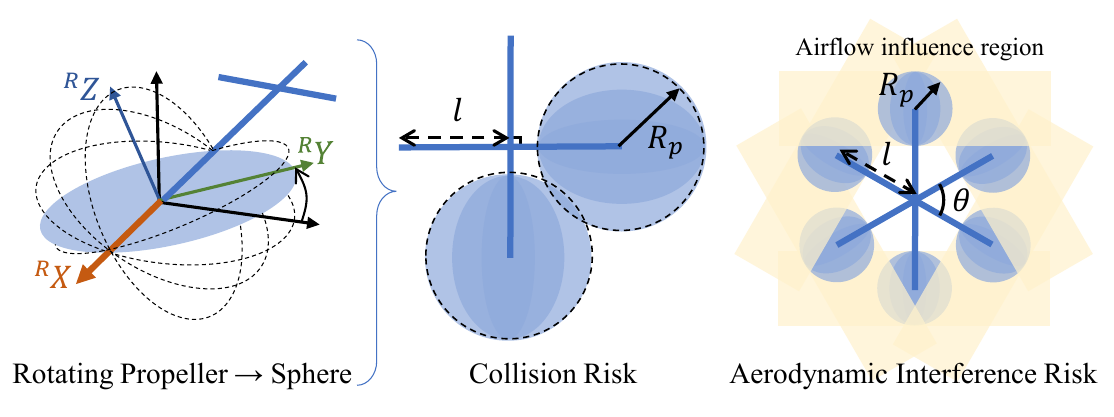}
        \label{fig:diff_aero_interference_illus}
    } \\ \vspace{-2mm}
    \subfloat[Feasible propeller size for tiltable-multirotors with different rotor numbers]{
        \includegraphics[trim=0 0 0 0,clip,width=3.4in]{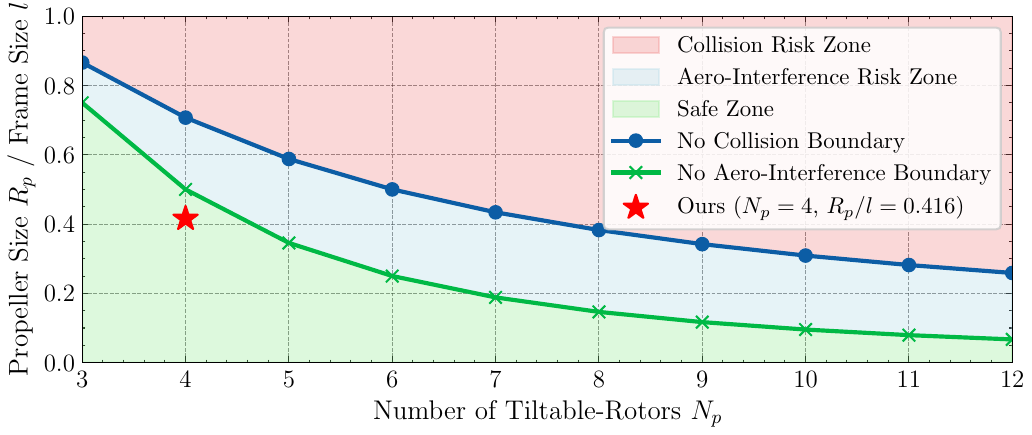}
        \label{fig:prop_size_select}
    }
    \caption{Rotor interference and feasible propeller sizes of tiltable-multirotors. The central region of the platform remains unaffected by rotor airflow for any tilting angle, making it a potentially suitable location for mounting an end-effector. Fewer rotors allow a larger feasible propeller size.}
    \label{fig:diff_aero_interference}
\end{figure}

As illustrated in Fig.~\ref{fig:diff_aero_interference_illus}, the tilting mechanism of tiltable-multirotors causes each propeller to sweep a spherical volume, which introduces the risk of physical collisions between adjacent rotors. Furthermore, unlike the fixed-direction rotors of conventional multirotors, tiltable rotors can change their airflow direction during flight, leading to potential aerodynamic interference among rotors.
To avoid these two types of interference, we analyze the corresponding propeller-size boundaries and use them as a design reference. 

Assuming that the rotors are evenly distributed, the condition for avoiding inter-rotor collision can be written as
\begin{equation}
    R_p / l <  \sin(\theta/2), \ \ \theta=2\pi/N_p,
\end{equation}
where $R_p$ and $l$ are defined in Fig.~\ref{fig:diff_aero_interference_illus}. 
Similarly, the condition for avoiding aerodynamic interference is given by
\begin{equation}
    R_p / l <  \left(1-\cos{\theta}\right) / 2.
\end{equation}
These geometric relations are sketched in Fig.~\ref{fig:diff_aero_interference_illus}.

The quantitative results of these design boundaries are shown in Fig.~\ref{fig:prop_size_select}.
In general, larger feasible propeller sizes are associated with higher flight efficiency.
However, Fig.~\ref{fig:prop_size_select} shows that increasing the number of tiltable rotors inevitably reduces the collision-free propeller size, thereby limiting flight efficiency. Notably, the aerodynamic-interference boundary decreases even more steeply, indicating far fewer propeller options if aerodynamic interference among tilted rotors is to be avoided. From this perspective, using fewer tiltable rotors makes it easier to design a more efficient tiltable-multirotor.

\subsubsection{Hovering Efficiency}

\begin{figure}[t]
\centerline{\includegraphics[trim=0cm 0cm 0cm 0cm,clip,width=3.4in]{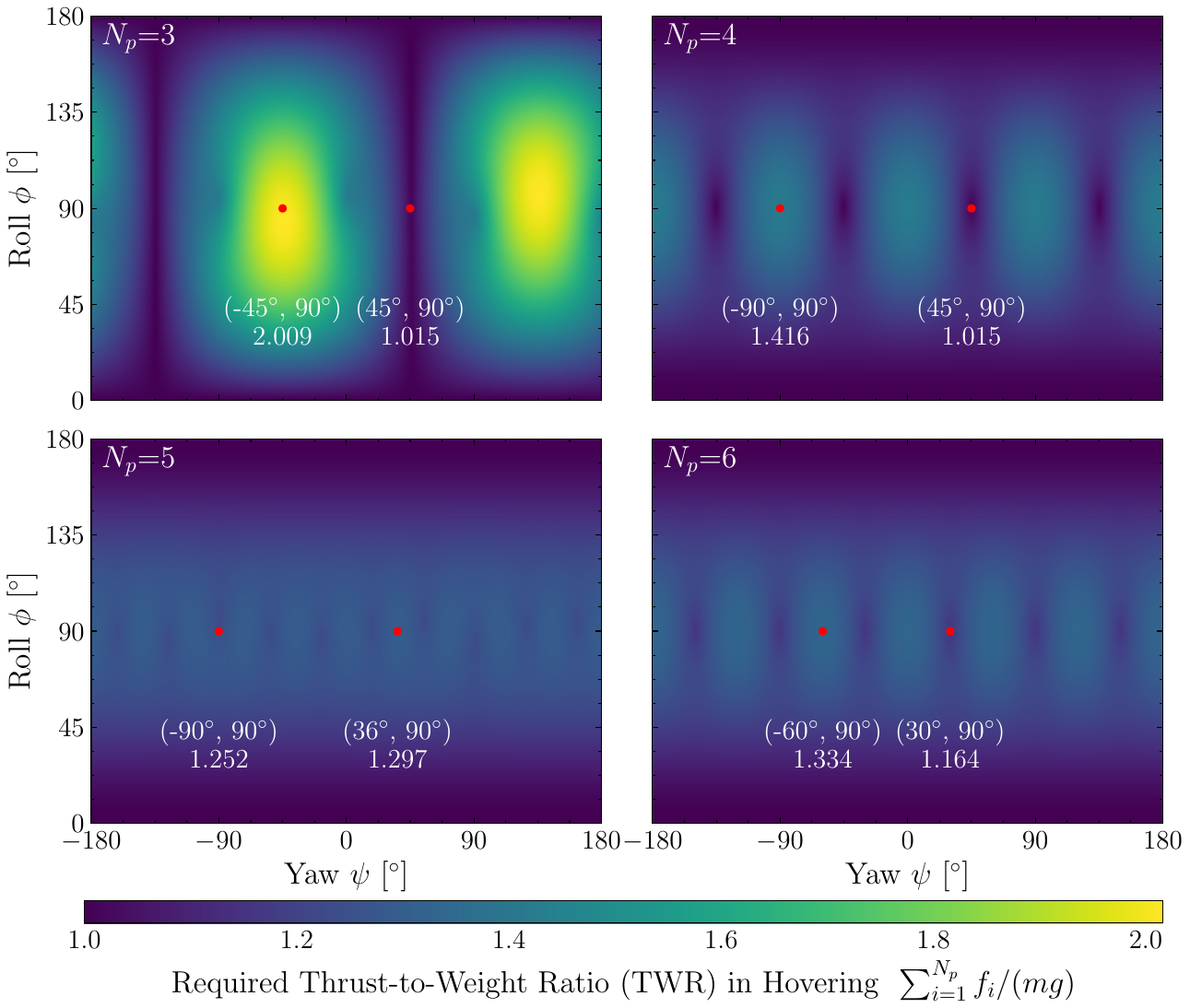}} 
    \vspace*{-3mm}
    \caption{Hovering TWR over different attitudes with rotor numbers from 3 to 6, where a higher value indicates lower efficiency. The trirotor adopts a T-shaped layout, whereas the other configurations distribute the rotors evenly. Red dots mark representative attitudes. As the rotor number increases, the required total thrust becomes more uniformly distributed over the attitude space.}
    \label{fig:hover_efficiency}
\end{figure}

Rotor number also affects the hovering thrust distribution across different attitudes. To quantitatively evaluate this effect, Fig.~\ref{fig:hover_efficiency} plots the required hovering thrust for tiltable-multirotors with rotor numbers ranging from 3 to 6. Because the hovering-thrust distribution is symmetric, plotting two attitude axes is sufficient to represent the entire attitude space.
For each attitude, represented by extrinsic Euler angles in the `zyx' order, the required servo angles and rotor thrusts are computed using the pseudo-inverse allocation method detailed in Sec.~\ref{sec:ref}. The rotor thrusts are then summed, and the total thrust-to-weight ratio (TWR) is used as a normalized metric for comparison. Obtained by removing one rotor from a tiltable-quadrotor, the trirotor adopts a T-layout to reflect existing work \cite{papachristos_dualauthority_2016} and the commercial product Voliro\footnote{More detailed information is available under \url{https://voliro.com/}}. The other configurations distribute the rotors evenly.

Fig.~\ref{fig:hover_efficiency} shows that a larger rotor number generally leads to less variation in the required hovering thrust across attitudes. Interestingly, the T-shaped trirotor maintains a continuously low thrust requirement over all roll angles when the yaw angle is $\psi=45^\circ\pm180^\circ$, making it suitable for force-only applications. However, it also exhibits the highest TWR in certain attitudes, such as $[\phi=90^\circ,\psi=90^\circ]$, which makes it less efficient for omnidirectional manipulation tasks. Another notable feature is the periodicity of the distribution: when $\phi=90^\circ$, configurations with an even rotor number exhibit $N_p$ waves, whereas the tiltable-pentarotor shows $2 N_p$ cycles with relatively high TWR even at the troughs. The symmetry of even rotor-number configurations leads to more regular thrust variation. In addition, this figure can be used to estimate the flight time at any attitude from the flight time in horizontal hovering, or to determine the maximum rotor thrust required for omnidirectional flight.

\setlength{\textfloatsep}{8pt plus 1.0pt minus 2.0pt}

\subsubsection{Rotor-Number Selection}

From the interference perspective, platforms with fewer rotors provide a larger design margin for propeller sizing. In contrast, increasing the rotor number improves the uniformity of the hovering thrust distribution across attitudes. Moreover, the analysis in Sec.~\ref{sec:ref} shows that the fully actuated tiltable-trirotor suffers from singularity-induced discontinuity, whereas overactuated platforms ($N_p \geq 4$) can overcome this issue. Therefore, \revised{for the analyzed one-DoF-per-arm tiltable-multirotors,} the tiltable-quadrotor appears to be a balanced compromise between design freedom and thrust uniformity for omnidirectional manipulation. Accordingly, the subsequent mounting-angle analysis focuses on the tiltable-quadrotor.

\subsection{Analysis of End-Effector Mounting Angle}

\subsubsection{Geometric Properties of Operation}

In addition to the number of rotors, another important design variable is the mounting angle of the end-effector. To avoid collisions with the tilting propellers, the end-effector is mounted at the center of the platform, making the mounting angle a primary design parameter. In this subsection, we analyze how this angle affects the \revised{required operation distance} and the feasible rod radius of the end-effector.

\setlength{\textfloatsep}{8pt plus 1.0pt minus 2.0pt}
\begin{figure}[t]
    \centering
    \subfloat[Illustration of the geometric relations used in the following analysis]{
    \includegraphics[trim=4.5cm 5cm 4.5cm 5.5cm,clip,width=3.4in]{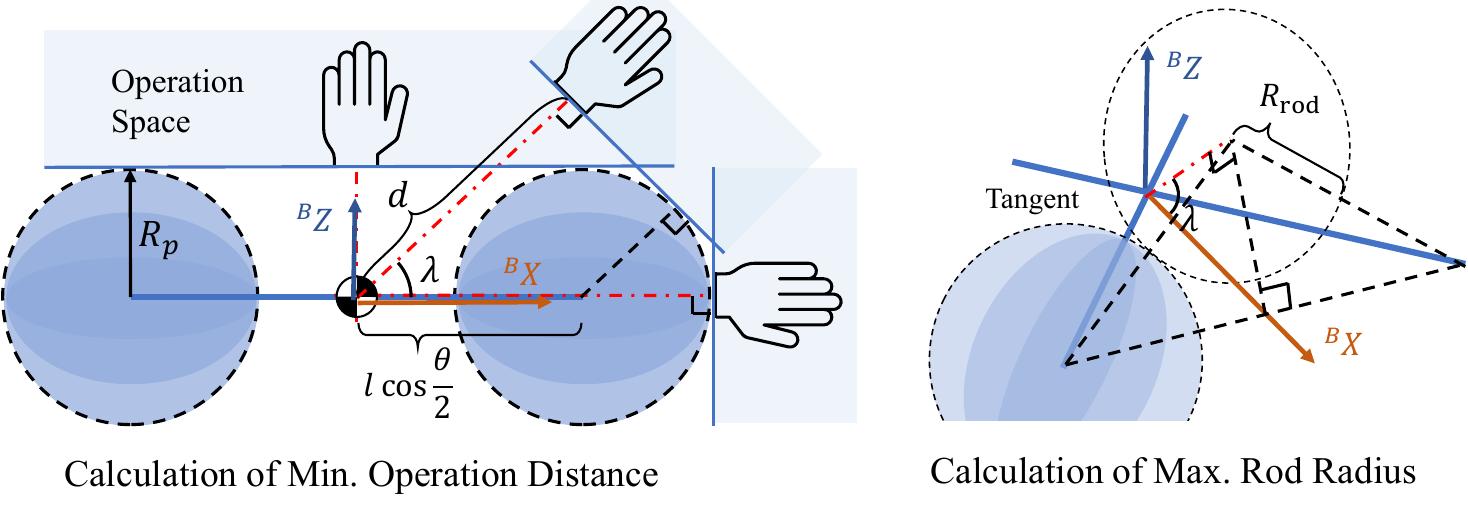}
        \label{fig:diff_opera_range_illus}
    } \\ \vspace{-2mm}
    \subfloat[Available operation distance and feasible rod radius for $N_p=4$]{
    \includegraphics[trim=0 0 0 0,clip,width=3.4in]{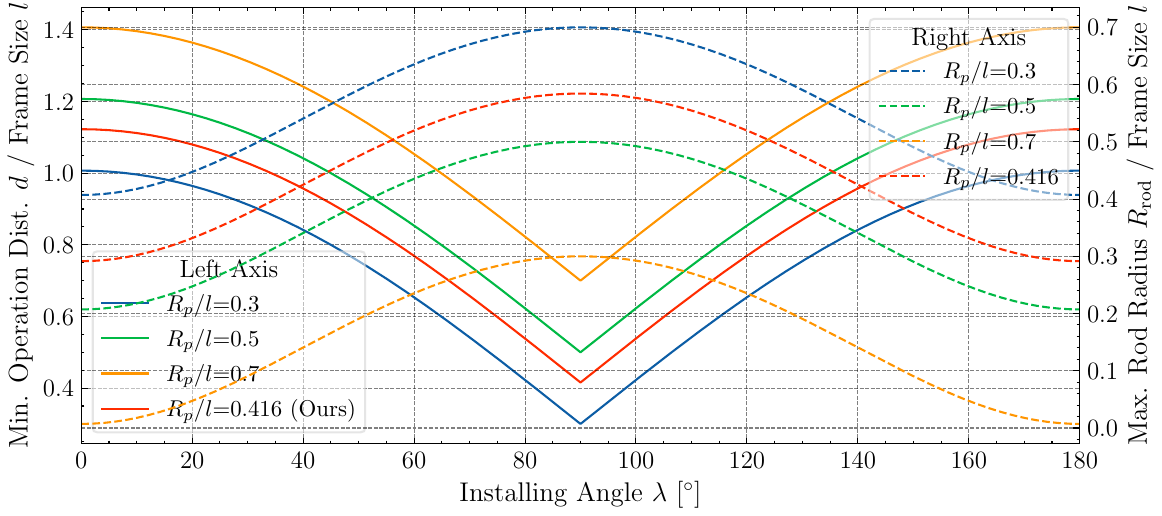}
        \label{fig:operate_dist_analysis}
    }
    \caption{Geometric analysis of the \revised{required operation distance} and feasible rod radius. Vertical mounting ($\lambda=90^\circ$) provides \revised{a short operation distance and a large allowable rod radius within the considered geometry}, making it suitable for large-wrench tasks.
    }
    \vspace*{-1mm}
    \label{fig:diff_ope_range}
\end{figure}

The geometric relations used in the analysis are illustrated in Fig.~\ref{fig:diff_opera_range_illus}. Based on these relations, varying the mounting angle changes the \revised{required operation distance} as
\begin{equation}
    \frac{d}{l} >
    \begin{cases}
        \cos(\frac{\theta}{2}) \ \cos(\lambda) + \frac{R_p}{l}, & \text{if} \ \lambda \leq \frac{\pi}{2}, \\
        \cos(\frac{\theta}{2}) \ \cos(\pi-\lambda) + \frac{R_p}{l}, & \text{if} \ \lambda > \frac{\pi}{2}.
    \end{cases}
    \label{eq:min_operate_dist}
\end{equation}
Another useful metric is the available space between two propellers for mounting the end-effector. We approximate the mounting rod as a cylinder. For the boundary case, its surface is tangent to the sphere swept by the tilted rotor. Based on the geometric relations, the \revised{allowable rod radius} varies with $\lambda$ as
\begin{equation}
    \frac{R_{\rm rod}}{l} <
    \begin{cases}
        \sqrt{\sin^2(\frac{\theta}{2})+\cos^2(\frac{\theta}{2}) \sin^2(\lambda)} - \frac{R_p}{l}, & \text{if} \; \lambda \leq \frac{\pi}{2}, \\
        \sqrt{\sin^2(\frac{\theta}{2})+\cos^2(\frac{\theta}{2}) \sin^2(\pi - \lambda)}- \frac{R_p}{l}, & \text{if} \; \lambda > \frac{\pi}{2}.
    \end{cases}
    \label{eq:max_rod_radius}
\end{equation}
Fig.~\ref{fig:operate_dist_analysis} plots these two quantities for the case of $N_p=4$ ($\theta=90^\circ$). Other rotor numbers follow the same trend.

Fig.~\ref{fig:operate_dist_analysis} shows that mounting the end-effector at $90^\circ$ yields \revised{a short operation distance within the considered geometry}, and a shorter distance generally enables more precise operation. 
At the same time, \revised{this mounting angle also provides a large feasible rod radius}, leaving more room for a stiffer mounting rod and reducing design difficulty. 
In addition, the curves shift upward or downward depending on the propeller-to-frame-size ratio $R_p/l$, revealing that although larger propellers may improve flight efficiency, they also increase the operation distance and decrease the available rod radius. As a compromise, we select $R_p/l=0.416$ for our robot.

\subsubsection{\revised{Available Wrench Envelope} in the Body Frame}

\setlength{\dbltextfloatsep}{8pt plus 1.0pt minus 2.0pt}
\begin{figure}[t] 
    \centering
    \subfloat[Illustration of force in (c), (d)]{\hspace{-0.0cm}
        \includegraphics[trim=8cm 7cm 19cm 7.5cm,clip,width=1.5in]{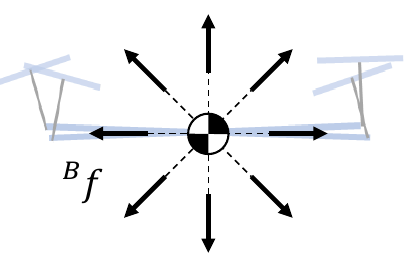}
        \label{fig:max_force_body_illu}
    }
    \subfloat[Illustration of torque in (e), (f)]{\hspace{+0.5cm}
        \includegraphics[trim=18cm 7cm 9cm 7.5cm,clip,width=1.5in]{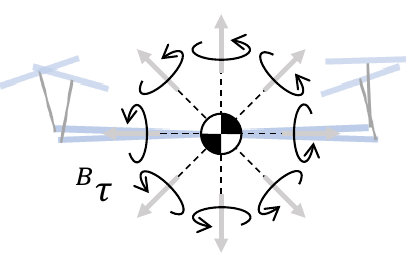}
        \label{fig:max_torque_body_illu}
    } \\
    \subfloat[3D \revised{available} force envelope]{\hspace{-0.2cm}
        \includegraphics[trim=0 0 0 0,clip,height=2.3in]{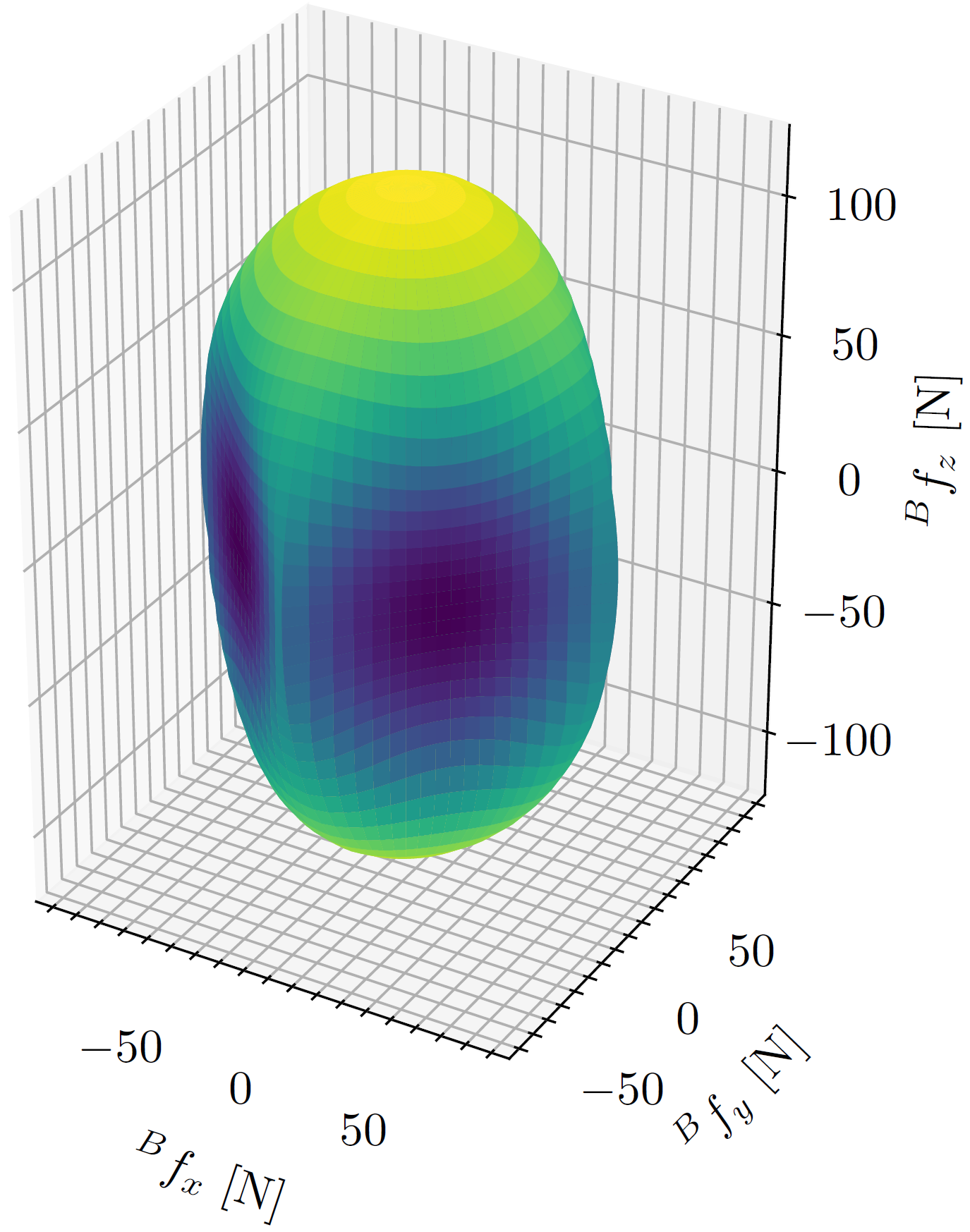}
        \label{fig:max_force_body_3d}
    }
    \subfloat[X-Z projection from -Y]{\hspace{-0.2cm}
        \includegraphics[trim=0 0 0 0,clip,height=2.3in]{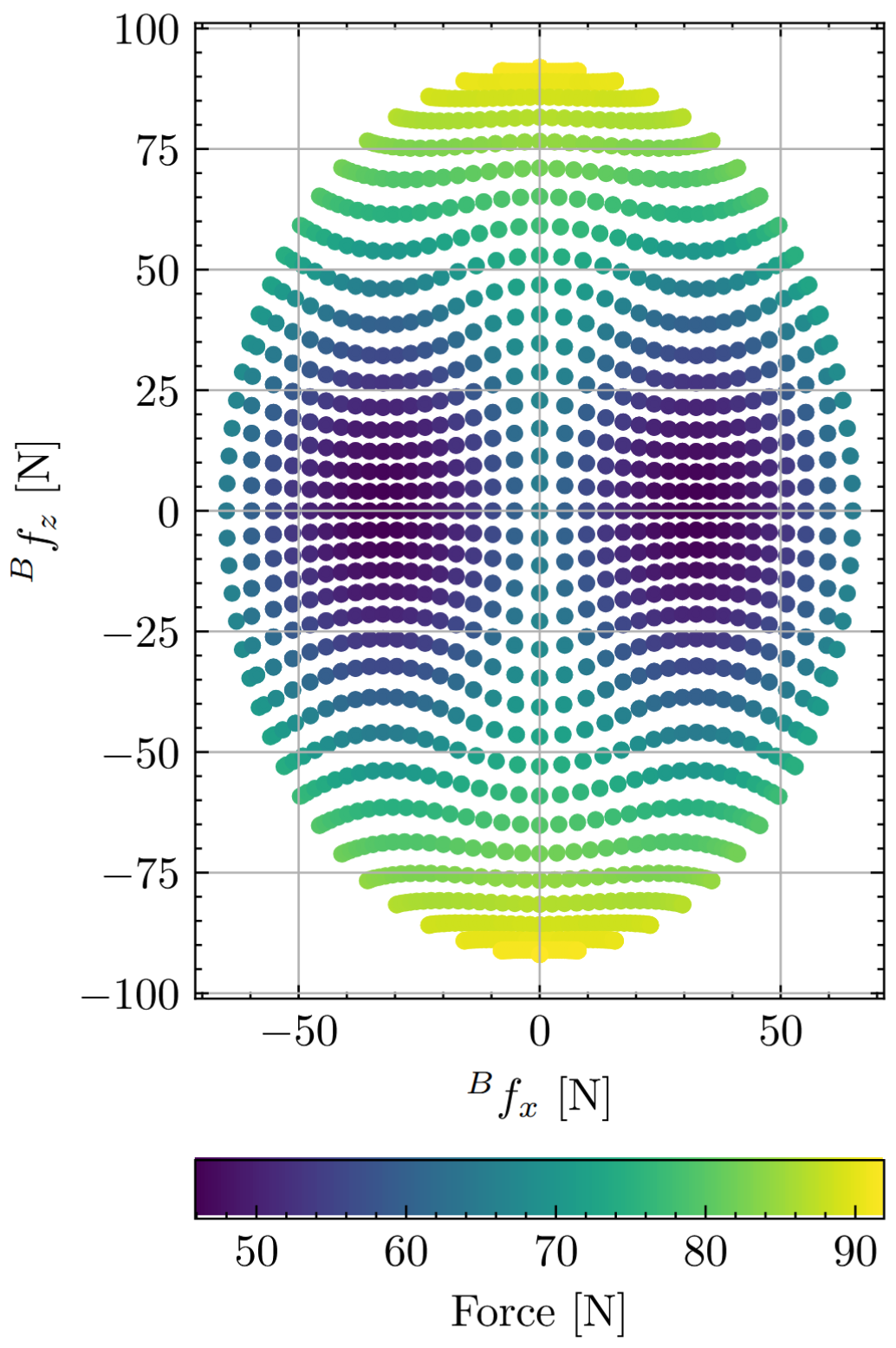}

        \label{fig:max_force_body_2d}
    }\vspace{-1mm}
    \subfloat[3D \revised{available} torque envelope]{\hspace{-0.2cm}
        \includegraphics[trim=0 0 0 0,clip,height=2.3in]{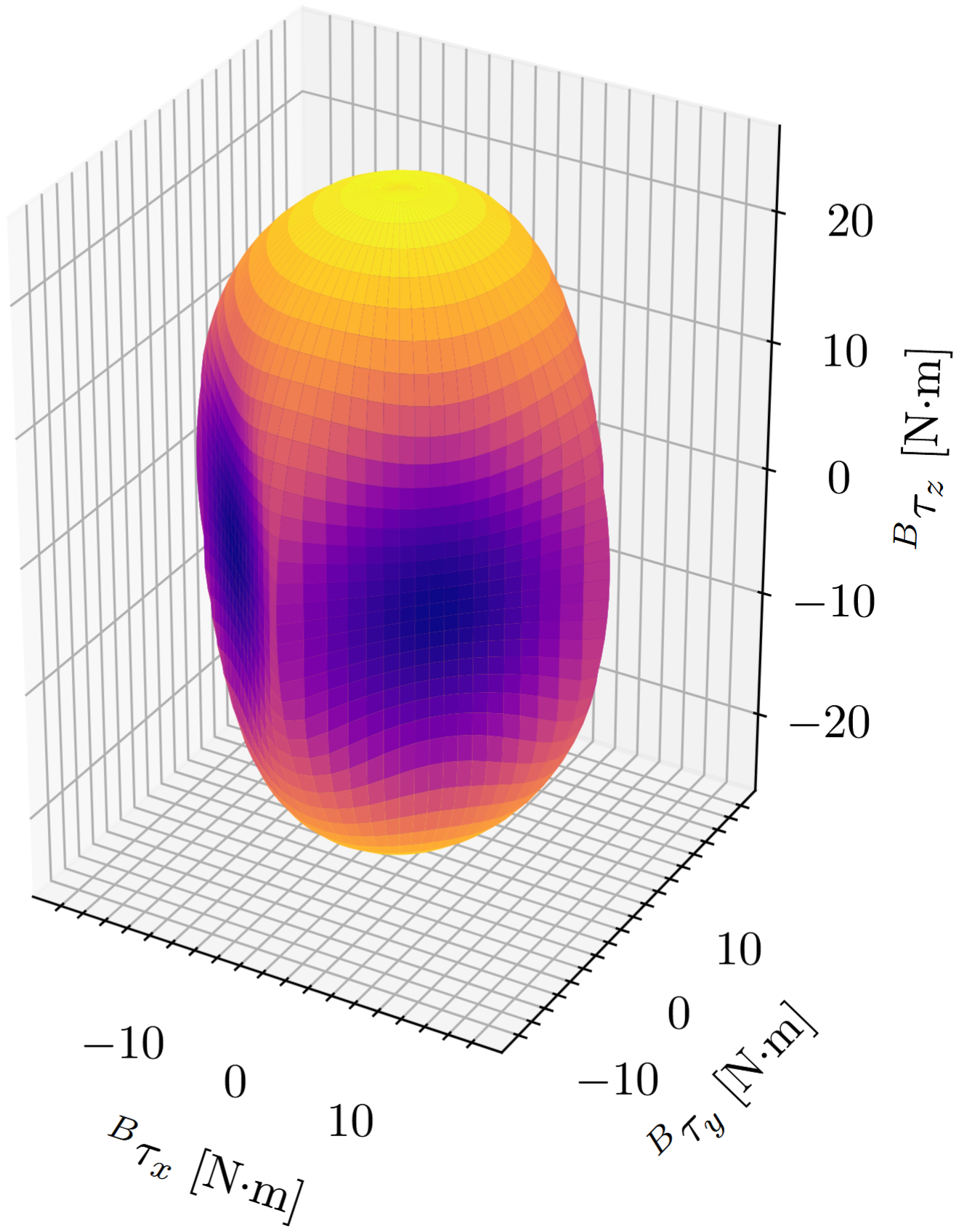}
        \label{fig:max_torque_body_3d}
    }
    \subfloat[X-Z projection from -Y]{\hspace{-0.2cm}
        \includegraphics[trim=0 0 0 0,clip,height=2.3in]{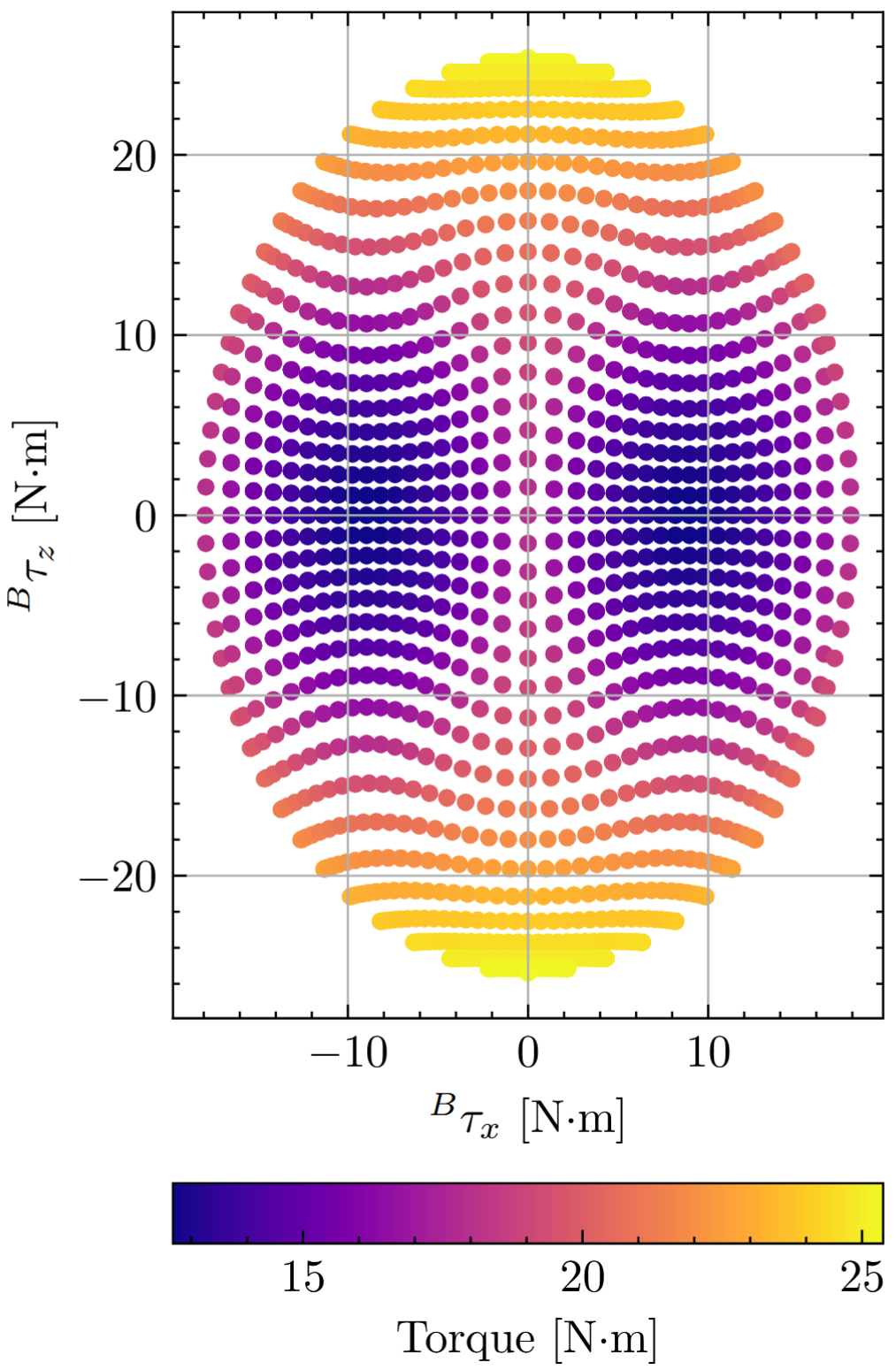}
        \label{fig:max_torque_body_2d}
    }\vspace{-1mm}
    \caption{Separated force and torque envelopes in the body frame at \SI{5}{\degree} resolution, neglecting gravity.
    Directions with smaller attainable wrench values correspond to larger internal force and hence lower efficiency. This trend remains unchanged when gravity is included.}
    \label{fig:max_wrench_body}
    \vspace*{-1mm}
\end{figure}

The \revised{available wrench envelope} is \revised{an important} criterion for determining a suitable mounting angle of the end-effector.
Since real-world manipulation tasks often require force and torque to be generated independently, we calculate the force envelope under zero torque and the torque envelope under zero force.
The algorithm samples wrench directions and performs a bisection search along each direction to obtain the boundary of feasible magnitude. The robot parameters are listed in Table~\ref{tab:mdl_ctrl_params}, and the results are shown in Fig.~\ref{fig:max_wrench_body}.
As shown in the figure, the computed body-frame envelope has \revised{large force and torque values} along the body Z-axis. 
Specifically, with a \revised{thrust limit} of \SI{23}{\newton} per rotor, the robot can generate \SI{90}{\newton} of force and more than \SI{22}{\newton\meter} of torque.
Since the underside is occupied by the landing gear, mounting the end-effector at the center and pointing it upward (as in Fig.~\ref{fig:coordinate}) \revised{seems to be a favorable option} for tasks requiring large wrench output.


\subsubsection{Task-Oriented Wrench Analysis}

\setlength{\textfloatsep}{8pt plus 1.0pt minus 2.0pt}
\begin{figure}[t]
    \centering
    \subfloat[Illustration of the task scenario and two work configurations]{
    \includegraphics[trim=4.5cm 4.9cm 4.5cm 5cm,clip,width=3.4in]{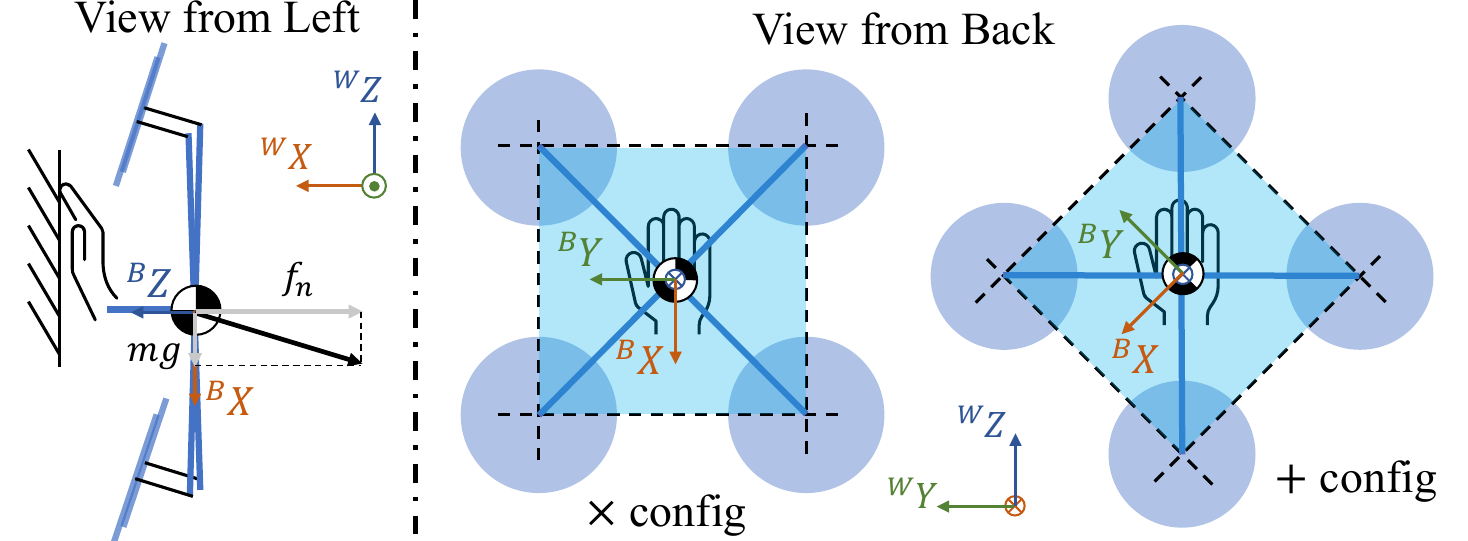}
        \label{fig:push_config_illus}
    } \\ \vspace{-2mm}
    \subfloat[\revised{Available} torque envelope under a required horizontal force]{
    \includegraphics[trim=0 0 0 0,clip,width=3.45in]{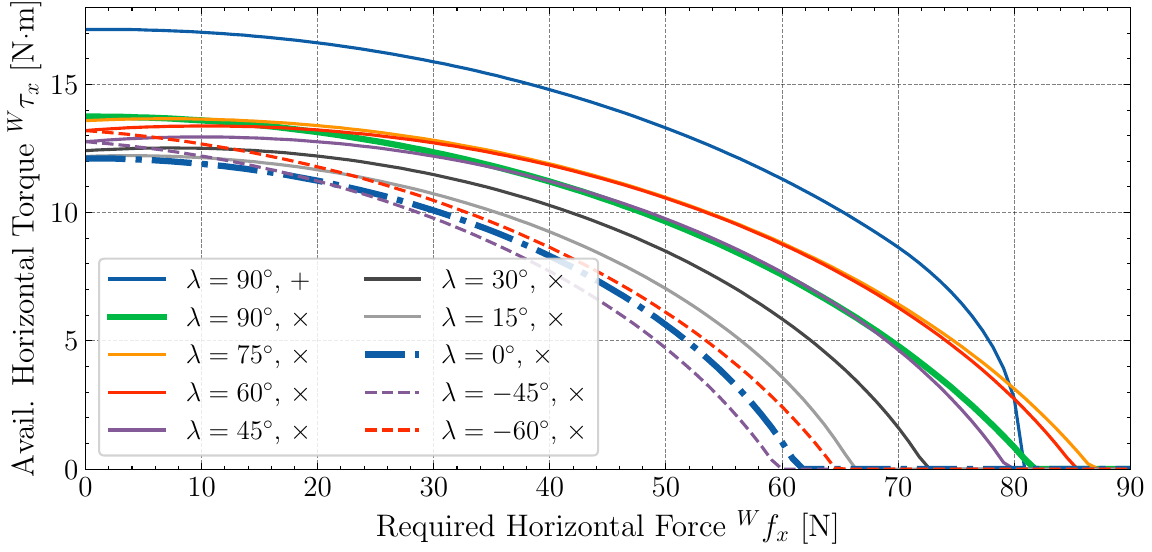}
    \label{fig:wrench_coupled_analysis}
    }
    \caption{Coupled wrench envelope for applying the wrench to a vertical surface. Vertical mounting ($\lambda=90^\circ$) provides a larger envelope than horizontal mounting ($\lambda=0^\circ$), and the $+$ configuration further enlarges the envelope. The $+$ configuration at other mounting angles may cause collisions with the propellers and is therefore omitted.
    }\vspace*{-1mm}
    \label{fig:wrench_coupled}
\end{figure}

\setlength{\dbltextfloatsep}{8pt plus 1.0pt minus 2.0pt}
\begin{figure}[t] 
    \centering
    \subfloat[Illustration of force in (c)]{\hspace{-0.0cm}
        \includegraphics[trim=8.5cm 7cm 20cm 7.5cm,clip,width=1.5in]{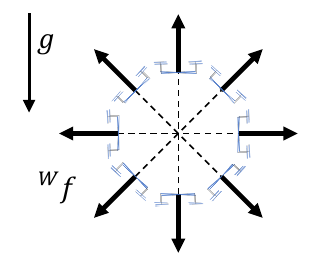}
        \label{fig:max_force_world_illu}
    }
    \subfloat[Illustration of torque in (d)]{\hspace{+0.5cm}
        \includegraphics[trim=19cm 7cm 9.5cm 7.5cm,clip,width=1.5in]{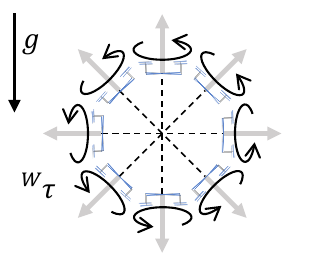}
        \label{fig:max_torque_world_illu}
    } \\
    \subfloat[\revised{Available} world force envelope]{\hspace{-0.2cm}
        \includegraphics[trim=0 0 0 0,clip,height=3.35in]{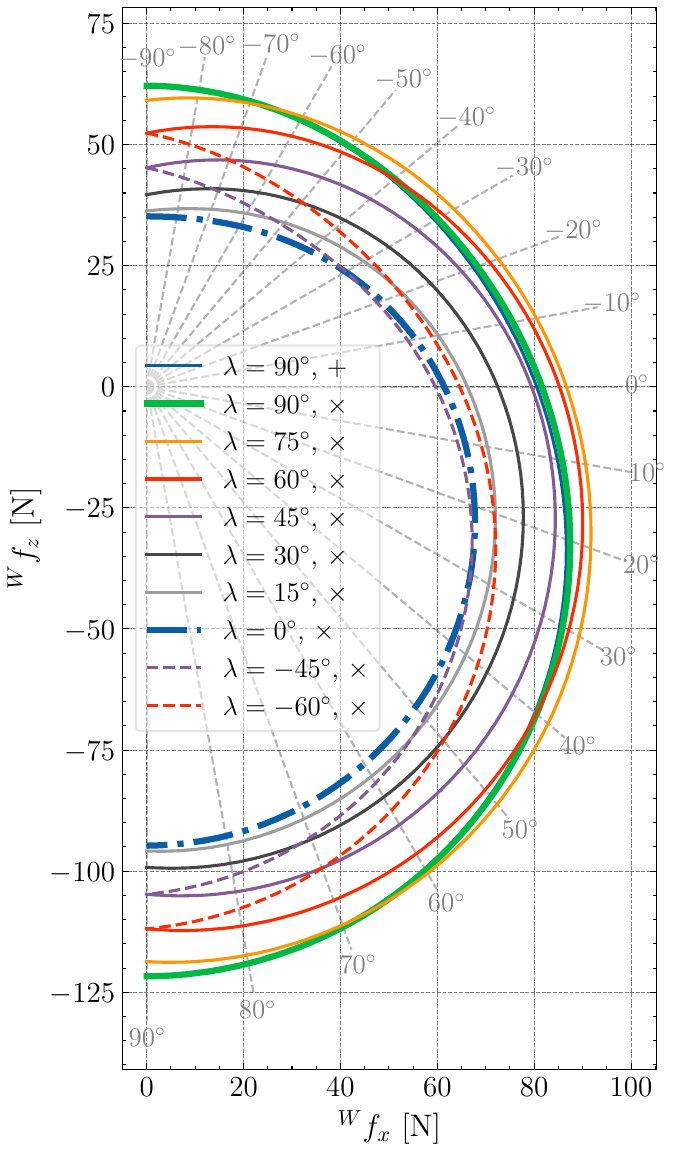}
        \label{fig:max_force_world_2d}
    }
    \subfloat[\revised{Available} world torque envelope]{\hspace{-0.2cm}
        \includegraphics[trim=0 0 0 0,clip,height=3.35in]{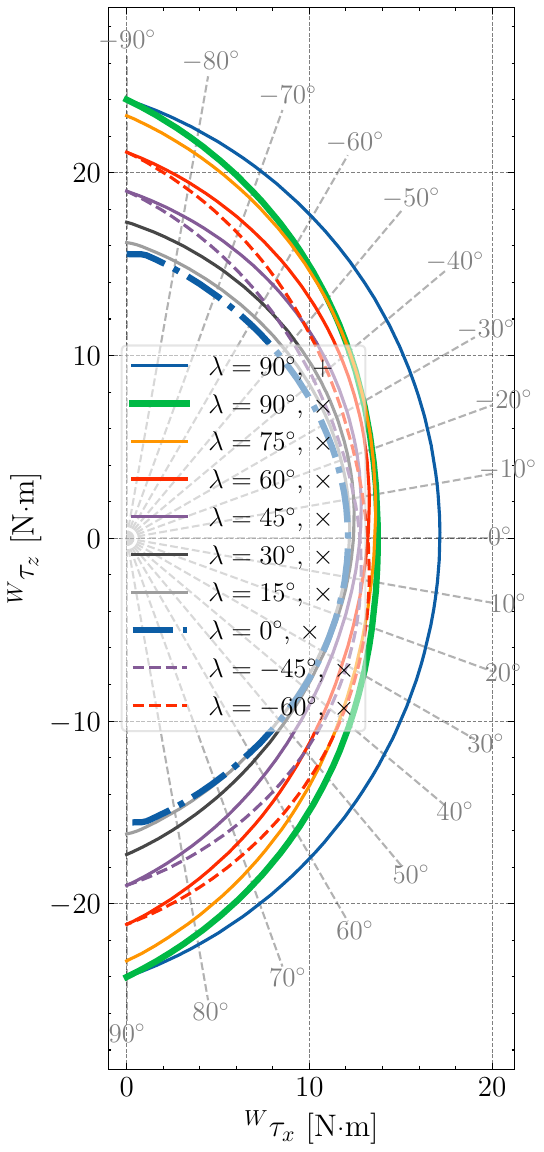}
        \label{fig:max_torque_world_2d}
    }
    \caption{Separated force and torque envelopes in the world frame for different mounting angles $\lambda$. The line connecting the origin to any point on the envelope indicates the pointing direction of the end-effector. The force envelope is clearly shifted downward by gravity, whereas the torque envelope remains symmetric about the horizontal plane. Vertical mounting ($\lambda=90^\circ$) provides larger envelopes than horizontal mounting ($\lambda=0^\circ$).}
    \vspace*{-1mm}
    \label{fig:max_wrench_world}
\end{figure}


Although the envelope in Fig.~\ref{fig:max_wrench_body} shows that \revised{large available force and torque values occur} along the body Z-axis, its implication in the gravity-influenced world frame still requires further analysis. We first consider a representative task that requires coupled horizontal force and torque generation. We then generalize the analysis by examining the \revised{available wrench envelope} in the world frame, which provides a reference for broader task scenarios.


The selected coupled-wrench task and the corresponding results are shown in Fig.~\ref{fig:wrench_coupled}.
This task requires the robot to generate torque while pushing a vertical wall, and can be regarded as a representative wall-contact task. 
From Fig.~\ref{fig:push_config_illus}, the \textit{aerial zero moment point (ZMP)} concept \cite{nishio_design_2024} provides \revised{an intuitive static interpretation, rather than a general stability proof}: when the contact force exceeds the weight, vertical mounting yields a larger projected support polygon and is expected to be more stable \revised{in this wall-contact scenario}.
Fig.~\ref{fig:wrench_coupled_analysis} further shows that vertical mounting provides a larger coupled-wrench envelope than horizontal mounting. Interestingly, \revised{the $75^\circ$ mounting angle forms the boundary of the coupled-wrench envelope}, and a possible reason is discussed together with the next analysis.

To generalize this wrench-based task to arbitrary attitudes, Fig.~\ref{fig:max_wrench_world} plots the separated force and torque envelopes in the world frame with gravity taken into account. Owing to symmetry about the world Z-axis, only a cross-section is shown. Compared with the body-frame envelope in Fig.~\ref{fig:max_wrench_body}, gravity influences force and torque differently: the force envelope is mainly shifted downward, whereas the torque envelope remains symmetric about the horizontal plane. Interestingly, the \revised{largest world-frame force envelope is achieved} around $\lambda=75^\circ$. We find that \revised{this angle} is theoretically related to the \revised{thrust-to-weight ratio limit} (MaxTWR) and is given by $\lambda^*=\arccos(1/\rm{MaxTWR})$. In our case, $\rm{MaxTWR}=3.086$ yields $71^\circ$, in good agreement with the observation. For torque generation, $\lambda=90^\circ$ provides the largest envelope. Moreover, the $+$ configuration (see Fig.~\ref{fig:push_config_illus} for the definition) further enlarges the overall torque envelope \revised{while maintaining a similar force envelope}, making it \revised{a useful} choice if the singularity can be properly handled.

\subsubsection{Mounting-Angle Selection}

Compared with horizontal mounting, vertical mounting provides a shorter operation distance, a larger allowable rod size, and larger wrench envelopes in the world frame.
Although \revised{$75^\circ$ provides a slightly larger force envelope under gravity}, vertical mounting remains \revised{competitive and provides a large torque envelope}.
Therefore, for tasks that require large wrench output, vertical mounting is \revised{a practical choice under the considered design criteria}. On the other hand, for tasks in which positioning accuracy or other criteria are more important than wrench capability, this advantage becomes less pronounced. Overall, the choice of mounting angle is task-dependent.


\section{Reference Generation \& Allocation} \label{sec:ref}

Although the NMPC optimizer can in theory work with only ${^W\boldsymbol{p}_r}$ and ${^W_B\boldsymbol{q}_r}$, providing additional reference states yields initial solutions closer to the optimum and often leads to faster convergence in practice. This procedure is worthwhile because it adds negligible computational burden. Accordingly, in this section, we leverage an allocation algorithm to generate full-state references for the controller. In addition, we propose a computationally efficient approach to handle singular points in omnidirectional flight. This approach exploits overactuation, highlighting the advantage of four tiltable rotors over three.

\revised{Note that related work~\cite{ryll_novel_2015,kamel_voliro_2018,cuniato_allocation_2026} generally adopts a control-allocation structure, where actuator dynamics and physical constraints can be considered together within the allocation~\cite{cuniato_allocation_2026}. By contrast, our approach uses allocation before the controller: the allocation module first generates coarse full-state references at negligible computational cost, and the NMPC then refines them considering the nonlinear dynamics and constraints.}

\subsection{Generation of Control Reference}






The control task can generally be divided into two categories: setpoint regulation and trajectory tracking. In the former, the same state is used throughout the NMPC horizon with zero reference velocity and acceleration. In the latter, references are time-shifted, and ${^W\boldsymbol{p}_{r}}$ and ${^W_B\boldsymbol{q}_r}$ are differentiated to obtain $^W\boldsymbol{v}_{r}$, $^W\dot{\boldsymbol{v}}_{r}$, $^B\boldsymbol{\omega}_r$, and $^B\dot{\boldsymbol{\omega}}_r$. Assuming zero ${^W\boldsymbol{f}_{dm}}$ and ${^B\boldsymbol{\tau}_{dm}}$, and providing ${^W\boldsymbol{f}_{de,r}}$ and ${^B\boldsymbol{\tau}_{de,r}}$ based on the task, the references ${^B\boldsymbol{f}_{u,r}}$ and ${^B\boldsymbol{\tau}_{u,r}}$ can be obtained from (\ref{eq:rigid_body_2}) and (\ref{eq:rigid_body_4}).
\revised{In this work, pure tracking uses trajectory references, whereas manipulation uses setpoint references generated from teleoperation commands.}

Based on these states, the reference rotor thrusts $f_{ir}$ and servo angles $\alpha_{ir}$ are calculated using pseudo-inverse allocation (PInv). First, we introduce a virtual input vector $\boldsymbol{z}$
\begin{equation} \label{eq:define_z}
    \begin{aligned}
        \boldsymbol{z} := \left[\begin{array}{c}
        f_{1r,h} \\
        f_{1r,v} \\
        \vdots \\
        f_{N_pr,h} \\
        f_{N_pr,v}
        \end{array}\right] = \left[\begin{array}{c}
        f_{1r} \sin{\alpha_{1r}} \\
        f_{1r} \cos{\alpha_{1r}} \\
        \vdots \\
        f_{N_pr} \sin{\alpha_{N_pr}} \\
        f_{N_pr} \cos{\alpha_{N_pr}}
        \end{array}\right],
    \end{aligned}
\end{equation}
which satisfies the equation
\begin{equation} \label{eq:Az}
    \left[^B\boldsymbol{f}_{u,r}, ^B\boldsymbol{\tau}_{u,r} \right]^\top = \boldsymbol{A} \ \boldsymbol{z},
\end{equation}
where the allocation matrix $\boldsymbol{A}$ can be derived from (\ref{eq:resultant_wrench}) using symbolic computation tools. Then the reference values $f_{ir}$ and $\alpha_{ir}$ are calculated as
\begin{subequations} \label{eq:pinv}
\begin{align}
\boldsymbol{z} &= \boldsymbol{A}^{\dagger} \ \left[^B\boldsymbol{f}_{u,r}, ^B\boldsymbol{\tau}_{u,r} \right]^\top, \\[5pt]
f_{ir} &= \sqrt{f_{ir,h}^2 + f_{ir,v}^2}, \quad \forall i=1,\cdots,N_p \\[5pt]
\alpha_{ir} &= {\rm atan2}\left(f_{ir,h}, f_{ir,v}\right), \quad \forall i=1,\cdots,N_p \label{eq:atan2}
\end{align}
\end{subequations}
where $(\cdot)^{\dagger}$ denotes the Moore--Penrose inverse. $\boldsymbol{A}^{\dagger}$ can either be kept fixed to reduce computation or updated at each control iteration using the current CoG for higher model accuracy. Note that $\alpha_{ir} \in \left[-\pi, \pi\right]$; thus, its value must remain continuous with the previous one for over-\SI{180}{\degree} rotation.


\subsection{Handling Singular Points and Physical Constraints}

\setlength{\dbltextfloatsep}{8pt plus 1.0pt minus 2.0pt}
\begin{figure}[t] 
    \centering
    \vspace{-2mm}
        \subfloat[Vertical cartwheel test case]{\hspace{-0.0cm}
        \includegraphics[trim=5.5cm 5.5cm 20cm 5.5cm,clip,width=1.4in]{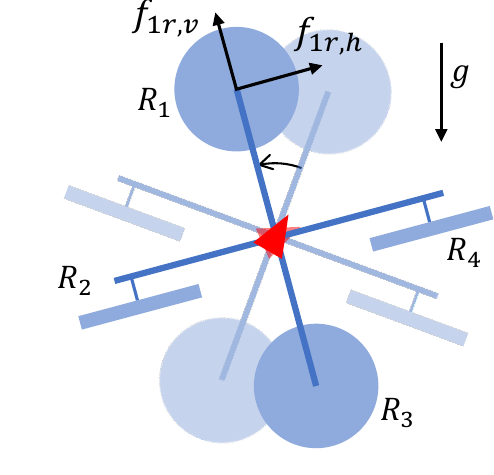}
        \label{fig:sing_exp_ill}
    }
    \subfloat[Illustration of the heuristic term (\ref{eq:heuristic})]{\hspace{1.0cm}
        \includegraphics[trim=19.5cm 6cm 7.0cm 6cm,clip,width=1.4in]{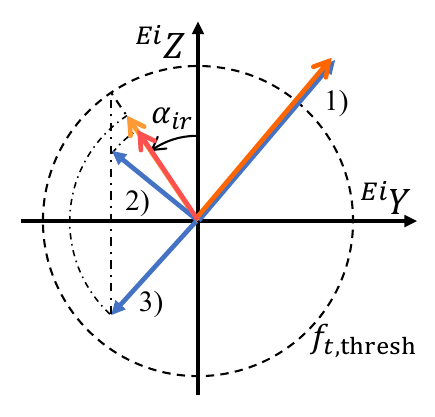}
        \label{fig:sing_mtd_ill}

    }\vspace{-1mm}
    \caption{Illustration of singularity test motion (a) and the heuristic term (b). In (b), if $f_{ir} \geq f_{t,\rm thresh}$, the angle is unchanged as in vector 1). If $f_{ir} < f_{t,\rm thresh}$ as in vectors 2) and 3), the heuristic term modifies them into the $+Z$ plane, avoiding motor-cable winding. When $f_{ir}=0$, $\alpha_{ir} \leftarrow \SI{0}{\degree}$.}
    \label{fig:illus_singular_heuristic_term}
\end{figure}

\begin{figure}[t]
    \centerline{\includegraphics[trim=0.2cm 0cm 0cm 0.2cm,clip,width=3.5in]{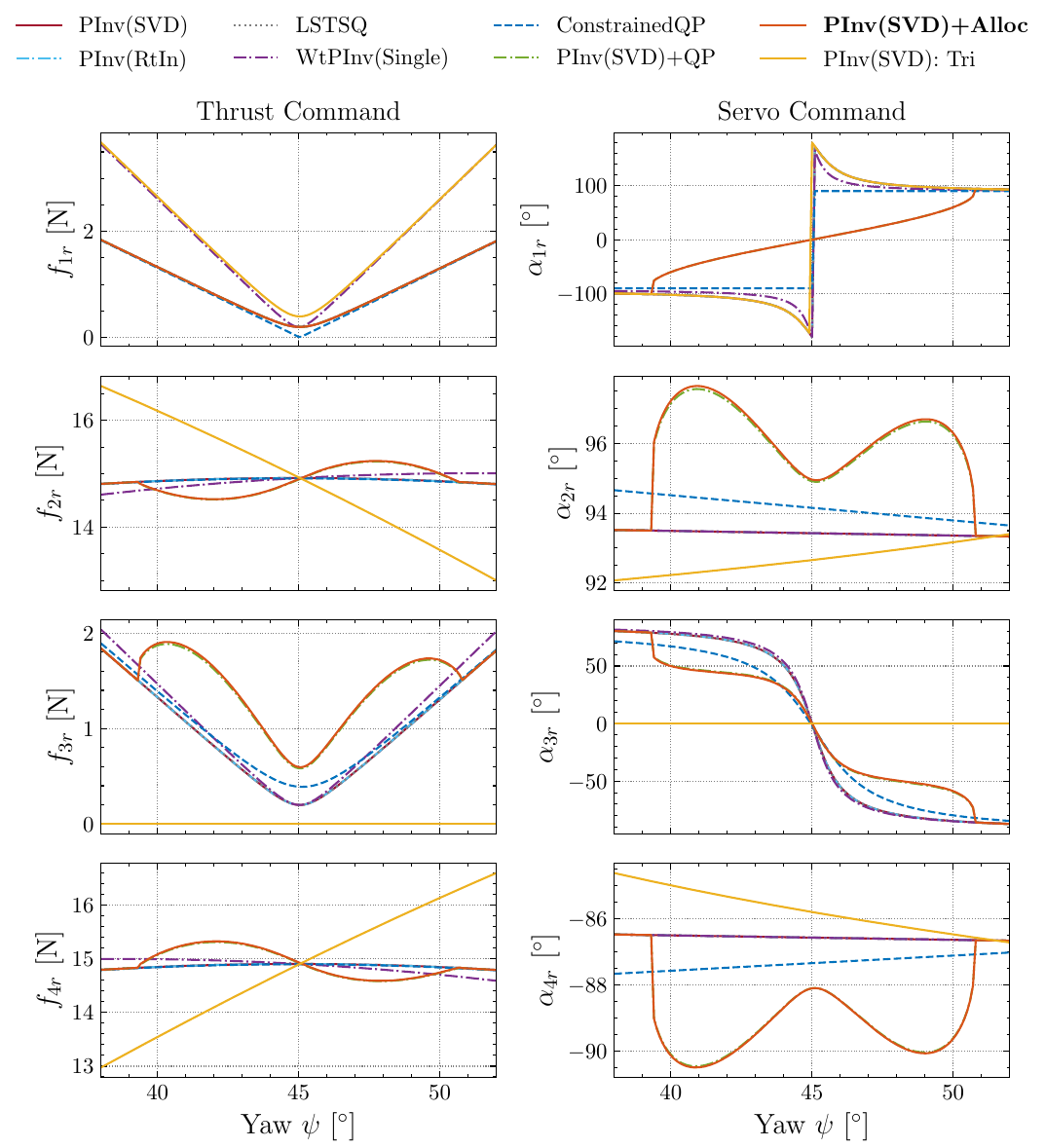}} 
    \vspace*{-3mm}
    \caption{Comparison of different allocation methods when traversing singular points. PInv: pseudo-inverse allocation. For the tiltable-trirotor test, rotor 3 is always shut down. The tiltable-quadrotor outperforms the tiltable-trirotor by exploiting the null space to pass singular points without winding the cables. 
    }
    \label{fig:singularity}
\end{figure}

The above reference generation method seeks solutions with minimum rotor speeds, which works well under most flight conditions. However, two problems arise in omnidirectional flight.
First, the method encounters sudden jumps at singular points. Note that these singularities do not originate from the allocation matrix $\boldsymbol{A}$, which can be viewed as constant and full rank. Instead, they stem from the $\texttt{atan2}(y,x)$ function in (\ref{eq:atan2}): when the desired force crosses the $i$th arm, $\alpha_{ir}$ changes abruptly by \SI{180}{\degree} \cite{cuniato_allocation_2026}. For example, when the robot rotates as shown in Fig.~\ref{fig:sing_exp_ill}, $f_{1r,v}$ is nearly zero while $f_{1r,h}$ changes from negative to 0 and then to positive, causing $\alpha_{1r}$ to jump from \SI{-90}{\degree} to \SI{0}{\degree} and then to \SI{+90}{\degree}.
Second, a more critical problem is the violation of physical constraints. The tilting structure typically powers each motor through a cable, which restricts the servo angle range. As illustrated in Fig.~\ref{fig:singularity}, \textit{PInv} allocation may cause $\alpha_{1r}$ to rotate in the opposite direction and wind the motor cable. Although an NMPC with a sufficiently long horizon can eventually determine the correct tilting direction, this issue still induces non-smooth motion.

To address these problems, \cite{allenspach_design_2020, cuniato_allocation_2026} employ \textit{differential allocation} and its variants to incorporate servo dynamics and constraints. 
Within the NMPC framework, the issues can be resolved either by modifying the NMPC cost or by adjusting the reference. We adopt the latter since it avoids increasing the complexity of NMPC optimization. Moreover, to keep reference generation simple and efficient while leaving dynamics to NMPC, we treat it purely at the geometric level without differentiation.

Specifically, we first perform pseudo-inverse allocation and then apply an additional refinement step. If $f_{ir}$ is smaller than a threshold, indicating proximity to a singular point, we manually introduce a heuristic term (visualized in Fig.~\ref{fig:sing_mtd_ill}) to correct the tilting angle $\alpha_{ir}$ of that rotor:
\begin{equation}
    \alpha_{ir} =
    \begin{cases}
        {\rm atan2}\left(f_{ir,h}, f_{ir,v}\right), & \text{if} \ f_{ir} \geq f_{t,\rm thresh}, \\
        \pi / 2 - {\rm arccos}\left( \frac{f_{ir,h}}{f_{t,\rm thresh}} \right), & \text{if} \ f_{ir} < f_{t,\rm thresh},
    \end{cases}
    \label{eq:heuristic}
\end{equation}

where $f_{t,\rm thresh}$ is a geometry-related constant that can be computed before takeoff through the torque balance between gravity and thrust:
\begin{equation}
    f_{t,\rm{thresh}} = \max_{1 \le i \le N_p} \frac{m \ g \ \lvert {^Bp_{{E_i},z}} \rvert}{ \bigl\lVert {^B\boldsymbol{p}_{E_{i},xy}}\bigr\rVert} + f_{t,\epsilon}, 
\end{equation}
and $f_{t,\epsilon}$ is a bias term to improve robustness.
After modifying $\alpha_{ir}$ while keeping $f_{ir}$ unchanged, the allocation downgrades from overactuation (eight free inputs) to full actuation (six free inputs), and its inverse is computed via LU decomposition and used to calculate the references of the other actuators.
In cases where two thrusts fall below the threshold, the larger-thrust rotor is selected.
\revised{The heuristic term (\ref{eq:heuristic}) only guides the NMPC and may be imperfect. It is discontinuous when the threshold is crossed with $f_{ir,v}<0$, which may cause chattering-like reference changes under repeated crossings. Nevertheless, for the unidirectional rotations considered in our experiments, this discontinuity could be handled by the NMPC without inducing observable chattering.}

\setlength{\textfloatsep}{8pt plus 1.0pt minus 2.0pt}

\vspace{-2mm}
\subsection{Evaluation}

We use a vertical cartwheel maneuver shown in Fig.~\ref{fig:sing_exp_ill} for evaluation. The tested allocation methods are
\begin{itemize}
    \item \textit{PInv(SVD)}: pseudo-inverse computed via singular value decomposition (SVD);
    \item \textit{PInv(RtIn)}: pseudo-inverse obtained using the right inverse;
    \item \textit{LSTSQ}: direct least-squares solution of (\ref{eq:Az});
    \item \textit{WtPInv(Single)}: weighted pseudo-inverse computed using SVD, with a weight of 1.5 on $f_{1r,h}$, indicating reduced preference for vertical force generation;
    \item \textit{ConstrainedQP}: quadratic programming (QP) with the constraint that $f_{1r,v}$ remains positive;
    \item \textit{PInv(SVD)+QP}: SVD-based initial allocation followed by fixing $\alpha_{1r}$ according to (\ref{eq:heuristic}), with the remaining references obtained via QP;
    \item \textit{\textbf{PInv(SVD)+Alloc} (proposed)}: SVD-based initial allocation followed by fixing $\alpha_{1r}$ according to (\ref{eq:heuristic}), with the remaining references computed via matrix inversion (LU decomposition);
    \item \textit{PInv(SVD): Tri}: pseudo-inverse computed via SVD for a tiltable-trirotor configuration.
\end{itemize}
All algorithms are implemented in Python and executed on a VIM4 onboard PC, with details provided in Sec. \ref{sec:robot}. The resulting reference values and computation times are presented in Fig.~\ref{fig:singularity} and Tab.~\ref{tab:alloc_time}, respectively.

The proposed \textit{PInv(SVD)+Alloc} approach is designed to overcome singularity points and satisfy physical constraints with an acceptable computational cost, and the results confirm this expectation.
As shown in Fig.~\ref{fig:singularity}, \textit{PInv(SVD)}, \textit{PInv(RtIn)}, and \textit{LSTSQ} yield identical results. \textit{WtPInv(Single)} has no significant impact on the result. \textit{ConstrainedQP} produces a step-like curve for $\alpha_{1r}$, failing to resolve the singularity-induced discontinuity. Both two-step methods, \textit{PInv(SVD)+QP} and \textit{PInv(SVD)+Alloc}, successfully generate the required references, with the latter offering significantly higher computational efficiency, as shown in Tab. \ref{tab:alloc_time}. Finally, the tiltable-trirotor using \textit{PInv(SVD)} exhibits the same singularity-induced discontinuity but lacks the additional null space to overcome it.
The ability to specify a rotor’s thrust and servo angle can also be extended to other applications, such as shutting down one rotor or enabling continuous pitch rotation \cite{cuniato_allocation_2026}.



\begin{table}[t]
\setlength{\abovecaptionskip}{0pt} 
\setlength{\belowcaptionskip}{-1pt}
\centering
\caption{Calculation Time of Different Allocation Methods}
\begin{tabular*}{\linewidth}{@{\extracolsep{\fill}} l c|l c }
 \toprule
 \textbf{Method} & \textbf{Time [ms]} & \textbf{Method} & \textbf{Time [ms]} \\
 \midrule
  PInv (SVD) & 0.103 & PInv (RtIn) & 0.101 \\
  LSTSQ & 0.181 & WtPInv (Single) & 0.103 \\
  Constrained QP & \hspace{-2.2mm} 18.780 & PInv (SVD) + QP & \hspace{-2.2mm} 14.815 \\
  \textbf{PInv (SVD) + Alloc.} & 0.310 & PInv (SVD) for Tri & 0.195 \\
 \bottomrule
\end{tabular*}
\label{tab:alloc_time}
\vspace{-1mm}
\end{table}


\section{Handling Model Error \& External Wrench} \label{sec:error}

For trajectory tracking of multirotors, external wrenches (e.g., wind gusts) are typically regarded as disturbances to be rejected and are often compensated together with model error.

In contrast, aerial manipulation generally requires the robot to react to external wrenches with different behaviors, necessitating separate treatment of the external wrench and model error. Existing works \cite{ryll_6d_2019, bodie_active_2021, zhao_versatile_2022} adopt the momentum-based method \cite{tomic_external_2017} for wrench estimation and typically use a pre-calibrated constant offset to compensate for model error.
The model-error compensator is often disabled during manipulation, possibly due to its potential conflict with wrench-based behavior.
In trajectory tracking, model error is usually compensated based on the pose error, such as through the integral term in PID control, with the aim of ensuring $\boldsymbol{x} - \boldsymbol{x}_r \rightarrow \boldsymbol{0}$. However, the controller in aerial manipulation may perform compliant behavior, changing the aim to $\boldsymbol{x} - \boldsymbol{x}_r \rightarrow \textbf{c} \neq \boldsymbol{0}$. This nonzero error is also observed during tracking when the reference is set far from the current state. In such cases, the pose-error-based integral term continues to accumulate and leads to overshoot. To avoid this unnecessary accumulation under such references, we exploit the prediction property of NMPC and propose a new type of integral term, which learns model error from the prediction error rather than the reference error. We also design an acceleration-based wrench estimator for tiltable-multirotors. Finally, we show how these two methods can be combined for realistic wrench-based tasks \revised{and analyze the frequency response}.

\setlength{\textfloatsep}{8pt plus 1.0pt minus 2.0pt}
\begin{figure}[t]
    \centerline{\includegraphics[trim=7cm 1cm 5.5cm 2.8cm,clip,width=3.5in]{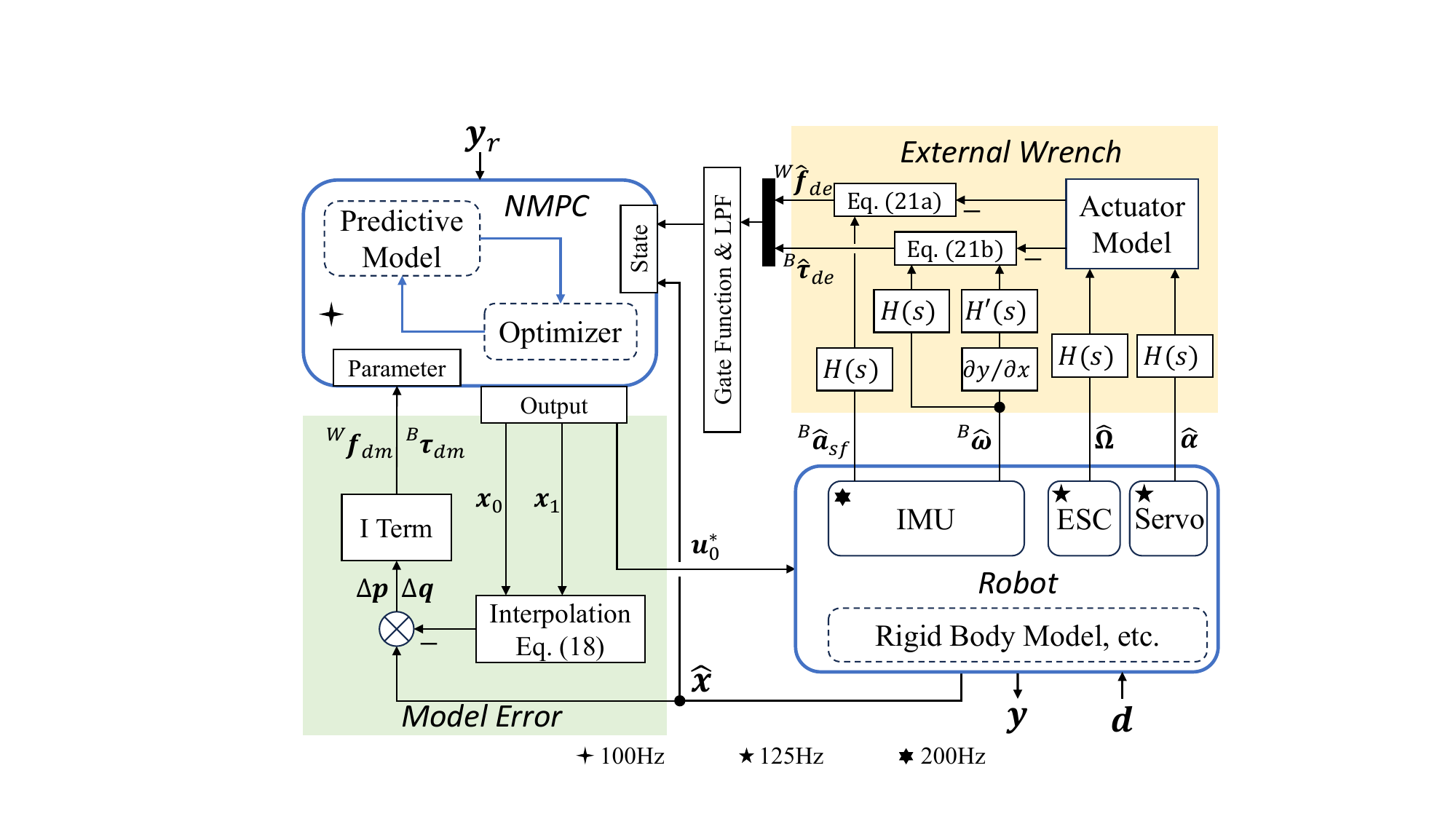}}
    \vspace*{-3mm}
    \caption{Diagram of the proposed framework for handling model error and external wrench in NMPC. All low-pass filters $H(s)$ use the same parameters to introduce identical delay. In practice, the estimates $^W\hat{\boldsymbol{f}}_{de}$ and $^B\hat{\boldsymbol{\tau}}_{de}$ are further filtered and processed by a gate function (\ref{eq:threshold_ext_wrench}) to enhance flight stability.}
    \label{fig:est_workflow}
\end{figure}

\subsection{Integral Term with Prediction Error}
\label{sec:integrator}

Inspired by the simplicity and effectiveness of the integral term in PID control, we propose a similar integrator to address model error in NMPC. Let the NMPC outputs be $\boldsymbol{x}_k$ and $\boldsymbol{u}_k$, where $k=0,\cdots,N-1$ denotes the prediction step. The errors in ${^W\boldsymbol{p}}$ and ${^W_B\boldsymbol{q}}$ are computed using \textit{spherical linear interpolation (slerp)} \cite{shoemake_animating_1985}:
\begin{subequations}  \label{eq:slerp}
\begin{align}
    \Delta \boldsymbol{p} &= {^W\hat{\boldsymbol{p}}}-\left({^W\boldsymbol{p}_0} + \left({^W\boldsymbol{p}_1} - {^W\boldsymbol{p}_0} \right) \cdot t_{s} \ / \ t_{\rm step}  \right), \\
    \Delta\boldsymbol{q} &= \left( \left({^W_B\boldsymbol{q}_1} \circ {^W_B\boldsymbol{q}^{-1}_0} \right)^{t_{s} / t_{\rm step}} \circ {^W_B\boldsymbol{q}_0}  \right)^{-1} \circ {^W_B\hat{\boldsymbol{q}}},
\end{align}
\end{subequations}
where $t_{s}$ and $t_{\rm step}$ denote the sampling time and prediction step in NMPC. Using this error, the model error is updated according to the following equations
\begin{subequations}  \label{eq:model_error}
\begin{align}
    {^W\boldsymbol{f}_{dm}} &= \text{Integrate} \left( \Delta\boldsymbol{p} \right), \\
    {^B\boldsymbol{\tau}_{dm}} &= \text{Integrate} \left( \text{sgn}\left(\Delta q_{w}\right) \cdot \mathcal{V}\left(\Delta \boldsymbol{q}\right) \right),
\end{align}
\end{subequations}
where ${\rm sgn}(\cdot)$ denotes the sign function, and ${\rm Integrate}(\cdot)$ represents the digital integrator with anti-windup (see \cite{beard_small_2012}, Ch. 6.5), expressed as
\begin{align}  \label{eq:digit_i_term}
    \begin{split}
        I'[k+1] &= I[k] + \frac{t_s}{2} \left(\Delta[k] + \Delta[k+1] \right), \\
        u'[k+1] &= k_I \ I'[k+1], \\
        u[k+1] &= \max\left( \min\left(u'[k+1], u_{\max}\right), u_{\min}\right), \\
        I[k+1] &= u[k+1] \, / \, k_I.
    \end{split}
\end{align}

\begin{figure}[t]
    \centering
    \includegraphics[trim=0 0 0 0,clip,width=3.5in]{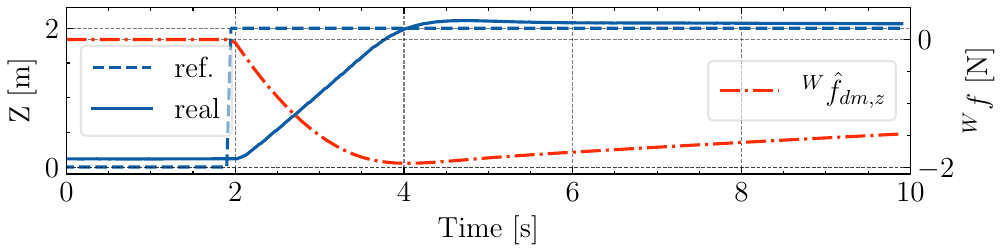}\\
    \includegraphics[trim=0 0 0 0,clip,width=3.5in]{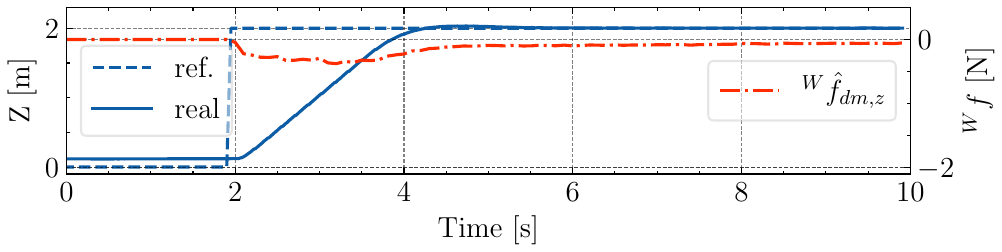} \vspace{-8mm}
    \caption{Simulated comparison of reference-error-based \cite{li_servo_2024} (\textbf{top}) and prediction-error-based (\textbf{bottom}) integral terms when tracking $^W p_z = \SI{2}{m}$. The former continues to accumulate until the reference is reached ($t=$\SI{4}{s}) and exhibits overshoot, whereas the latter is independent of the reference error and remains well behaved.}
    \label{fig:sim_iterm}
\end{figure}

To demonstrate the performance of the proposed integral term against the baseline, we conduct a waypoint-tracking task in Gazebo simulation and show the results in Fig.~\ref{fig:sim_iterm}. The robot is commanded to reach $^Wp_z=$\SI{2}{m} at about \SI{2}{s}. Under this condition, the reference-based baseline \cite{allenspach_design_2020, li_servo_2024} continues to accumulate until the state exceeds the reference at around \SI{4}{s}, whereas the prediction-based integral term remains well behaved. Consequently, the former causes overshoot and requires a longer time to settle. Although only the Z-axis response is shown for clarity, the same difference is observed along the other translational and attitude axes.

\setlength{\textfloatsep}{8pt plus 1.0pt minus 2.0pt}

\subsection{Acceleration-Based Wrench Estimation}  \label{sec:acc_wrench_est}

The principle of acceleration-based wrench estimation is straightforward: with rotor speed measurements and the actuator model, we can compute the nominal wrench generated by the robot. Simultaneously, with data from the inertial measurement unit (IMU), we can obtain the actual wrench experienced by the robot.
Then the external wrench is estimated as the difference between the actual and nominal wrenches. With $\hat{\boldsymbol{\Omega}}$ measured from the ESC 
and $\hat{\boldsymbol{\alpha}}$ from the servos, the nominal inputs ${^B\hat{\boldsymbol{f}}_{u,\rm nom}},{^B\hat{\boldsymbol{\tau}}_{u,\rm nom}}$ are computed from (\ref{eq:fu_and_tauu}) or (\ref{eq:Az}). Meanwhile, the specific force ${}^B\hat{\boldsymbol{a}}_{sf}$ and angular velocity ${}^B\hat{\boldsymbol{\omega}}$ are obtained from the IMU. Thus, we have
\begin{subequations} \label{eq:wrench_est}
\begin{align}
    {^W\hat{\boldsymbol{f}}_{de}} &= {^W_B\boldsymbol{R}}(\boldsymbol{q}) \ \left( m \, {^B\hat{\boldsymbol{a}}_{sf}} \;-\; {^B\hat{\boldsymbol{f}}_{u,\rm nom}} \right) - {^W\boldsymbol{f}}_{de,b}, \\
    {^B\hat{\boldsymbol{\tau}}_{de}} &= \boldsymbol{I} \, {^B\hat{\dot{\boldsymbol{\omega}}}} + {^B \hat{\boldsymbol{\omega}}} \times\left({\boldsymbol{{I}}} \ {^B\hat{\boldsymbol{\omega}}}\right) - {^B\hat{\boldsymbol{\tau}}_{u,\rm nom}} - {^B{\boldsymbol{\tau}}_{de,b}},
\end{align}
\end{subequations}
where $m$ is the mass, $\boldsymbol{I}$ is the inertia matrix, and ${^W\boldsymbol{f}}_{de,b}$ and ${^B{\boldsymbol{\tau}}_{de,b}}$ denote the initial bias obtained during CALIB (see next subsection).
Because all measurements contain noise and require filtering, it is necessary to ensure that all filtered data have the same delay \cite{smeur_adaptive_2016}. Specifically, $\hat{\boldsymbol{\Omega}},\hat{\boldsymbol{\alpha}},{{}^B\hat{\boldsymbol{a}}_{sf}}$ should be passed through the same low-pass filter (LPF) for synchronization, while the differentiator of $\hat{\boldsymbol{\omega}}$ and its associated LPF should be carefully designed to have a comparable total delay. Further details are provided in Sec.~\ref{sec:sysid_filter}.

\subsection{Combining the Integral Term and Wrench Estimation}  \label{sec:combinition}


The above integral term and wrench estimator are combined as shown in Fig.~\ref{fig:est_workflow}, where model errors are treated as NMPC parameters and external wrenches are treated as NMPC states.

To improve practical stability, we design a state machine with three modes: STOP, CALIB, and RUN. \revised{The robot starts in STOP and enters CALIB when the linear and angular velocities are below $v_{\rm th,calib}$ and $\omega_{\rm th,calib}$, respectively, and the estimated wrench magnitude is sufficiently small. During CALIB, wrench measurements are collected for $t_{\rm calib}$ and then averaged to obtain the bias terms ${^W\boldsymbol{f}}_{de,b}$ and ${^B\boldsymbol{\tau}}_{de,b}$. 
These bias terms arise from errors in the IMU, actuator measurements, inertial parameters, and assembly-related geometry.
The state then switches to RUN. If the linear or angular velocity exceeds $v_{\rm th,stop}$ or $\omega_{\rm th,stop}$, respectively, the state switches back to STOP,}
following the principle that the robot should remain stiff during manipulation for accuracy, but become compliant during motion for safety.



\begin{figure}[t]
    \centerline{\includegraphics[trim=0 0 0 0,clip,width=3.5in]{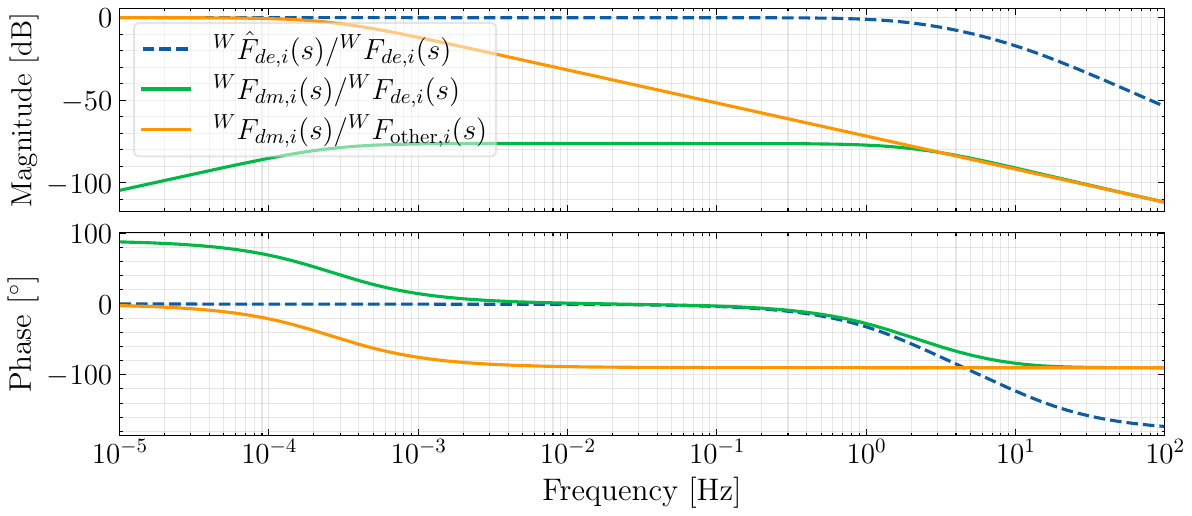}}
    \vspace*{-3mm}
    \caption{\revised{Frequency-response comparison of the wrench estimator and the integral term. The estimator captures external wrenches, while the integral term compensates for near-static residuals and responds weakly to captured wrenches, indicating their complementary rather than competing roles.}}
    \label{fig:freq_response}
\end{figure}

\revised{To clarify the complementary behaviors of these two modules, we analyze the force channel in the frequency domain.
The robot disturbance is decomposed into ${}^{W}\boldsymbol f_{de}+{}^{W}\boldsymbol f_{\rm other}$, where ${}^{W}\boldsymbol f_{de}$ is the external force and ${}^{W}\boldsymbol f_{\rm other}$ represents a residual force mainly caused by model errors.
Since the acceleration-based estimator reconstructs ${{}^{W}\hat{\boldsymbol f}_{de}}$ from measured acceleration and actuator states, its response is dominated by the filters and can be approximated as}
\begin{equation}
    \revised{\frac{{}^{W}\hat{F}_{de,i}(s)}{{}^{W} F_{de,i}(s)} \approx H_e(s)=H_{\rm post}(s)H_{\rm pre}(s), \; i\in\{x,y,z\},} \label{eq:freq_estimator}
\end{equation}
\revised{where $H_{\rm pre}(s)$ and $H_{\rm post}(s)$ denote the LPFs applied to the measurements and to the estimated result, respectively. The LPF parameters are given in Sec.~\ref{sec:sysid_filter}.} 
\revised{
Meanwhile, a second-order Taylor expansion of the measured and linearly interpolated predicted positions yields}
\begin{equation}
\revised{
{}^W\dot{\boldsymbol{f}}_{dm}
= \boldsymbol{K}_I \Delta\boldsymbol{p} \approx \frac{\boldsymbol{K}_I t_s^2}{2m}
\left(\boldsymbol f_{de} - \hat{\boldsymbol f}_{de}
+ \boldsymbol f_{\mathrm{other}} - \boldsymbol f_{dm}\right).
}
\end{equation}
\revised{Converting this approximation into the frequency domain and substituting (\ref{eq:freq_estimator}) yield the axis-wise transfer functions:}
\begin{equation} \label{eq:freq_response_integrator}
    \revised{\frac{{}^{W}{F}_{dm,i}(s)}{{}^{W}{F}_{de,i}(s)} \approx \frac{\omega_I}{s+\omega_I}\left(1-H_e(s)\right), \;
    \frac{{}^{W}{F}_{dm,i}(s)}{{}^{W}{F}_{{\rm other},i}(s)} \approx \frac{\omega_I}{s+\omega_I},} 
\end{equation}
\revised{where $\omega_I=k_{I,i} \; t_s^2/(2m)$ and $i\in\{x,y,z\}$. The frequency responses of (\ref{eq:freq_estimator}) and (\ref{eq:freq_response_integrator}) are plotted in Fig.~\ref{fig:freq_response}.}

\revised{This analysis and Fig.~\ref{fig:freq_response} show that the wrench estimator and the integral term have separated roles: the former acts as a filtered feedforward term for external wrenches, whereas the latter compensates for the remaining unmodeled wrenches under near-static conditions.
This separation is reflected by the factor $1-H_e(s)$ in the transfer function ${}^{W}{F}_{dm,i}(s)/{}^{W}{F}_{de,i}(s)$: when an external wrench is captured by the estimator, the corresponding residual seen by the integral term is attenuated, leading to a weak integral response. Hence, the two terms play complementary roles rather than competing with each other.}

\revised{The torque channel follows the same trend but with a lower effective bandwidth, because ${^B\hat{\dot{\boldsymbol{\omega}}}}$ is obtained by numerical differentiation and therefore requires stronger filtering.}


\setlength{\textfloatsep}{8pt plus 1.0pt minus 2.0pt}

\section{Nonlinear Model Predictive Control} \label{sec:control}

In this section, we propose an effector-centric NMPC framework for tiltable-multirotors. The NMPC controller serves as the central module that integrates the results from previous sections, including the system model from Sec.~\ref{sec:modeling}, the vertical mounting of an end-effector analyzed in Sec.~\ref{sec:design}, the references generated in Sec.~\ref{sec:ref}, and the model error term \& external wrench from Sec.~\ref{sec:error}. We first introduce the selection of states, inputs, and parameters, and then present the formulation of the NMPC optimization problem.

\subsection{Basic Configuration}

The NMPC involves three types of variables \cite{verschueren_acadosmodular_2022}: states $\boldsymbol{x_k}$, which are optimized; inputs $\boldsymbol{u_k}$, which are optimized and output as control commands; and parameters $\boldsymbol{p_k}$, which are not optimized but updated externally. In our formulation, we select $\boldsymbol{x} = \left[{^W\boldsymbol{p}}, {^W\boldsymbol{v}}, {^W_B\boldsymbol{q}}, {^B\boldsymbol{\omega}}, {\boldsymbol{\alpha}}, {^W\boldsymbol{f}_{de}}, {^B\boldsymbol{\tau}_{de}} \right]^\top$, $\boldsymbol{u} = \left[\boldsymbol{f}_c, \boldsymbol{\alpha}_c\right]^\top$, and $\boldsymbol{p} = \left[{^W\boldsymbol{f}_{dm}}, {^B\boldsymbol{\tau}_{dm}} \right]^\top$, where ${^W\boldsymbol{f}_{de}}, {^B\boldsymbol{\tau}_{de}}$ are the estimated wrenches from (\ref{eq:wrench_est}), and ${^W\boldsymbol{f}_{dm}}, {^B\boldsymbol{\tau}_{dm}}$ are the model errors accumulated in (\ref{eq:model_error}).
The parameter set can be further enriched to $\boldsymbol{p} = [{^W\boldsymbol{f}_{dm}}, {^B\boldsymbol{\tau}_{dm}}, \boldsymbol{I}, {^B\boldsymbol{p}_{E_i}, \ldots}, {^B\boldsymbol{p}_{E_{N_p}}} ]^\top$ or even extended to include all physical parameters, thereby dynamically accounting for CoG changes. Note that ${^W\boldsymbol{f}_{de}}, {^B\boldsymbol{\tau}_{de}}$ are included in the states to allow greater flexibility in modeling their dynamics, while ${^W\boldsymbol{f}_{dm}}, {^B\boldsymbol{\tau}_{dm}}$ are treated as parameters, assuming these slowly varying quantities remain constant over the prediction horizon.

\subsection{Effector-Centric Error Definition} \label{sec:nmpc_error}

To simplify reference generation for manipulation tasks, trajectories are generated in the end-effector frame {$\mathcal{T}$}, denoted as ${^W\boldsymbol{p}_{T_o,r}}$, ${^W\boldsymbol{v}_{T_o,r}}$, ${^W_T\boldsymbol{q}_r}$ and ${^T\boldsymbol{\omega}_r}$, where $T_o$ indicates the origin of {$\mathcal{T}$}. In contrast, the states used in model prediction ${^W\boldsymbol{p}_{B_o}}$, $^W\boldsymbol{v}_{B_o}$, ${^W_B\boldsymbol{q}}$, and ${^B\boldsymbol{\omega}}$, remain in the body frame {$\mathcal{B}$} for computational simplicity. Accordingly, a coordinate transformation is applied to construct the cost function with respect to the reference as
\begin{subequations}
\begin{align}
    {^W\boldsymbol{p}_{T_o}} &= {^W\boldsymbol{p}_{B_o}} + {^W_B\boldsymbol{R}}(\boldsymbol{q}) \; {^B\boldsymbol{p}_{T_o}}, \\
    {^W\boldsymbol{v}_{T_o}} &= {^W\boldsymbol{v}_{B_o}} + {^W_B\boldsymbol{R}}(\boldsymbol{q}) \; \left[^B\boldsymbol{\omega}\right]_\times  {^B\boldsymbol{p}_{T_o}}, \\
    {^W_T\boldsymbol{q}} &= {^W_B\boldsymbol{q}} \; \circ \; {^B_T\boldsymbol{q}}, \\
    {^T\boldsymbol{\omega}} &= {^B_T\boldsymbol{R}^\top} \; {^B\boldsymbol{\omega}},
\end{align}
\label{eq:ee-centric}
\end{subequations}
where ${^B\boldsymbol{p}_{T_o}},\ {^B_T\boldsymbol{q}}$ define the relative pose of the end-effector in {$\mathcal{B}$}, and $[\boldsymbol{\omega}]_\times$ denotes the skew-symmetric matrix of $\boldsymbol{\omega}$.
Using the transformed values above, the complete definition of the error terms is summarized in Tab.~\ref{tab:error_term}.

\begin{table}[t]
\setlength{\abovecaptionskip}{0pt} 
\setlength{\belowcaptionskip}{-1pt}
\centering
\caption{Definition of Error Terms in Effector-Centric NMPC}
\begin{tabular*}{\linewidth}{@{\extracolsep{\fill}} l l|l l }
 \toprule
 \textbf{Symbol} & \textbf{Definition} & \textbf{Symbol} & \textbf{Definition} \\
 \midrule
  ${^W\overline{\boldsymbol{p}}_k}$ & ${^W\boldsymbol{p}_{T_o}} - {^W\boldsymbol{p}_{T_o,r}}$ & ${^W\overline{\boldsymbol{v}}_{T_o,k}}$ & ${^W\boldsymbol{v}_{T_o}} - {^W\boldsymbol{v}_{T_o,r}}$ \\
  \rule{0pt}{3ex} ${^W_T\overline{\boldsymbol{q}}_k}$ & $\mathcal{V}\left({^W_T\boldsymbol{q}}^*_r \circ {^W_T\boldsymbol{q}} \right)$ & ${^T\overline{\boldsymbol{\omega}}_k}$ & ${^T\boldsymbol{\omega}} - {^T\boldsymbol{\omega}_r}$ \\
  \rule{0pt}{3ex} ${\overline{\boldsymbol{\alpha}}_k}$ & ${\boldsymbol{\alpha}}-{\boldsymbol{\alpha}_{c,r}}$ & ${^W\overline{\boldsymbol{f}}_{de,k}}$ &
  $-{^W\boldsymbol{f}_{de}} - \boldsymbol{0}$ \\
   ${^B\overline{\boldsymbol{\tau}}_{de,k}}$ & 
   $-{^B\boldsymbol{\tau}_{de}} - \boldsymbol{0}$
   & ${\overline{\boldsymbol{f}}_{c,k}}$ & $\boldsymbol{f}_{c} - \boldsymbol{f}_{c,r}$ \\
   ${\overline{\boldsymbol{\alpha}}_{c,k}}$ & ${\boldsymbol{\alpha}_{c} - \boldsymbol{\alpha}}$\cite{li_servo_2024} & & \\
 \bottomrule
\end{tabular*}
\label{tab:error_term}
\vspace{-1mm}
\end{table}

\subsection{NMPC Controller}

NMPC reformulates the control task as a constrained nonlinear optimization problem. With the cost function defined in Sec.~\ref{sec:nmpc_error}, the optimal control problem can be posed as a nonlinear least-squares problem:
\begin{subequations}  \label{eq:nmpc_opt_problem}
\begin{align}
    \underset{\boldsymbol{x}_k,\boldsymbol{u}_k}{\rm min \ } \quad &\sum\limits_{k=0}^{N-1}\left(\overline{\boldsymbol{x}}^\top_k \boldsymbol{{Q}} \overline{\boldsymbol{x}}_k + \overline{\boldsymbol{u}}^\top_k \boldsymbol{{R}} \overline{\boldsymbol{u}}_k\right) + \overline{\boldsymbol{x}}^\top_N \boldsymbol{{Q}}_N\overline{\boldsymbol{x}}_N, \label{eq:cost} \\
    {\rm s.t.} \quad 
    &\boldsymbol{x}_{k+1} = \boldsymbol{f}\left(\boldsymbol{x}_{k}, \boldsymbol{u}_{k},\boldsymbol{p}_{k}\right), \ k \in \{0,...,N\!-\!1\}, \label{eq:dyn_constraint} \\
    &\boldsymbol{x}_0 =\hat{\boldsymbol{x}}, \label{eq:init_constraint} \\
    &\left|v_{x,y,z}\right| \leq v_{\rm limit}, \quad \left|\omega_{x,y,z}\right| \leq \omega_{\rm limit}, \label{eq:state_constraint} \\
    &\boldsymbol{u}_{\min} \leq \boldsymbol{u}_k \leq \boldsymbol{u}_{\max}, \ k \in \{0,...,N\!-\!1\} \label{eq:input_constraint}
\end{align}
\end{subequations}
where $\boldsymbol{Q}$, $\boldsymbol{R}$, and $\boldsymbol{Q}_N$ are positive-definite matrices representing the state cost, control-energy cost, and terminal cost, respectively. The errors $\overline{\boldsymbol{x}}_k$ and $\overline{\boldsymbol{u}}_k$ are defined in Tab.~\ref{tab:error_term} in the previous subsection. 
Constraint (\ref{eq:dyn_constraint}) enforces the system dynamics, where $\boldsymbol{f}(\cdot)$ denotes the full tiltable-multirotor model established from (\ref{eq:servo_model}) to (\ref{eq:rigid_body}). The model is discretized by the \textit{fourth-order Runge-Kutta} method with time step $t_{\rm step}$.
Constraint (\ref{eq:init_constraint}) imposes the initial condition, which is essential for feedback in NMPC; $\hat{\boldsymbol{x}}$ denotes the estimated state from the estimator.
Finally, constraints (\ref{eq:state_constraint}) and (\ref{eq:input_constraint}) define the state and input limits, respectively, which are primarily determined by safety considerations and physical boundaries.

For the estimated wrench contained in $\hat{\boldsymbol{x}}$, we introduce a sigmoid function to suppress its influence at small values:
\begin{subequations}  \label{eq:threshold_ext_wrench}
\begin{align}
    {^W\hat{\boldsymbol{f}}_{de,0}} &= {^W\hat{\boldsymbol{f}}_{de}} \cdot \frac{1}{1+e^{-\lambda_{f} \ \left( \left\Vert {^W\hat{\boldsymbol{f}}_{de}} \right\Vert - f_{de,\rm th} \right)}}, \\
    {^B\hat{\boldsymbol{\tau}}_{de,0}} &= {^B\hat{\boldsymbol{\tau}}_{de}} \cdot \frac{1}{1+e^{-\lambda_{\tau} \ \left( \left\Vert {^B\hat{\boldsymbol{\tau}}_{de}} \right\Vert - \tau_{de,\rm th} \right)}}.
\end{align}
\end{subequations}

After optimization, the first element of the sequence $\boldsymbol{u}^*$ is sent to the robot for execution as $\boldsymbol{f}_c,\boldsymbol{\alpha}_c \leftarrow \boldsymbol{u}^*_0$.

In implementation, we use the software package \textit{CasADi} \cite{andersson_casadi_2019} to model the system and \textit{acados} \cite{verschueren_acadosmodular_2022} as the NMPC solver. The solver incorporates several standard techniques to accelerate computation, including \textit{warm starting}, \textit{real-time iteration (RTI)}, and \textit{multiple shooting}. \textit{Warm starting} accelerates convergence by providing an initial guess close to the final solution, taken from the result of the previous iteration. RTI computes the sequential quadratic programming (SQP) step only once per control iteration, prioritizing speed over exact optimality. \textit{Multiple shooting} divides the original OCP into smaller subproblems that can be solved in parallel. We refer interested readers to \cite{gros_linear_2020} for further details. The complete framework proposed in this article is summarized in Alg.~\ref{alg:all_framework}.

\begin{algorithm}[t]
\caption{Offset-Free Effector-Centric NMPC}\label{alg:all_framework}
\begin{algorithmic}[1]
\Require ${^W\boldsymbol{p}_{T_o,r}}, {^W_T\boldsymbol{q}_{r}}$, references for the effector pose and its derivatives; $\boldsymbol{f}(\cdot)$, the model defined in (\ref{eq:servo_model})--(\ref{eq:rigid_body}); $t_s$, the NMPC control period
\Ensure $\boldsymbol{f}_c,\boldsymbol{\alpha}_c$, the control commands
\State Initialize NMPC and servo angles with $\boldsymbol{x}_{r,k}, \boldsymbol{u}_{r,k}$;
\While{control is triggered every $t_s$}
    \State Update $\boldsymbol{I}, {^B\boldsymbol{p}_{E_i}}$ via URDF; 
    \Comment{\textit{Optional}}
    \State Update the state machine in Sec. \ref{sec:combinition};
    \If{CALIB is finished} \Comment{\textit{Event trigger}}
        \State Store ${^W\hat{\boldsymbol{f}}_{de,b}}, {^B\hat{\boldsymbol{\tau}}_{de,b}}$ as bias for (\ref{eq:wrench_est});
    \EndIf
    \If{${^W\boldsymbol{p}_{T_o,r}}, {^W_T\boldsymbol{q}_{r}}$ are updated}  \Comment{\textit{Event trigger}}
        \State $\boldsymbol{f}_{c,r,k}, \boldsymbol{\alpha}_{c,r,k} \gets$ initial allocation via (\ref{eq:pinv});
        \State $\boldsymbol{\alpha}_{c,r,k} \gets$ refinement via (\ref{eq:heuristic});
        \State Construct $\boldsymbol{x}_{r,k}, \boldsymbol{u}_{r,k}$;
    \EndIf
    \If{$\hat{\boldsymbol{\Omega}}, \hat{\boldsymbol{\alpha}}, {^B\hat{\boldsymbol{a}}}_{sf}, {^B\hat{\boldsymbol{\omega}}}$ are updated} \Comment{\textit{Event trigger}}
        \State Differentiate ${^B\hat{\boldsymbol{\omega}}}$ and low-pass filter all data;
        \State ${^W\hat{\boldsymbol{f}}_{de}}, {^B\hat{\boldsymbol{\tau}}_{de}} \gets$ estimate external wrench via (\ref{eq:wrench_est});
    \EndIf
        \State ${^W{\boldsymbol{f}}_{dm}}, {^B{\boldsymbol{\tau}}_{dm}} \gets$ handle model error via (\ref{eq:slerp})--(\ref{eq:digit_i_term});
        \State $\hat{\boldsymbol{x}} \gets$ state estimation;
        \State Modify ${^W\hat{\boldsymbol{f}}_{de}}, {^B\hat{\boldsymbol{\tau}}_{de}}$ via (\ref{eq:threshold_ext_wrench});
        \State $\boldsymbol{x}^*_k, \boldsymbol{u}^*_k \gets$ solve (\ref{eq:nmpc_opt_problem});
        \State $\boldsymbol{f}_c, \boldsymbol{\alpha}_c \gets \boldsymbol{u}^*_0$. 
    
\EndWhile
\end{algorithmic}
\end{algorithm}

\subsection{Feasibility and Trade-Offs}

Real-time feasibility is achieved using the RTI-based NMPC implementation described above. Although the EE-centric formulation increases the average NMPC solve time from \SI{3.74}{ms} to \SI{4.98}{ms}, the complete control loop remains below \SI{10}{ms}, enabling onboard execution at \SI{100}{Hz}.
A rigorous recursive-feasibility proof is difficult for the proposed NMPC because it would require terminal-set design and additional assumptions for the strongly nonlinear dynamics in (\ref{eq:dyn_constraint}). Therefore, as in related NMPC studies \cite{sun_comparative_2022,alharbat_predictive_2025}, feasibility is mainly evaluated experimentally. In practice, the solver is initialized from the known hover solution, and the references are kept within admissible ranges and close to the current state.

\setlength{\textfloatsep}{8pt plus 1.0pt minus 2.0pt}
\begin{figure}[t]
    \centerline{\includegraphics[trim=0 0 0 0,clip,width=3.5in]{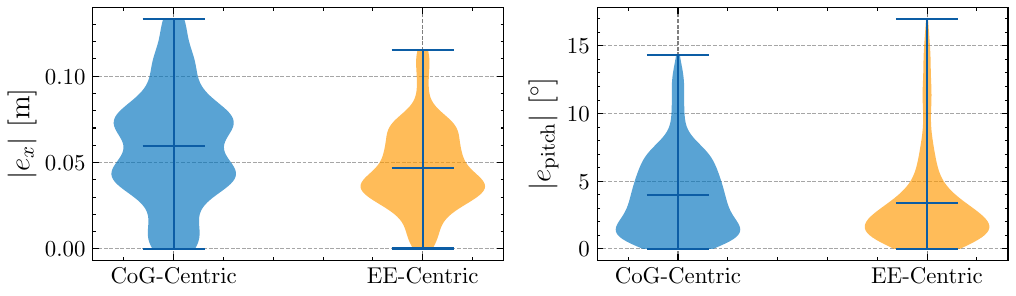}}
    \vspace*{-3mm}
    \caption{
        \revised{Experimental comparison between CoG-centric and EE-centric formulations during a \SI{0}{\degree}--\SI{180}{\degree} pitch rotation. In this representative tracking task, the EE-centric formulation shows lower RMSE in the X direction (\SI{0.053}{m} vs. \SI{0.068}{m}) and pitch angle (\SI{4.837}{\degree} vs. \SI{5.039}{\degree}).}
        }
    \label{fig:pitch_track_comparison}
\end{figure}

\revised{The EE-centric formulation also introduces trade-offs. Compared with a CoG-centric one, it requires additional computation for the transformations in (\ref{eq:ee-centric}), but it specifies the tracking objective directly in the EE frame and simplifies the planning part.
In addition, when the EE is horizontally pushed by about \SI{0.3}{m} during hovering without wrench compensation, the EE-centric formulation shows 54.8\% less pitch rotation than the CoG-centric formulation, suggesting its greater emphasis on EE tracking under disturbances.
We further compare it with a CoG-centric baseline in a representative \SI{0}{\degree}--\SI{180}{\degree} pitch-rotation task, where the EE reference is converted into an equivalent CoG reference for the baseline. As shown in Fig.~\ref{fig:pitch_track_comparison}, the EE-centric formulation gives a more concentrated error distribution and modestly lower RMSE values in the X direction and pitch angle. Overall, the EE-centric formulation is mainly valuable at the task-representation level, with a modest performance improvement.}

\section{\revised{Platform Description}} \label{sec:robot}

\subsection{Robot Platform}

\begin{figure}[t]
    \centerline{\includegraphics[trim=8cm 3cm 8cm 3cm,clip,width=3.5in]{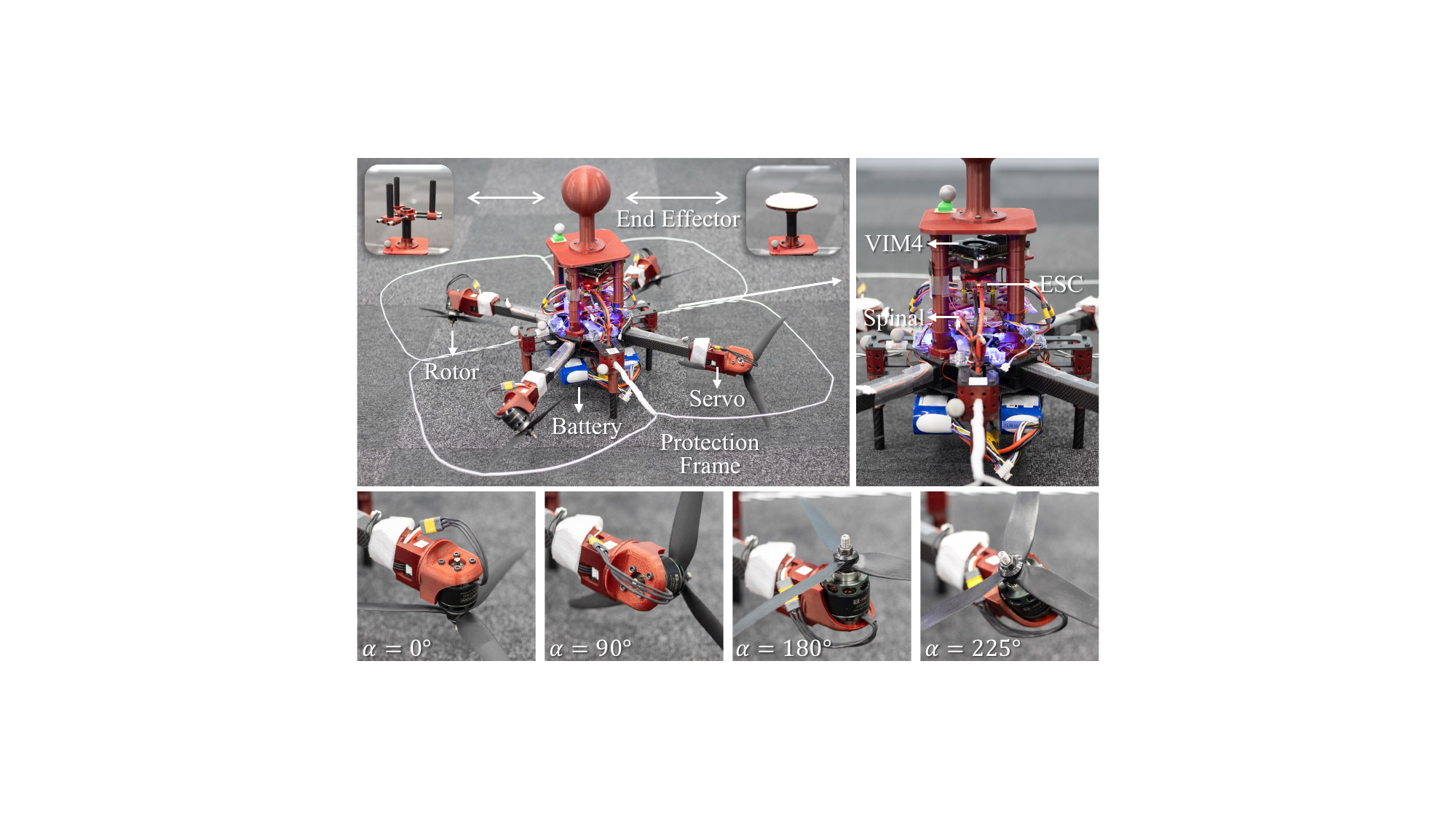}}
    \vspace*{-3mm}
    \caption{Snapshot of the self-designed omnidirectional tiltable-quadrotor, 
    \textit{Beetle-Art-Omni}, 
    together with images of its tilting module at various angles.}
    \label{fig:robot_detail}
\end{figure}

We designed the tiltable-quadrotor
shown in Fig.~\ref{fig:robot_detail} to evaluate the proposed framework, with its dimensions listed in Tab.~\ref{tab:mdl_ctrl_params}.
The design is both assembly- and repair-oriented, enabling quick recovery from crashes.
The system adopts a ``brain-spinal'' structure: a Khadas VIM4 computer with \SI{2.2}{\giga\hertz} quad-core ARM Cortex-A73 and \SI{2.0}{\giga\hertz} quad-core Cortex-A53 CPUs serves as the onboard PC, while a custom ``spinal'' module with an STM32H7-series processor functions as the low-level MCU.
For actuators, a T-Motor F55A PROII 6S 4IN1 ESC powers four T-Motor AT2814 KV900 motors with three-blade 9045 propellers, and XC330-T181-T servos provide a tilting range of more than \SI{180}{\degree}. Each servo is mounted at the arm tip to reduce mechanical complexity. The motors are installed facing downward, increasing lift by about 15\% compared to upward installation.
Two 6S LiPo \SI{3000}{\milli\ampere\hour} batteries supply power, supporting approximately \SI{7}{\minute} of horizontal hovering. Reflective markers are distributed around the robot for motion capture, and an iron frame provides structural protection. Three interchangeable end-effectors are designed.

\setlength{\textfloatsep}{8pt plus 1.0pt minus 2.0pt}

\begin{figure}[t]
    \centerline{\includegraphics[trim=7.8cm 3.5cm 7.8cm 3.5cm,clip,width=3.4in]{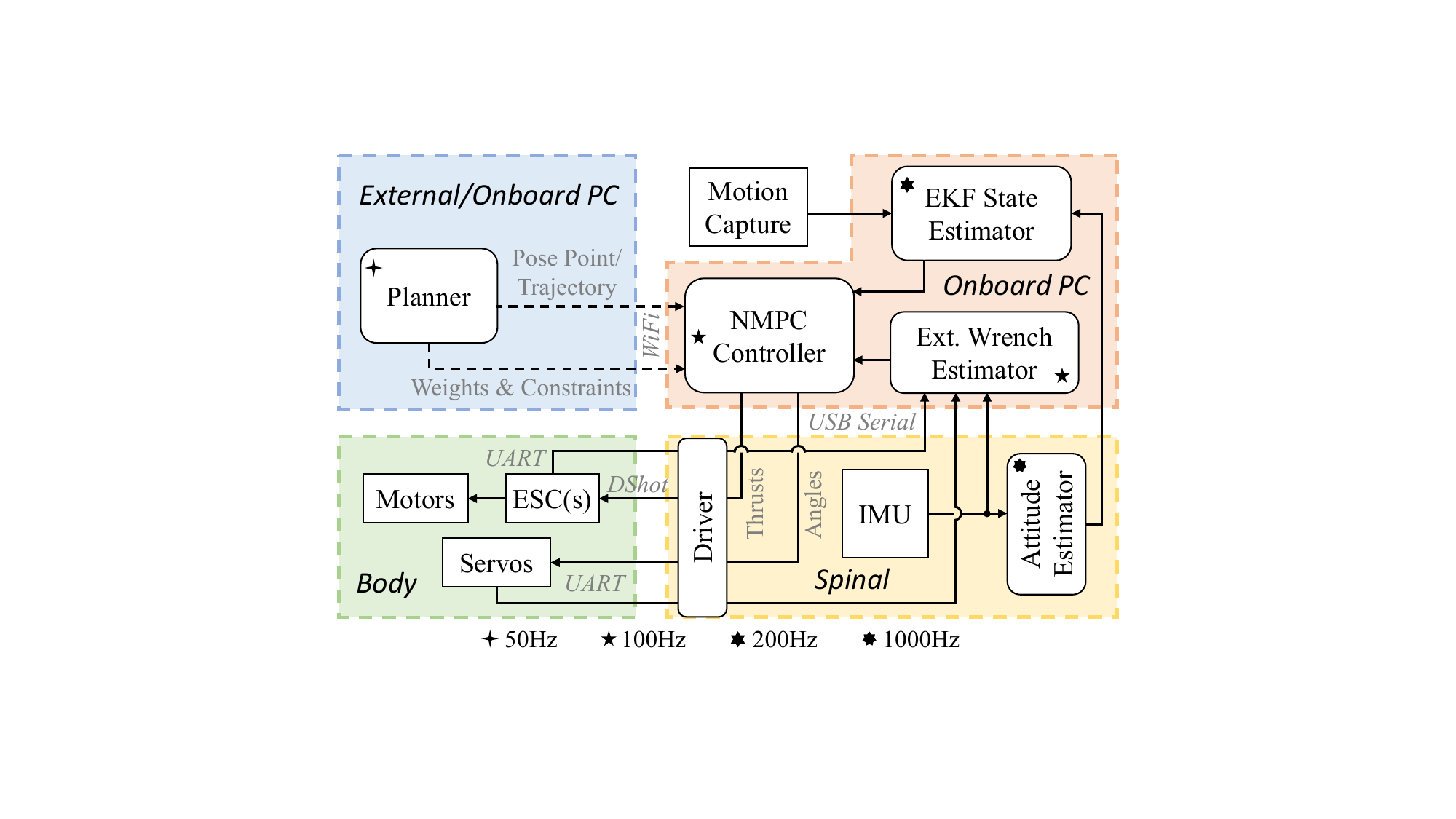}}
    \vspace*{-3mm}
    \caption{Information flow of our system, where normal rectangles represent hardware and rounded rectangles represent software.}
    \label{fig:workflow}
\end{figure}

The software information flow of our system is illustrated in Fig.~\ref{fig:workflow}. On the onboard computer, the Robot Operating System (ROS-O) serves as the core scheduler, and \textit{acados} is used for the proposed NMPC framework. We employ \textit{Partial Condensing HPIPM} as the QP solver, \textit{Explicit Runge-Kutta} as the integrator, and the parameters in Tab.~\ref{tab:mdl_ctrl_params} as control settings.
In each control cycle, the control command is first transmitted to the ``spinal'' module via USB and then forwarded to the actuators. The DShot protocol is used to send throttle signals and receive rotor speed measurements $\hat{\boldsymbol{\Omega}}$, and half-duplex UART is used to communicate with the servos for $\hat{\boldsymbol{\alpha}}$.
Regarding state estimation, to avoid the singularity issues associated with Euler angles, we use the rotation matrix ${^W_B\boldsymbol{R}^\top}=[{^B\boldsymbol{e}_x}, {^B\boldsymbol{e}_y}, {^B\boldsymbol{e}_z}]$ as the attitude representation, where each column denotes an axis of $\{\mathcal{W}\}$ expressed in $\{\mathcal{B}\}$. In particular, ${^B\boldsymbol{e}_z}$ is primarily determined by gravity and is therefore first estimated on the ``spinal'' module using a \SI{1}{kHz} complementary filter based on IMU measurements $\hat{\boldsymbol{a}}_{sf}$ and $\hat{\boldsymbol{\omega}}$. Since a pure IMU cannot provide yaw information, the estimated ${^B\boldsymbol{e}_z}$ is then transmitted to the onboard PC and fused with the full rotation matrix obtained from the MoCap system. A \SI{200}{Hz} EKF is designed for this fusion process.
More implementation details can be found in \cite{zhao_versatile_2021}.

\setlength{\textfloatsep}{8pt plus 1.0pt minus 2.0pt}

\subsection{Parameter Identification and Filter Design}  \label{sec:sysid_filter}

Identification is required to obtain the robot's physical parameters for both simulation and control. Specifically, geometric and inertial parameters are extracted from CAD software.
Rotor parameters and the throttle--thrust mapping are identified using a Leptrino 6-axis force/torque sensor (PFS055YA501U6).
For the servos, we first tune the PID gains of the inner controller to achieve fast response, guided primarily by the Ziegler-Nichols method \cite{astrom_revisiting_2004}. A preliminary time constant $t_{\rm servo}$ is predefined for the first hover test. After collecting flight data, $t_{\rm servo}$ is identified using the MATLAB SYSID Toolbox. All identified parameters are listed in Tab.~\ref{tab:mdl_ctrl_params}.


Finally, we introduce the practical filter-design procedure for the acceleration-based wrench estimator in Sec.~\ref{sec:acc_wrench_est}. We use a second-order Butterworth filter as the LPF, whose group delay can be approximated as $t_{BW}=1/\left(\pi \ f'_c\right)$ when $f'_c \ll f'_s$, where $f'_c$ and $f'_s$ denote cutoff and sampling frequencies, respectively.
For differentiation, we employ a five-point derivative $\frac{1}{12}\{-1, 8, 0, -8, 1\}$ (with the first element corresponding to the newest data) to compute ${^B\hat{\dot{\boldsymbol{\omega}}}}$, which introduces a group delay of $2 \cdot t_{s,\rm IMU}=\SI{10}{\milli\second}$.
If an LPF with a \SI{15}{\hertz} cutoff frequency is applied to the differentiator, the LPF adds a group delay of \SI{21}{\milli\second}, resulting in a total delay of \SI{31}{\milli\second}. Consequently, the LPFs for the other measurements use a cutoff frequency of \SI{10.27}{\hertz} to introduce the same delay.
To further improve flight stability, we finally apply an additional \SI{2}{\hertz} filter to the estimated results.



\begin{table}[t]
\setlength{\abovecaptionskip}{0pt} 
\setlength{\belowcaptionskip}{-1pt}
\centering
\caption{Model \& Control Parameters}
\begin{tabular*}{\linewidth}{@{\extracolsep{\fill}} l c|l c|l c }
 \toprule
 \textbf{Param.} & \textbf{Value} & \textbf{Param.} & \textbf{Value} & \textbf{Param.} & \textbf{Value} \\
 \midrule
 \multicolumn{6}{c}{\textit{Model, Sec.~\ref{sec:modeling} \& Design, Sec.~\ref{sec:design} \& Reference, Sec.~\ref{sec:ref}}}  \\
 \midrule
  {$l$} & \SI{0.275}{\meter} & $R_p$ & \SI{0.114}{\meter} & {$m$} & \SI{3.039}{kg}  \\
 {$I_{xx}$} & $0.063$ & {$I_{yy}$} & {$0.062$} & {$I_{zz}$} & {$0.095$} \\
$N_p$ & $4$ & {$\alpha_{\rm limit}$} & \SI{\pm225}{\degree} & $t_{\rm servo}$ & 
\SI{0.048}{\second}  \\
$k_q/k_t$ & \SI{0.0165}{m} & $f_{t,\epsilon}$ & \SI{1.5}{\newton} & $\lambda$ & \SI{90}{\degree} \\
 \midrule
  \multicolumn{6}{c}{\textit{Integral Term \& Wrench Estimator, Sec.~\ref{sec:error}}} \\
 \midrule
 $k_{I,xy}$ & $100$ & $k_{I,z}$ & $500$ & $f_{de, \rm th}$ & \SI{0.6}{\newton}\\
 $k_{I,qxy}$ & $10$ & $k_{I,qz}$ & $50$ & $\tau_{de, \rm th}$ & \SI{0.3}{\newton\meter} \\
 $\lambda_{f}$ & $10$ & $\lambda_{\tau}$ & $10$ & & \\
  $v_{\rm th,calib}$ & \SI{0.1}{\meter\per\second} & $\omega_{\rm th,calib}$ & \SI{0.1}{\radian\per\second} & $t_{\rm calib}$ & \SI{3}{\second} \\
 $v_{\rm th,stop}$ & \SI{0.5}{\meter\per\second} & $\omega_{\rm th,stop}$ & \SI{0.5}{\radian\per\second} & & \\
 \midrule
   \multicolumn{6}{c}{\textit{NMPC Controller, Sec.~\ref{sec:control}}} \\
 \midrule
 $N$ & $20$ & $t_{\rm step}$ & \SI{0.1}{\second} & $t_s$ & \SI{0.01}{\second} \\
 $Q_{p,xy}$ & $300$ & $Q_{p,z}$ & $400$ & $Q_{v,xy}$ & $10$ \\
 $Q_{v,z}$ & $10$ & $Q_{q,xy}$ & $300$ & $Q_{q,z}$ & $600$ \\
 $Q_{\omega, xy}$ & $5$ & $Q_{\omega,z}$ & $5$ & $Q_{\alpha}$ & $2$ \\ 
 $R_{f}$ & $2$ & $R_{\delta\alpha}$ & $250$ &  &  \\
 $v_{\rm limit}$ & \SI{\pm 1}{\meter\per\second} & $\omega_{\rm limit}$ & \SI{\pm 6}{\radian\per\second} & $\alpha_{i,{\rm limit}}$ & $\pm 3.15$ \\
 $f_{i,\min}$ & \SI{0}{\newton} & $f_{i,\max}$ & \SI{23}{\newton} & $\alpha_{ic,{\rm limit}}$ & $\pm 3.15$ \\
 \bottomrule
\end{tabular*}
\parbox{\linewidth}{\footnotesize
Note: The unit of $I_{xx}$, $I_{yy}$, and $I_{zz}$ is
\si{\kilogram\meter\squared}.}
\label{tab:mdl_ctrl_params}
\vspace{-3mm}
\end{table}


\section{Experiments} \label{sec:exp}


We conducted several experiments to evaluate the performance of the proposed NMPC framework.
First, we performed omnidirectional trajectory tracking to verify singularity handling and effector-centric behavior. Next, we conducted experiments to evaluate the wrench estimator, laying the foundation for disturbance compensation. Finally, supported by a teleoperation framework, we conducted two applications that combined the above capabilities: pushing a sliding whiteboard and continuously turning a vertically installed valve.

\subsection{Effector-Centric Omnidirectional Flight}

In this part, we evaluated the proposed framework for omnidirectional flight in an effector-centric manner. A ball-shaped end-effector was mounted on the robot, and the goal was to track trajectories around its center.
\revised{The wrench estimator was disabled because no intentional external wrench was applied,}
while the integral term was activated for all six axes to compensate for model error.

Visualizing the attitude during omnidirectional flight was nontrivial. The raw attitude states were represented as quaternions but then converted to Euler angles for readability. 
To avoid the jumps caused by singularities, we first computed the quaternion error, converted it into Euler angles, and then added it to the reference. Note that in the following figures, the abrupt changes in roll and yaw when pitch reaches $\pm90^\circ$ are caused by the singularity of Euler-angle notation.

\setlength{\dbltextfloatsep}{8pt plus 1.0pt minus 2.0pt}
\begin{figure}[t] 
    \centering
    \subfloat[Snapshots of \SI{360}{\degree} Pitch and Roll rotations about the end-effector]{
        \includegraphics[trim=0 0 0 0,clip,width=3.45in]{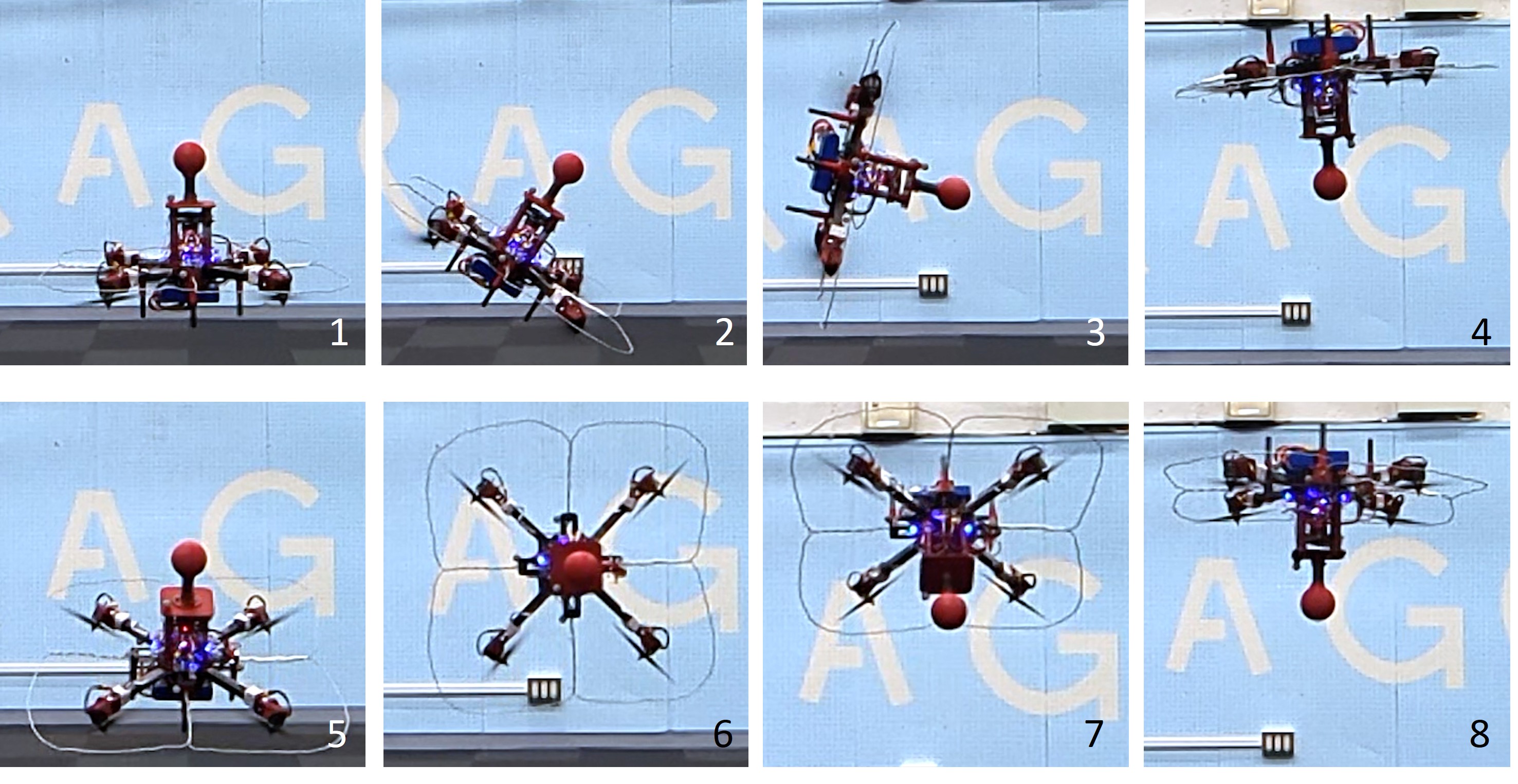}
        \label{fig:exp_pitch_roll_snapshots}
    } \\
    \subfloat[Data of \SI{360}{\degree} Pitch and Roll rotations about the end-effector]{
        \includegraphics[trim=1mm 0 0 0,clip,width=3.45in]{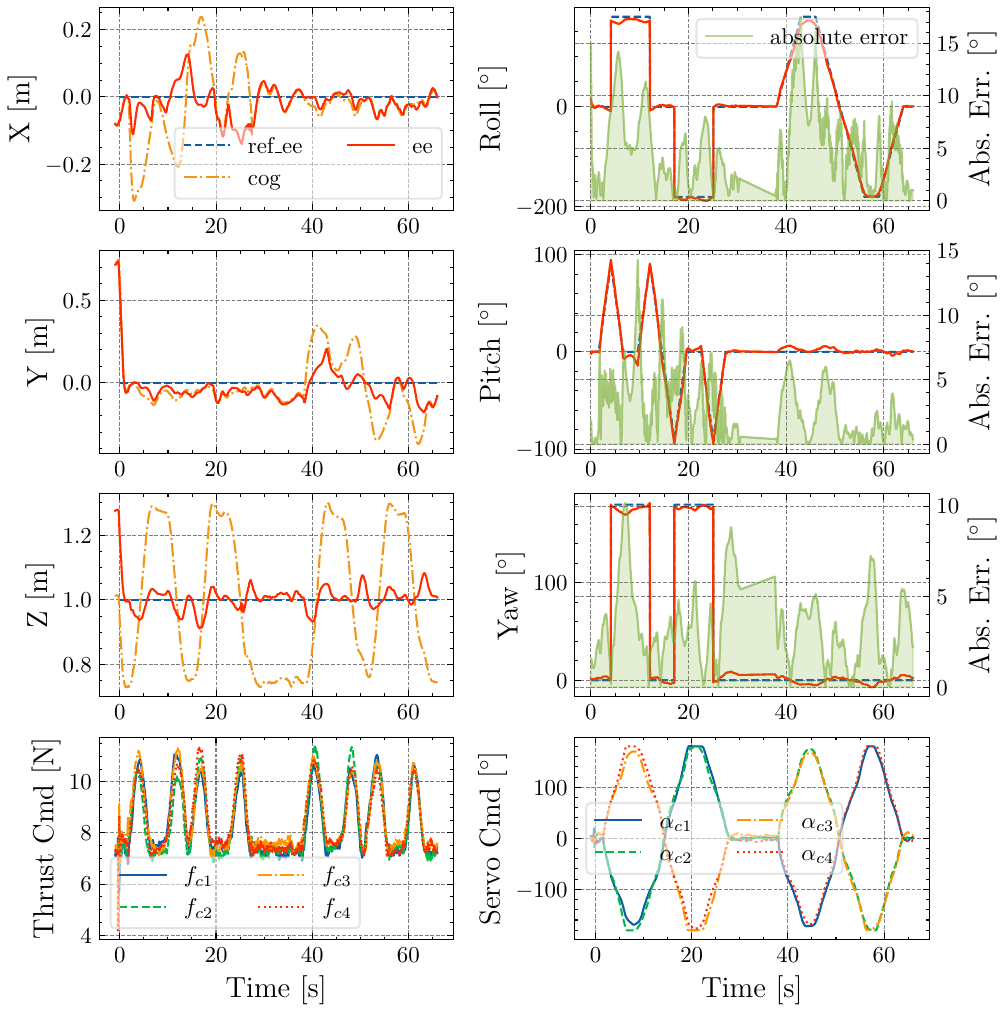}
        \label{fig:exp_pitch_roll_data}
    }\vspace{-1mm}
    \caption{Rotational flight about the pitch and roll axes. The RMSEs are: $p_x$: \SI{0.0478}{\meter}, $p_y$: \SI{0.1199}{\meter}, $p_z$: \SI{0.0467}{\meter}; roll: \SI{6.3287}{\degree}, pitch: \SI{3.7819}{\degree}, yaw: \SI{3.9590}{\degree}, where Euler angle errors are converted from quaternion errors for readability. The CoG curve denotes the change of the CoG point during this flight.
    Complete entry/exit data are retained to present the full process.}
    \label{fig:exp_pitch_roll_omni}
\end{figure}

\subsubsection{Pitch \& Roll \SI{360}{\degree} Rotation}

The first experiment evaluated trajectory tracking along the pitch and roll directions.
Due to the limited actuation range, we commanded the robot to move linearly from \SI{0}{\degree} to \SI{180}{\degree}, then back through a full circle covering $\left[-180^\circ, 180^\circ \right]$, and finally returned it to the starting pose. At both \SI{180}{\degree} and \SI{-180}{\degree}, the robot remained stationary for \SI{3}{\second}.
The duration of one cycle was set to \SI{10}{\second}. Snapshots of this experiment are shown in Fig.~\ref{fig:exp_pitch_roll_snapshots}, and the corresponding data are plotted in Fig.~\ref{fig:exp_pitch_roll_data}.

As shown in Fig.~\ref{fig:exp_pitch_roll_omni}, the robot successfully executes both motions,
demonstrating the reliability of the tilting structure.
The plots also reveal how the CoG is shifted to maintain the end-effector's pose. Since the end-effector is mounted vertically above the body, its orientation aligns with that of the CoG.
Regarding control performance, the rotational motion influences positional accuracy, with errors of less than \SI{0.1}{\meter} in X, \SI{0.2}{\meter} in Y, and \SI{0.05}{\meter} in Z. The X error is primarily affected by pitch rotation, and the Y error by roll rotation. This correlation between angular and positional errors likely results from model inaccuracies due to assembly imperfections and aerodynamic effects. Nevertheless, the controller tracks all angular changes successfully, providing an initial verification of omnidirectional capability.

\setlength{\dbltextfloatsep}{8pt plus 1.0pt minus 2.0pt}
\begin{figure}[t] 
    \centering
    \subfloat[Snapshots of continuous vertical cartwheel flight]{
        \includegraphics[trim=0 0 0 8mm,clip,width=3.45in]{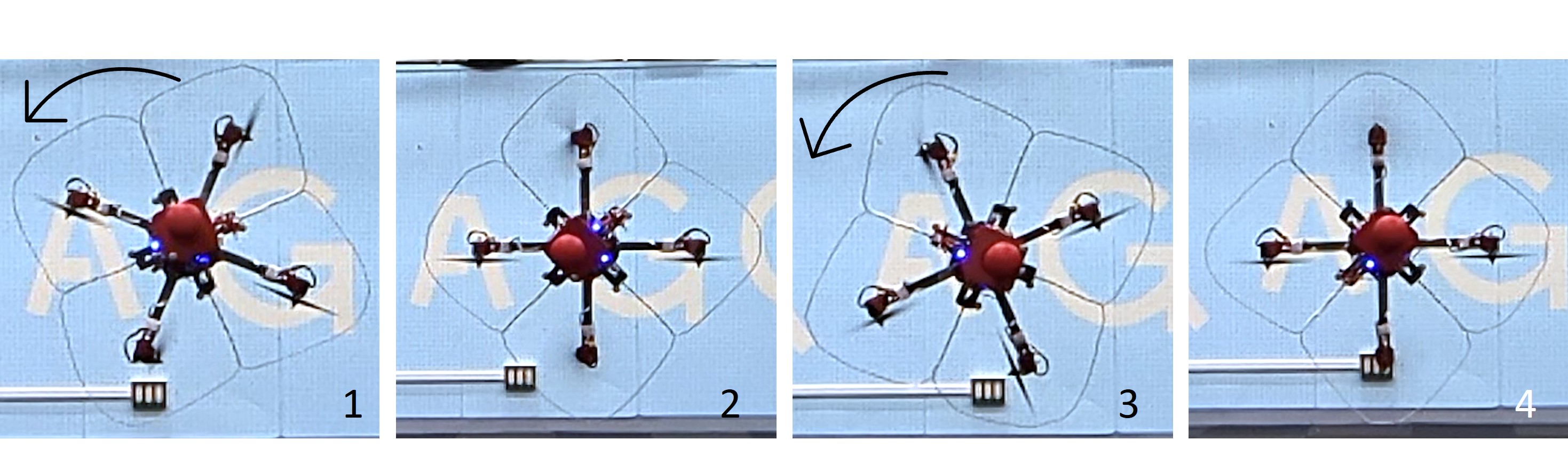}
        \label{fig:exp_omni_vertical_yaw}
    } \\
    \subfloat[Data of continuous vertical cartwheel flight about the end-effector]{
        \includegraphics[trim=0 0 0 0,clip,width=3.45in]{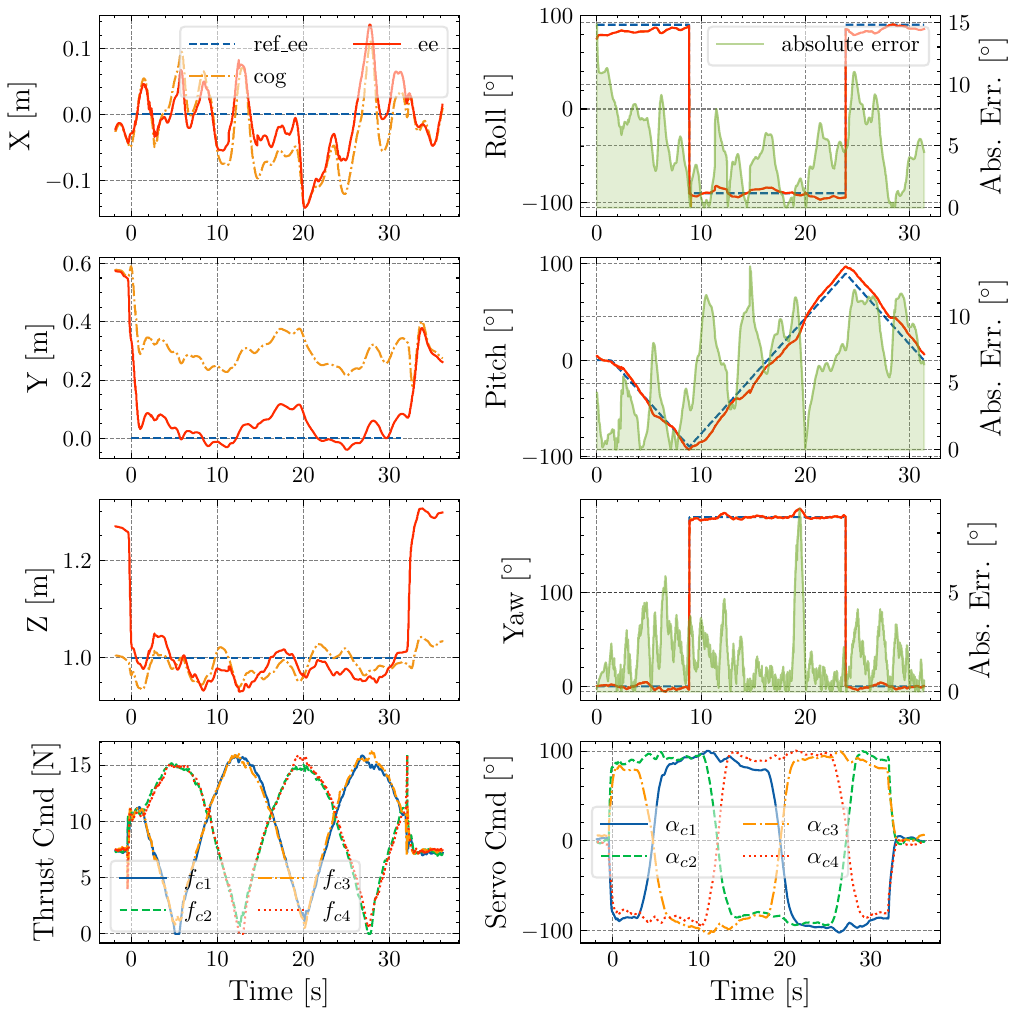}
        \label{fig:exp_vertical_yaw_data}
    }\vspace{-1mm}
    \caption{Continuous vertical cartwheel flight around the end-effector. The RMSEs are: $p_x$: \SI{0.0514}{\meter}, $p_y$: \SI{0.1616}{\meter}, $p_z$: \SI{0.1110}{\meter}; roll: \SI{5.4626}{\degree}, pitch: \SI{7.4434}{\degree}, yaw: \SI{2.3075}{\degree}. Complete entry/exit data are retained to present the full experimental process.}
    \label{fig:exp_vertical_yaw}
\end{figure}

\subsubsection{Vertical Cartwheel Rotation}

As discussed in Sec.~\ref{sec:design}, mounting the end-effector on top of the robot \revised{can provide large available torque}, and a typical application is to turn a vertically installed valve by continuous rotation.
Performing this motion required careful handling of several singular points and thus served as a meaningful test case. For this experiment, the robot was initialized at a roll angle of \SI{90}{\degree} and then commanded to perform a full circular rotation about the body Z-axis over \SI{30}{\second}. Similar to the previous experiments, the angular speed was constant. Snapshots and data from this flight are shown in Fig.~\ref{fig:exp_omni_vertical_yaw} and Fig.~\ref{fig:exp_vertical_yaw_data}, respectively.

From Fig.~\ref{fig:exp_vertical_yaw}, the robot successfully performs a continuous cartwheel rotation, bypassing all singular points. Despite the singularity-handling strategy in Sec.~\ref{sec:ref}, the rotation speed is reduced when a rotor is tilted to the opposite side. However, this imperfection has \revised{limited} impact on real-world torque-demanding tasks. Another notable observation is the large variation in thrust: since only two rotors compensate for gravity at singular poses, thrust values fluctuate from zero up to twice the hovering throttle---a condition harsher than the near-hovering regime of conventional multirotors. Finally, the robot consistently points the rotor backward when passing singular points to avoid cable winding, thereby validating the method in Sec.~\ref{sec:ref}.

\setlength{\dbltextfloatsep}{8pt plus 1.0pt minus 2.0pt}
\begin{figure}[t] 
    \centering
    \subfloat[Snapshots of discrete vertical cartwheel flight to test singularity points]{
        \includegraphics[trim=0 0 0 8mm,clip,width=3.45in]{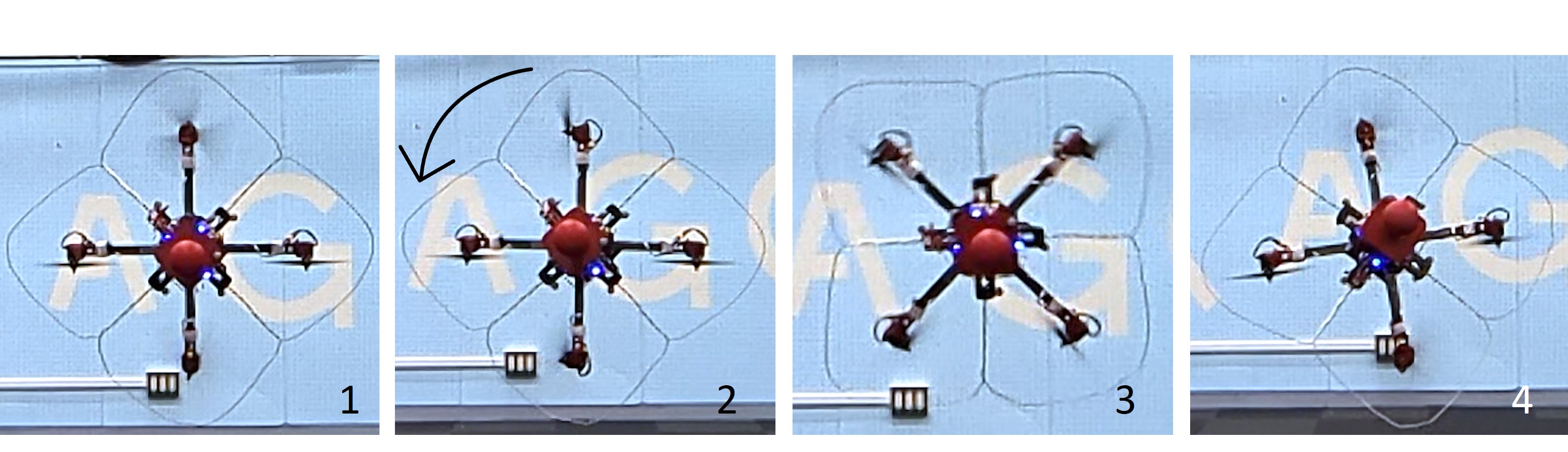}
        \label{fig:exp_omni_singularity_pts}
    } \\
    \subfloat[Data of discrete vertical cartwheel flight to test singularity points]{
        \includegraphics[trim=0 0 0 0,clip,width=3.45in]{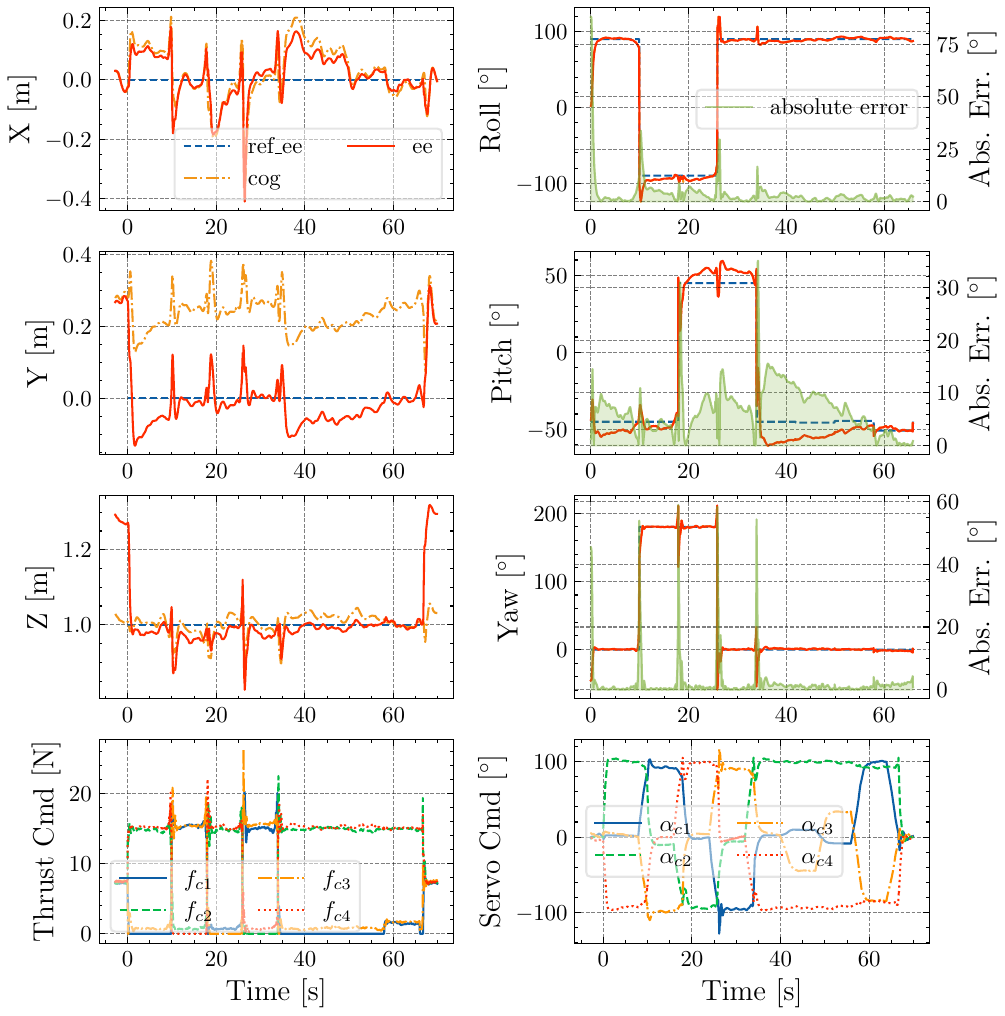}
        \label{fig:exp_singularity_data}
    }\vspace{-1mm}
    \caption{Discrete vertical cartwheel flight to evaluate singularity points. The RMSEs are: $p_x$: \SI{0.0784}{\meter}, $p_y$:  \SI{0.0875}{\meter}, $p_z$: \SI{0.0867}{\meter}; roll: \SI{7.7260}{\degree}, pitch: \SI{8.1508}{\degree}, yaw: \SI{6.5117}{\degree}. \revised{The transient peaks in the attitude error are caused by the step changes of the body-Z angle reference.}}
    \label{fig:exp_singular_pts}
\end{figure}

\subsubsection{Small Angles around Singular Poses}

Our last experiment challenged the limits of the proposed controller by introducing step changes of \SI{90}{\degree} in vertical flight. In addition, it aimed to thoroughly evaluate performance near singular points.
As shown in Fig.~\ref{fig:exp_singular_pts}, the reference signal comprises two parts. In the first part (\SI{8}-\SI{40}{\second}), \revised{after the robot is initialized at a \SI{90}{\degree} roll angle,} the \revised{body-Z angle reference} starts at ${\pi}/{4}$ and switches sequentially among ${{3\pi}/{4}},{5\pi}/{4},{7\pi}/{4}$, and back to ${\pi}/{4}$, with each step lasting \SI{8}{\second}.
In the second part (\SI{40}-\SI{65}{\second}), the \revised{body-Z angle reference} is set sequentially to ${{\pi}/{4},{\pi}/{4}+0.01,{\pi}/{4}-0.01}$, and ${\pi}/{4}+0.1$, again with \SI{8}{\second} per step. Flight snapshots and data are presented in Fig.~\ref{fig:exp_omni_singularity_pts} and Fig.~\ref{fig:exp_singularity_data}, respectively.

From Fig.~\ref{fig:exp_singular_pts}, the robot swiftly rotates to each new singularity pose following the step commands. Although overshoot is observed, the system shows a rapid transient response and settles within two oscillations.
Regarding accuracy, all three axes converge to near-zero offset within each \SI{8}{\second} interval owing to the integral term. However, this compensation at one singular pose also introduces larger deviations when transitioning to the next singular pose. A possible solution is to identify the model error with an integral term in all directions and use it as a feedforward term. For most slow-motion aerial manipulation tasks, however, the proposed NMPC framework with the integral term already provides sufficient accuracy.

The second half of Fig.~\ref{fig:exp_singular_pts} illustrates the behavior near singular points. As shown in Fig.~\ref{fig:exp_singularity_data}, the servo angle is zero when tilted exactly upward, and is positively correlated with the robot’s yaw rotation within a small range (\SI{43}-\SI{58}{\second}). When the yaw angle exceeds this range, the servo angle switches back to \SI{180}{\degree} or \SI{-180}{\degree} (\SI{58}-\SI{65}{\second}), precisely matching the design in Sec.~\ref{sec:ref}. Overall, these experiments demonstrate singularity-included omnidirectional flight.

\subsection{External Wrench Estimation and Compensation}


\subsubsection{Open-Loop Wrench Estimation}
\label{sec:exp:open-loop_wrench_estimation}

\begin{figure}[t]
    \centerline{\includegraphics[trim=0 0 0 0,clip,width=3.45in]{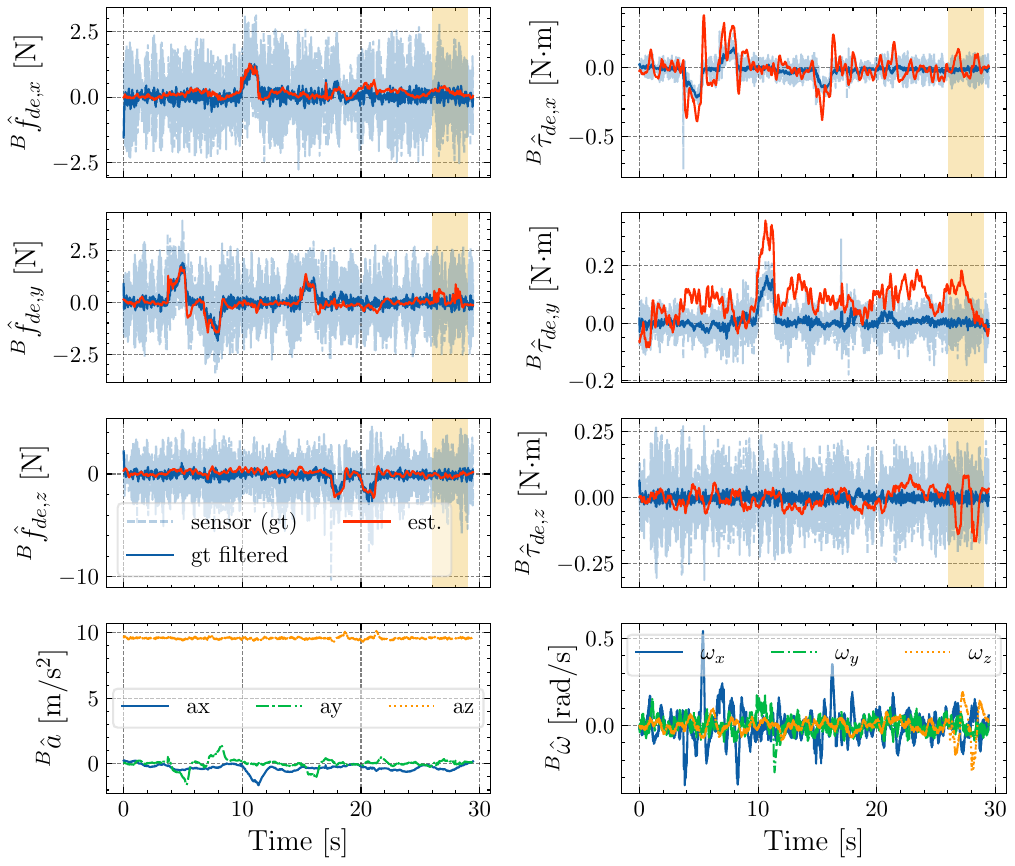}}
    \vspace*{-3mm}
    \caption{\revised{Open-loop wrench-estimation data. The offsets of the wrench sensor and the estimator are removed using the average over the first \SI{3}{\second}. Compared with the filtered ground truth (gt) over the initial \SI{25}{\second}, the force RMSE is X: \SI{0.234}{\newton}, Y: \SI{0.264}{\newton}, Z: \SI{0.424}{\newton}; the torque RMSE is X: \SI{0.087}{\newton\meter}, Y: \SI{0.090}{\newton\meter}, Z: \SI{0.034}{\newton\meter}.
    }
    The yellow region denotes the period when the protection frame, rather than the end-effector, is poked, highlighting the estimator’s ability to sense external wrenches applied away from the EE.}
    \label{fig:exp_wrench_est_data}
\end{figure}

In this part, we evaluate the accuracy of the acceleration-based wrench estimator. A 6-axis wrench sensor is mounted between the robot base and the end-effector to obtain ground-truth measurements. During horizontal hovering, an operator pokes the robot with an iron rod to generate external wrenches. Both the wrench estimator and the sensor record the data as visualized in Fig.~\ref{fig:exp_wrench_est_data}.

\setlength{\textfloatsep}{8pt plus 1.0pt minus 2.0pt}

From Fig.~\ref{fig:exp_wrench_est_data}, the vibrations induced by the rotors make the raw sensor data extremely noisy. Therefore, we apply an $N=10$ moving average filter to smooth them.
\revised{
Overall, the estimator is able to capture the wrench trend, with the force estimate showing smaller error than the torque estimate.
This difference originates from the sensing paths: force is obtained directly from the accelerometer, while torque is obtained indirectly from the gyroscope, where numerical differentiation amplifies gyro noise.}
The estimation delay appears to be negligible.

{The yellow region (\SI{26}-\SI{29}{\second}, Fig.~\ref{fig:exp_wrench_est_data}) illustrates that when the protection frame is poked instead of the end-effector, the estimator detects this wrench but the end-effector sensor does not. Thus, the estimator can capture robot-level external wrenches such as collisions or wind gusts, but cannot localize the contact. Therefore, in some windy manipulation scenarios, an end-effector wrench sensor may still be useful.}

\subsubsection{Closed-Loop Wrench Compensation}

\setlength{\textfloatsep}{8pt plus 1.0pt minus 2.0pt}
\begin{figure}[t]
    \centerline{\includegraphics[trim=0 0 0.2cm 0,clip,width=3.45in]{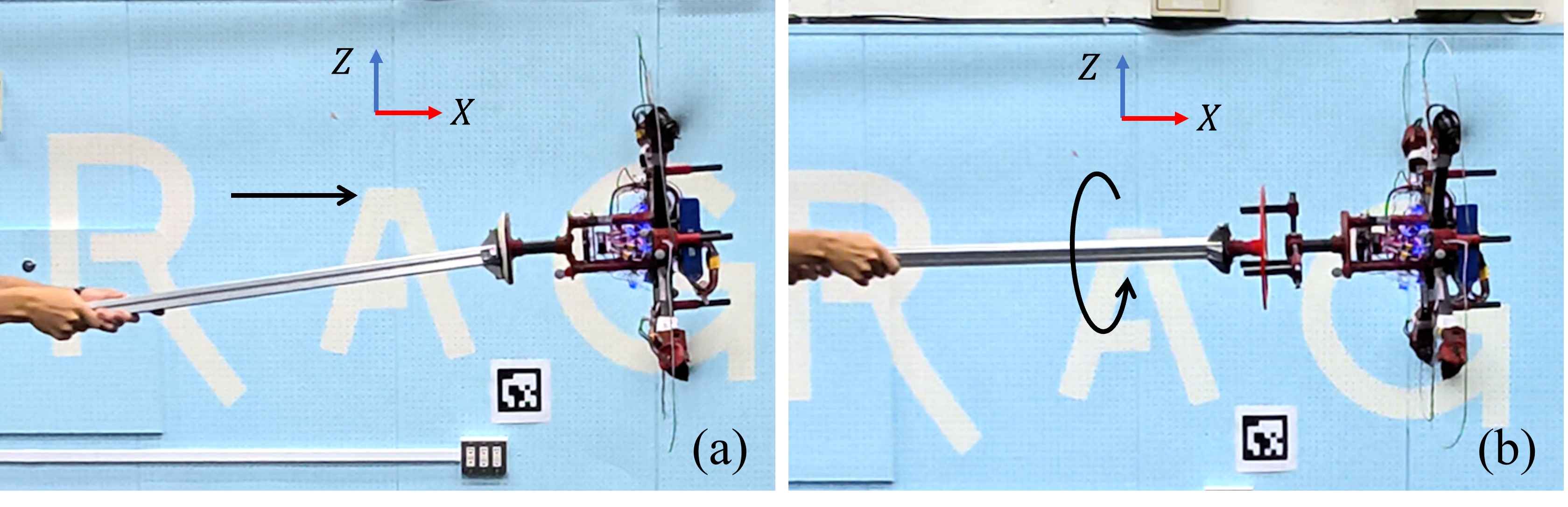}}
    \vspace*{-3mm}
    \caption{Snapshots of closed-loop external wrench compensation tests: (a) force and (b) torque.}
    \label{fig:exp_wrench_est_close_loop_photo}
\end{figure}

\setlength{\textfloatsep}{8pt plus 1.0pt minus 2.0pt}
\begin{figure}[t]
    \centering
    \subfloat[Lateral force without compensation]{
        \includegraphics[trim=0 0 0 0,clip,width=3.45in]{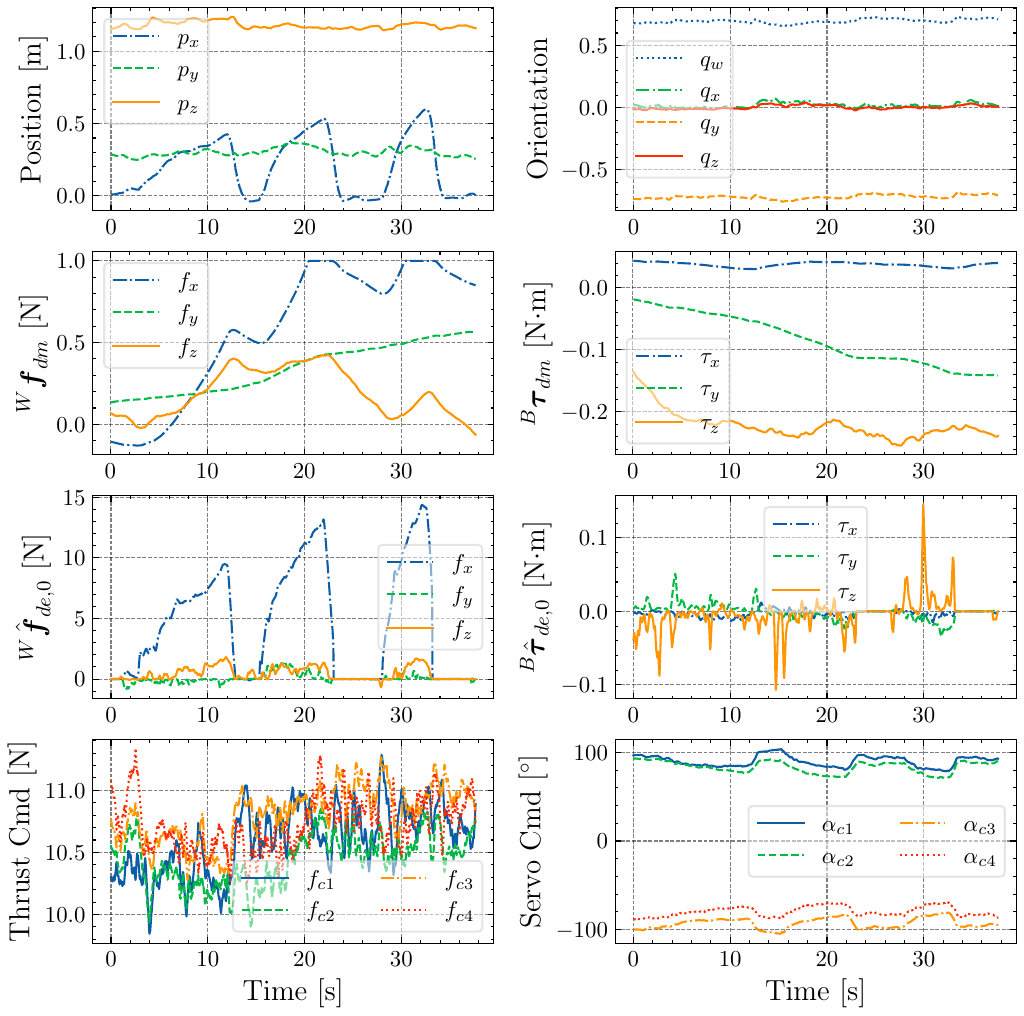}
        \label{fig:exp_vertical_force_wo_comp}
    } \\
    \subfloat[Lateral force with compensation]{
        \includegraphics[trim=0 0 0 0,clip,width=3.45in]{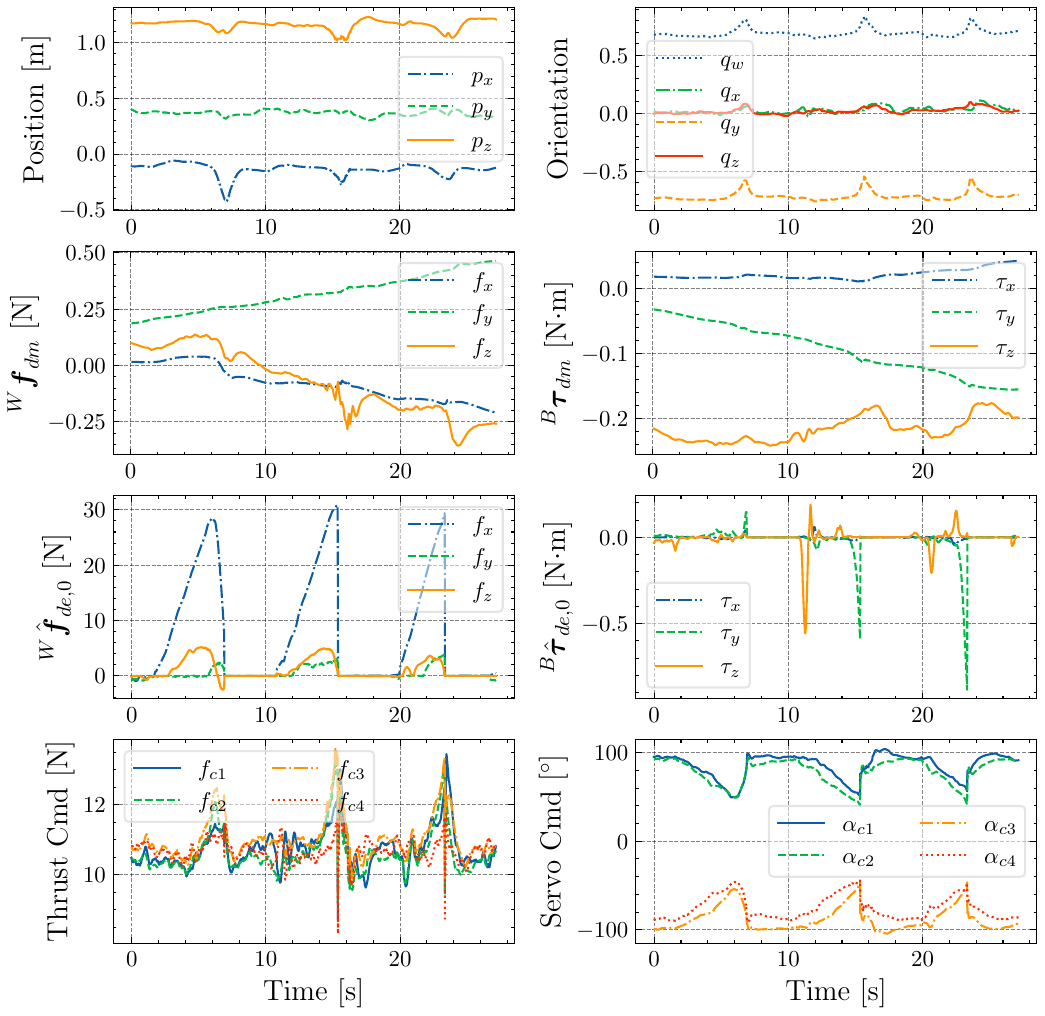}
        \label{fig:exp_vertical_force_w_comp}
    }\vspace{-1mm}
    \caption{Flight data without and with wrench compensation corresponding to Fig.~\ref{fig:exp_wrench_est_close_loop_photo}a. Without compensation, the robot is easily pushed far away, and the tracking controller generates a maximum force of about \SI{14}{\newton}. With compensation, the robot remains near the starting point and produces up to \SI{30}{\newton}. Some integral terms (e.g., ${^Wf_{dm,y}}$ and ${^B\tau_{dm,y}}$, (\ref{eq:model_error})) continue to increase, since model error is not fully compensated.
    }
    \label{fig:exp_vertical_force}
\end{figure}

\setlength{\textfloatsep}{8pt plus 1.0pt minus 2.0pt}
\begin{figure}[t]
    \centering
    \subfloat[Lateral torque without compensation]{
        \includegraphics[trim=0 0 0 0,clip,width=3.45in]{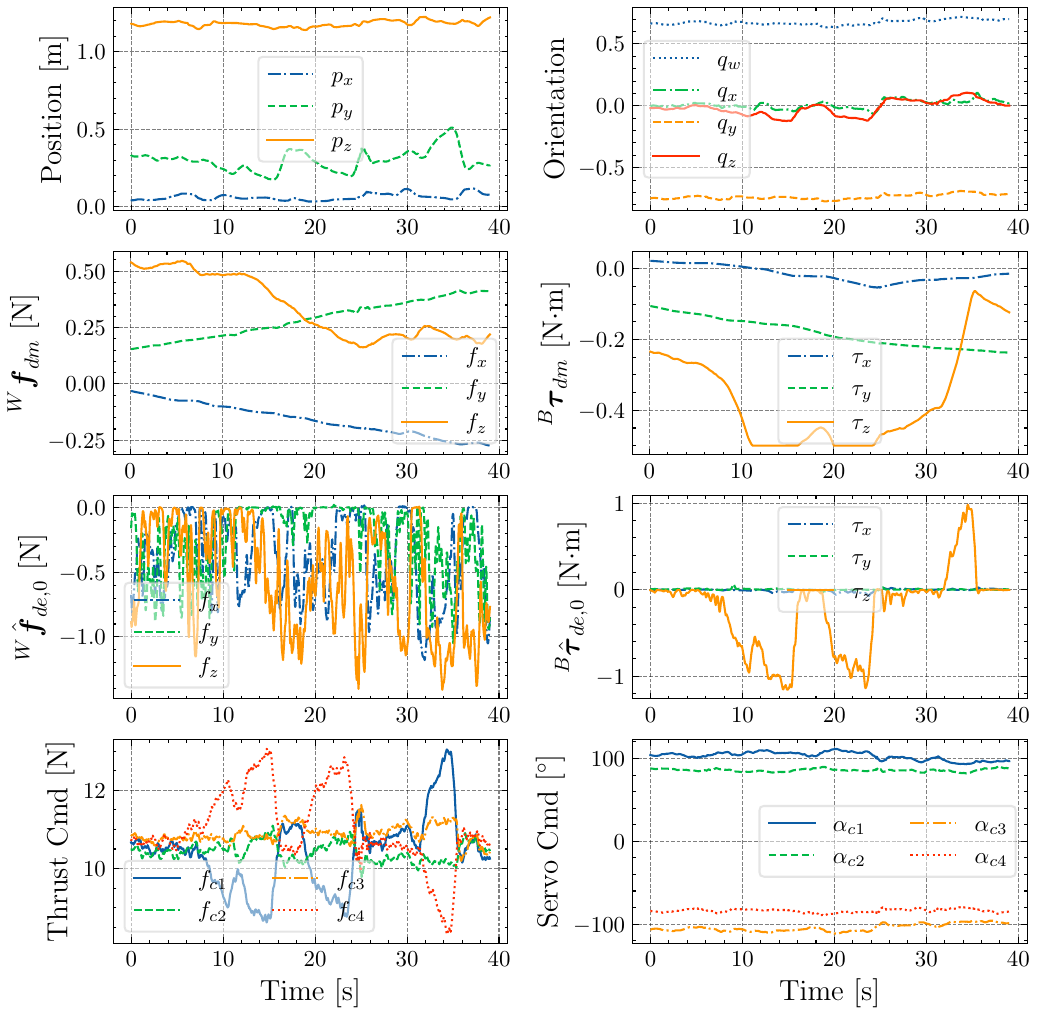}
        \label{fig:exp_vertical_torque_wo_comp}
    } \\
    \subfloat[Lateral torque with compensation]{
        \includegraphics[trim=0 0 0 0,clip,width=3.45in]{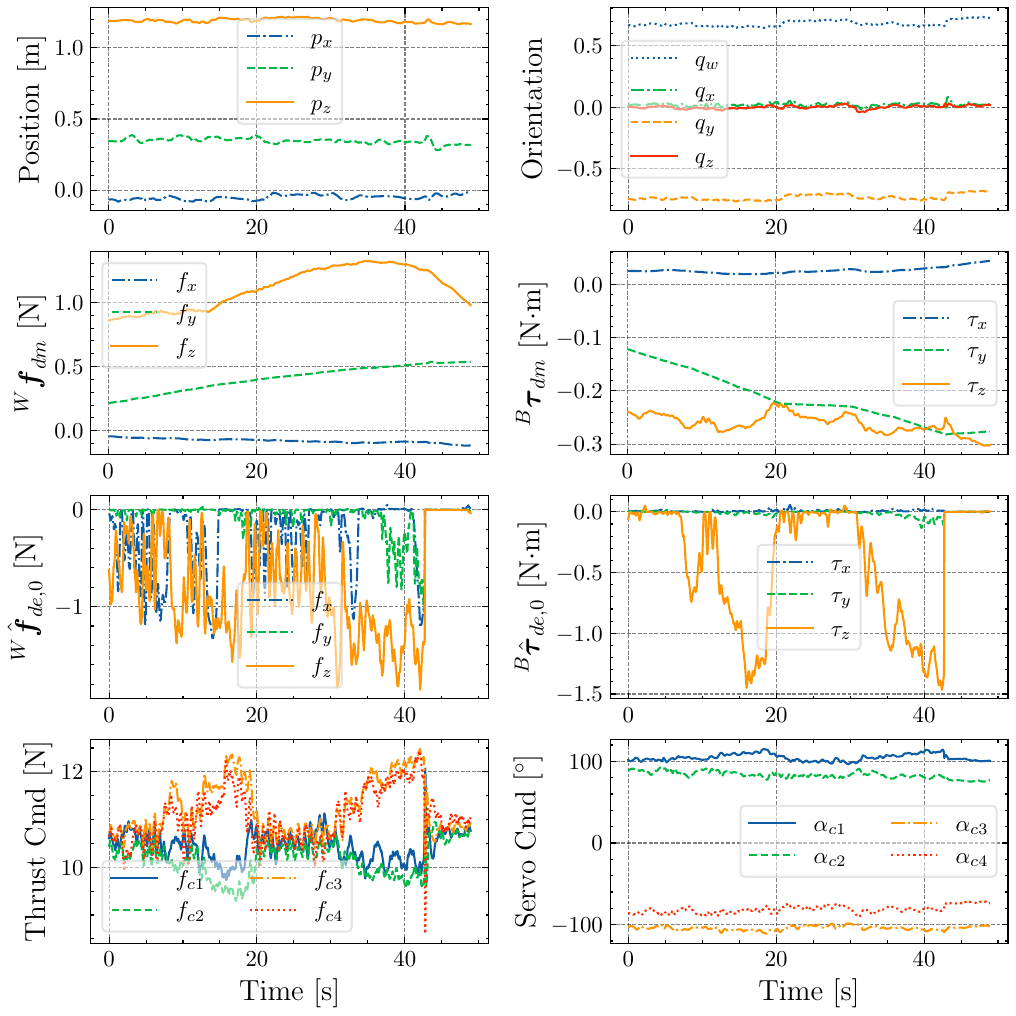}
        \label{fig:exp_vertical_torque_w_comp}
    }\vspace{-1mm}
    \caption{Flight data without and with wrench compensation corresponding to Fig.~\ref{fig:exp_wrench_est_close_loop_photo}b. Without compensation, torque induces a position change, whereas with compensation this effect is eliminated. Some integral terms (e.g., ${^Wf_{dm,y}}$, ${^B\tau_{dm,y}}$, (\ref{eq:model_error})) keep increasing, as model error is not fully compensated.}
    \label{fig:exp_vertical_torque}
\end{figure}

After verifying the performance of the wrench estimator, we integrated it into the control loop for wrench compensation. Direct integration was too sensitive to noise; therefore, as described in Sec.~\ref{sec:sysid_filter}, a \SI{2}{\hertz} low-pass filter was applied to the estimated wrench. We also found that using thrust measurements was more stable than using thrust commands. In all experiments, the robot was flown vertically to demonstrate that wrench estimation did not influence omnidirectional capability. We first tested force and then torque, with external wrench applied by an operator using a brush or a fork (Fig.~\ref{fig:exp_wrench_est_close_loop_photo}). In each test, performance was compared without and with wrench compensation.

The results for lateral force disturbance are shown in Fig.~\ref{fig:exp_vertical_force}. The wrench-compensated controller keeps the robot highly resistant to external force, whereas the baseline tracking controller is easily pushed away.

We separately plot the force generated by the integral term $^W\boldsymbol{f}_{dm}$ and the estimated external wrench $^W\hat{\boldsymbol{f}}_{de,0}$ to highlight their differences. Throughout the process, the integral term continuously corrects model error. However, when an external force is applied, Fig.~\ref{fig:exp_vertical_force_wo_comp} shows that $^W\boldsymbol{f}_{dm}$ accumulates compensation too slowly. Increasing the gain would risk large overshoot during normal tracking. In contrast, with the aid of the wrench estimator, Fig.~\ref{fig:exp_vertical_force_w_comp} shows that the robot responds promptly to the applied wrench. Moreover, unlike $^W{f}_{dm,x}$ in Fig.~\ref{fig:exp_vertical_force_wo_comp}, which exhibits large fluctuations under external force, the corresponding term in Fig.~\ref{fig:exp_vertical_force_w_comp} remains nearly steady, indicating that $^W\boldsymbol{f}_{dm}$ primarily accounts for model error even in the presence of external wrenches. Another noteworthy finding concerns $^B\boldsymbol{\tau}_{de,0}$ in Fig.~\ref{fig:exp_vertical_force_w_comp}: as the applied force increases, torque estimation error also increases, causing a downward rotation that limits further force application.
One possible remedy may be to use the model-error estimate in the controller to correct modeling errors in the wrench estimator.

The results of applying a lateral torque are shown in Fig.~\ref{fig:exp_vertical_torque}. Unlike the force case, the baseline tracking controller alone already counteracts torque about the body Z-axis (Fig.~\ref{fig:exp_vertical_torque_wo_comp}). However, this compensation comes at the cost of a horizontal offset, observed as $p_y$ in the first subplot. In contrast, the robot with wrench compensation avoids this offset (Fig.~\ref{fig:exp_vertical_torque_w_comp}).

Similar to the force case, we plot the wrench generated by the integral term $^B\boldsymbol{\tau}_{dm}$ and the estimated external wrench $^B\boldsymbol{\tau}_{de,0}$ separately to highlight their differences. The same trend is observed: without compensation, $^B\boldsymbol{\tau}_{dm}$ accumulates to counteract the applied torque, whereas with compensation, $^B\boldsymbol{\tau}_{dm}$ primarily corrects model error with the aid of $^B\boldsymbol{\tau}_{de,0}$. This complementarity further verifies the effectiveness of the proposed disturbance-handling framework.






\setlength{\dbltextfloatsep}{8pt plus 1.0pt minus 2.0pt}
\begin{figure}[t] 
    \centering
    \subfloat[Snapshots of pushing a whiteboard in the vertical pose]{
        \includegraphics[trim=0 0 0 0,clip,width=3.45in]{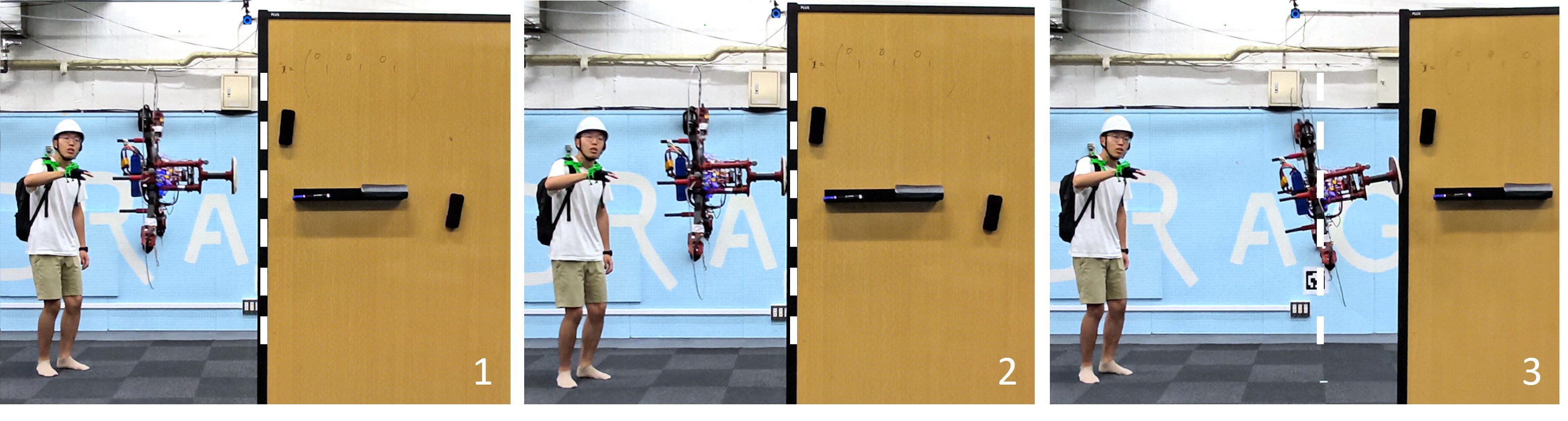}
        \label{fig:exp_door_push_snapshot}
    } \\ \vspace{-2mm}
    \subfloat[Data of pushing a whiteboard in the vertical pose]{
        \includegraphics[trim=0 0 0 0,clip,width=3.45in]{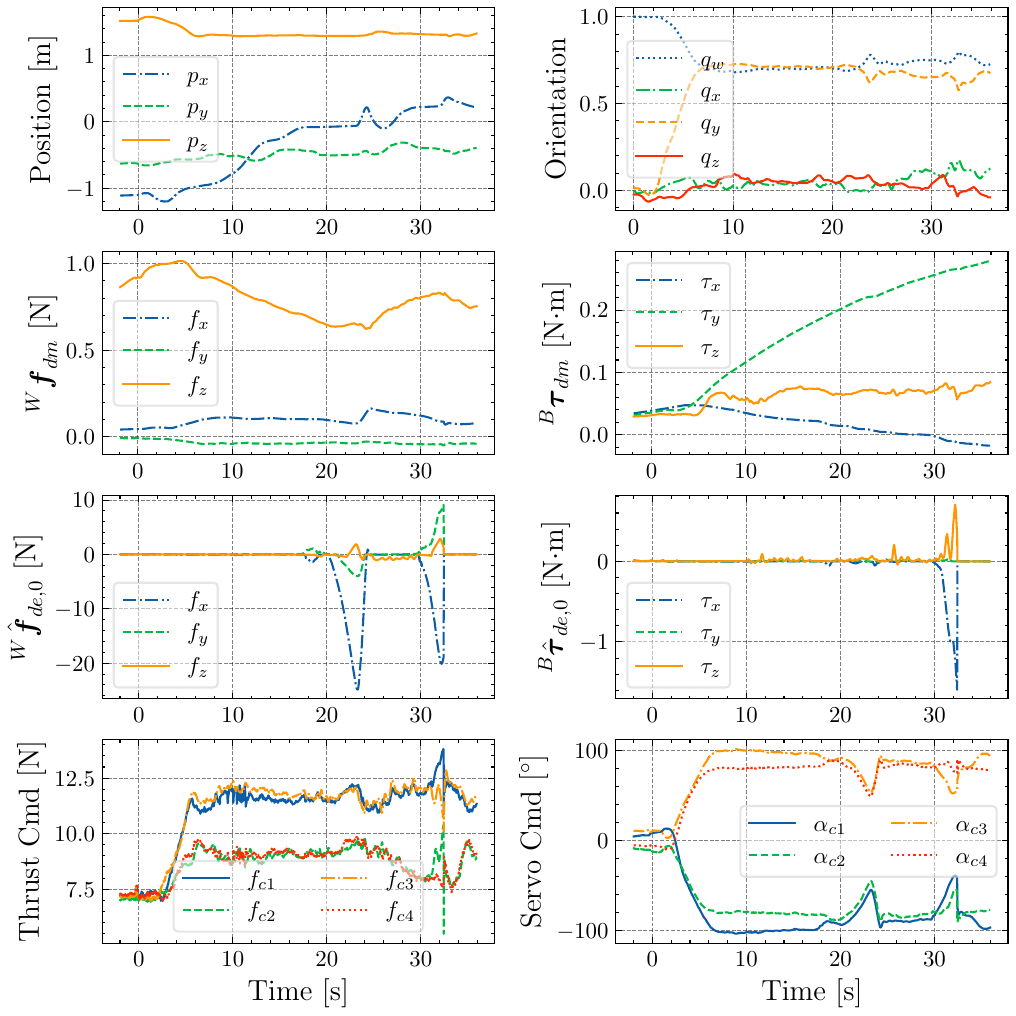}
        \label{fig:exp_door_push_data}
    }\vspace{-1mm}
    \caption{An operator pushes the side of a whiteboard via teleoperation. With wrench compensation activated, the robot’s force accumulates until it exceeds static friction, causing the board to slide. The external-wrench estimate is absent during some intervals (e.g., \SI{24}{\second} to \SI{30}{\second}) because the estimator enters the STOP and CALIB states.}
    \label{fig:exp_vertical_door_push}
\end{figure}

\setlength{\dbltextfloatsep}{8pt plus 1.0pt minus 2.0pt}
\begin{figure}[t] 
    \centering
    \subfloat[Snapshots of turning a vertically installed valve \revised{for \SI{360}{\degree}}]{
        \includegraphics[trim=0 3mm 0 0,clip,width=3.45in]{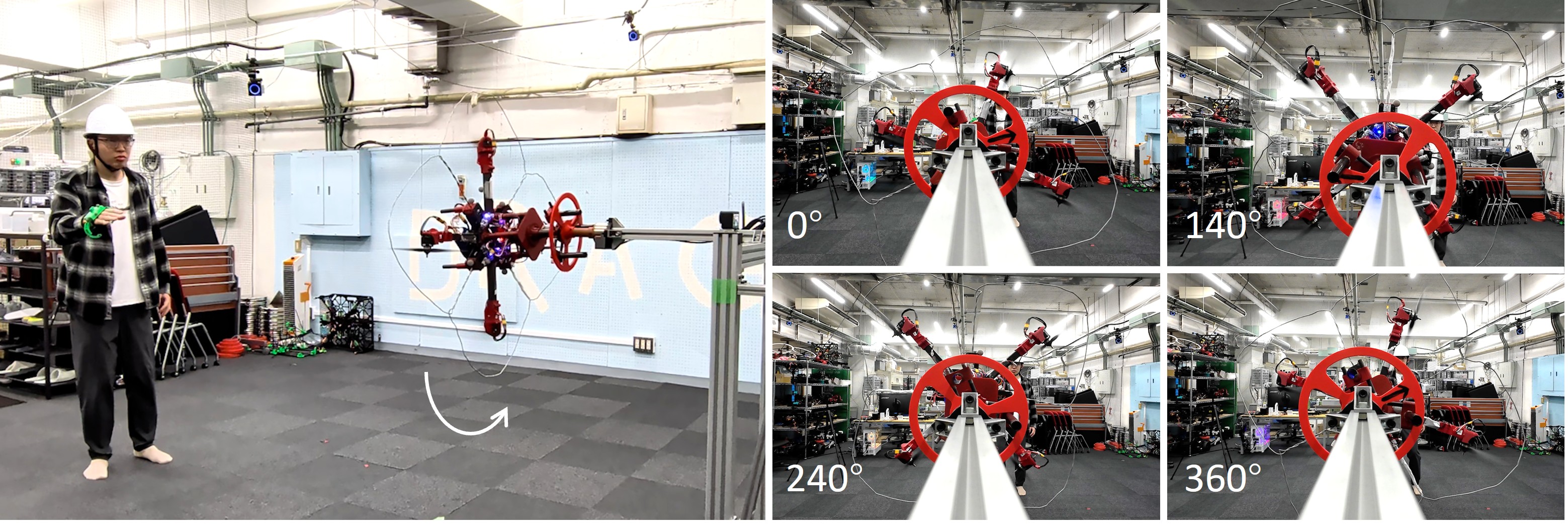}
        \label{fig:exp_hand_valve_turn_snapshot}
    } \\ \vspace{-2mm}
    \subfloat[Data of turning a vertically installed valve \revised{for \SI{360}{\degree}}]{
        \includegraphics[trim=0 0 0 0,clip,width=3.45in]{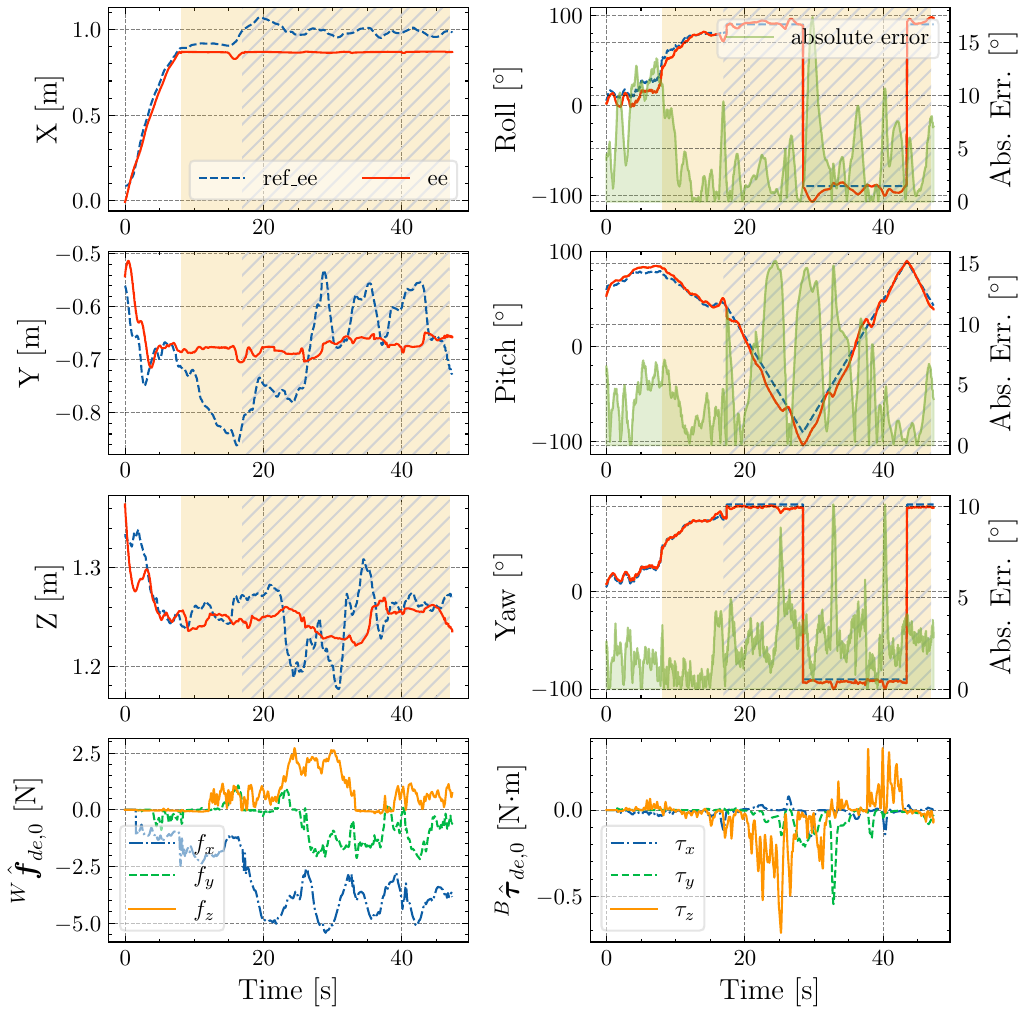}
        \label{fig:exp_hand_valve_turn_data}
    }\vspace{-1mm}
    \caption{Shared-control valve-turning experiment. The yellow and gray regions indicate valve engagement and commanded continuous rotation, respectively. The RMSEs are: $p_x$: \SI{0.108}{\meter}, $p_y$: \SI{0.082}{\meter}, $p_z$: \SI{0.028}{\meter}; roll: \SI{6.120}{\degree}, pitch: \SI{6.930}{\degree}, yaw: \SI{2.925}{\degree}.
    The external wrench is shown only for information and is not used for compensation.
    }
    \label{fig:exp_vertical_hand_valve_turn}
\end{figure}

\subsection{Aerial Manipulation}

\revised{We evaluate the established capabilities in two representative manipulation tasks: whiteboard pushing, which combines pitch rotation and wrench compensation, and continuous turning of a vertically installed valve, which demonstrates omnidirectional flight. References for both tasks are provided through a teleoperation framework~\cite{li_six-dof_2025}.}

\subsubsection{Pushing a Whiteboard}

In this experiment, a brush was selected as the end-effector, and an operator controlled the robot using hand markers. The end-effector's pose was aligned with the operator’s hand with a position offset. 
To align the brush with the whiteboard, the robot pitched by \SI{90}{\degree}, demonstrating its ability to push in different directions.

As shown in Fig.~\ref{fig:exp_door_push_data}, the whiteboard begins to slide after the estimated external force increases to about \SI{20}{\newton}.
This value exceeds the approximately \SI{14}{\newton} force level reached by the controller without compensation in Fig.~\ref{fig:exp_vertical_force_wo_comp}.
A torque of about \SI{1}{\newton\meter} in $^B\boldsymbol{\tau}_{de,0}$ is also observed at around \SI{32}{\second}, possibly because the robot is not perfectly aligned with the \SI{5}{\centi\meter} board frame during the second push. Overall, with the aid of wrench estimation, the robot completes the pushing task.

\subsubsection{\revised{360\si{\degree}} Continuous Valve Turning}

\revised{In this feasibility demonstration,} a fork was selected as the end-effector, and the robot was controlled through the operator’s hand motion. Because continuous valve rotation exceeds the natural flexibility of the human wrist and forearm, we introduced a \textit{shared control} approach: when the operator’s hand rotation exceeded a threshold, the robot executed a continuous vertical rotation of \SI{360}{\degree}. Although the orientation was governed by the robot, the position remained under operator control for safety. Since the proposed framework did not include compliance control, wrench compensation was disabled to avoid excessive internal forces during turning.

\revised{As shown in Fig.~\ref{fig:exp_hand_valve_turn_data}, the robot achieves a continuous \SI{360}{\degree} rotation of the vertically mounted valve under teleoperation.
Due to imperfect manual alignment between the fork and the valve, a contact-induced interaction wrench is observed during rotation.
In the reported trials, the rear view offers better visual alignment and more reliable fork engagement than the side view; the robot completes the \SI{360}{\degree} rotation in all three trials.
These results demonstrate the feasibility of continuous vertical valve turning with a tiltable-quadrotor, while MoCap-free operation, robust autonomous insertion, and compliant contact remain future work.}

\subsection{\revised{Discussion}}

\revised{
The experiments provide several practical lessons.
First, omnidirectional flight requires both algorithmic methods to handle singularities in attitude representation and control allocation \& hardware methods to prevent interference between wires and propellers.
Second, NMPC relaxes the requirement for reference quality, allowing even coarse but directionally correct references to be used. The controller can then generate smooth and dynamically feasible control commands to drive the robot.
Third, large-angle flight makes attitude model-error compensation important because CoG-height uncertainty, assembly imperfections, and tilted-rotor airflow can introduce errors that are negligible in horizontal hovering.
Finally, torque is more difficult to estimate than force, and the two estimates can be coupled in practice. We observed that large external forces can amplify torque-estimation errors and induce downward rotation, suggesting that further calibration is required.
We believe that these lessons, together with our successful demonstrations, can benefit the community and help advance tiltable-quadrotors toward practical aerial manipulation.
}


\section{Conclusion} \label{sec:conclusion}

This work investigated tiltable-multirotors for aerial manipulation from both design and control perspectives. The results suggested that a tiltable-quadrotor with an upward end-effector placement \revised{appears to be a balanced design choice under the considered criteria of} rotor geometry, hovering efficiency, singularity traversal, and wrench generation. To enable wrench-based interaction with environments, model error and external wrench were treated separately through a modified integral term and an acceleration-based estimator. Building on these findings, an effector-centric NMPC framework was developed to unify design insights, singularity handling, and disturbance compensation for omnidirectional manipulation.
The framework was extensively validated in real-world experiments, including whiteboard pushing and continuous turning of a vertical valve under teleoperation.
Future work will extend the NMPC framework to support more compliant behaviors, aiming to enable a wider range of manipulation tasks. \revised{In addition, a more comprehensive task-level benchmark across different rotor numbers and layouts remains an important future direction.}





\bibliographystyle{bibtex/IEEEtran}
\bibliography{bibtex/IEEEabrv, bibtex/my_config, bibtex/references_simplified}


 
\newcommand{\biographysep}{\vspace{-32pt}}

\biographysep

\begin{IEEEbiography}[{\includegraphics[width=1in,height=1.25in,clip,keepaspectratio]{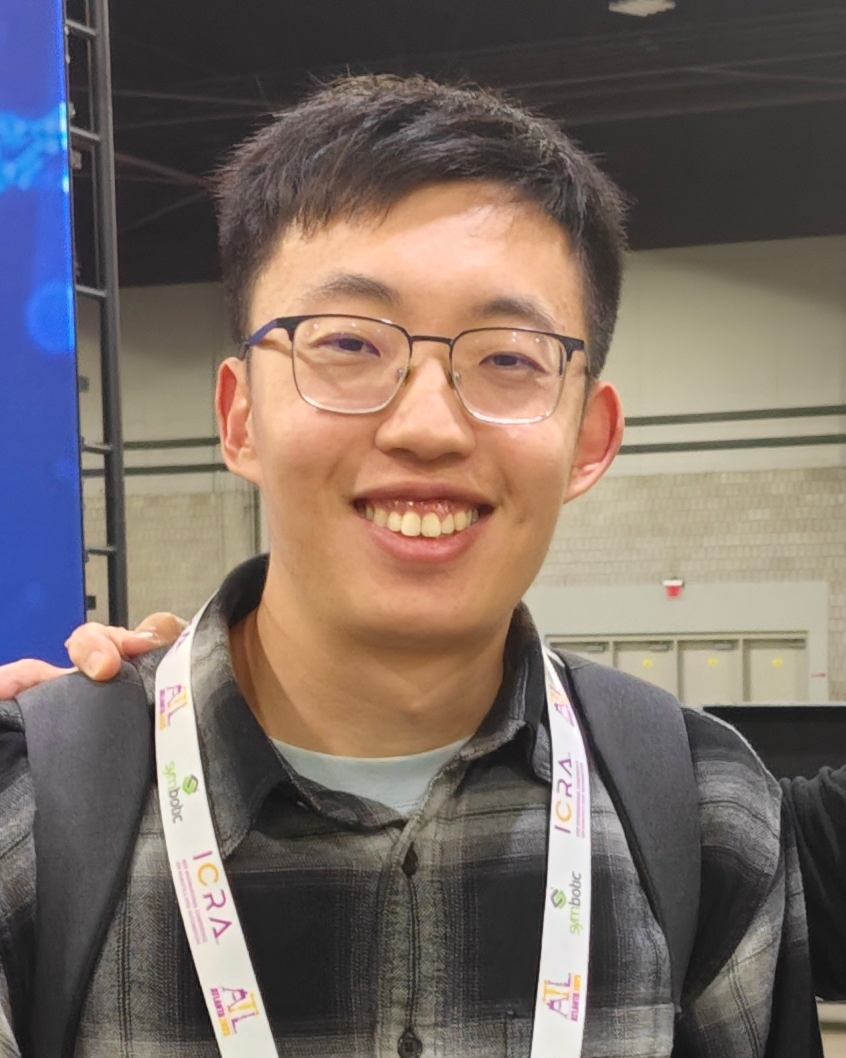}}]{Jinjie Li}
(Graduate Student Member, IEEE) received the B.Eng.\ in automation and the M.Sc.\ in control science and engineering from Beihang University, Beijing, China, in 2020 and 2023, respectively. He is currently pursuing the Ph.D. degree with the Department of Mechanical Engineering, The University of Tokyo, Tokyo, Japan. His research interests include optimization-based control for aerial manipulation, aiming to make aerial robots function as flying hands rather than just eyes.
\end{IEEEbiography}

\biographysep

\begin{IEEEbiography}[{\includegraphics[width=1in,height=1.25in,clip,keepaspectratio]{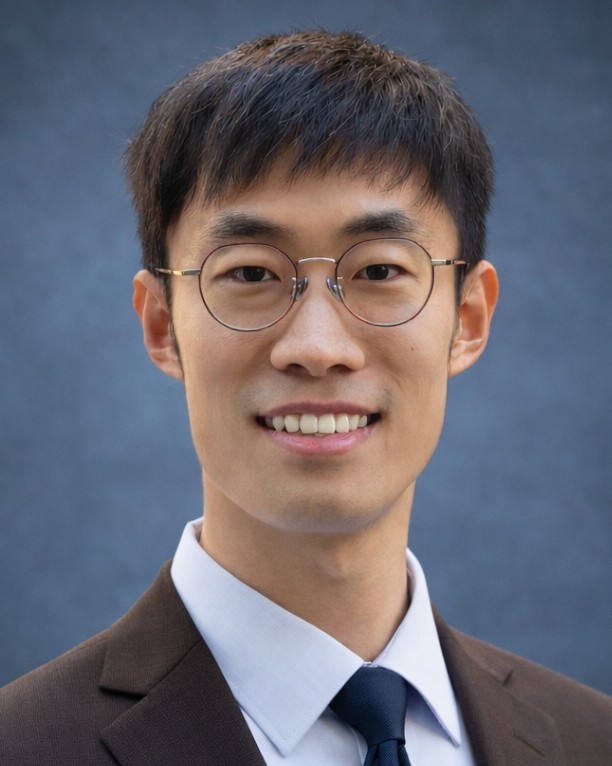}}]{Yicheng Chen}
(Graduate Student Member, IEEE) received the B.E.\ and M.Sc.\ degrees from Beihang University, Beijing, China, in 2021 and 2024, respectively. He is currently working toward the Ph.D. degree in mechanical engineering with the Department of Mechanical Engineering, The University of Tokyo, Tokyo, Japan. His research interests include motion planning in aerial robotics and the intersection of optimization and learning-based methodologies.
\end{IEEEbiography}

\biographysep

\begin{IEEEbiography}[{\includegraphics[width=1in,height=1.25in,clip,keepaspectratio]{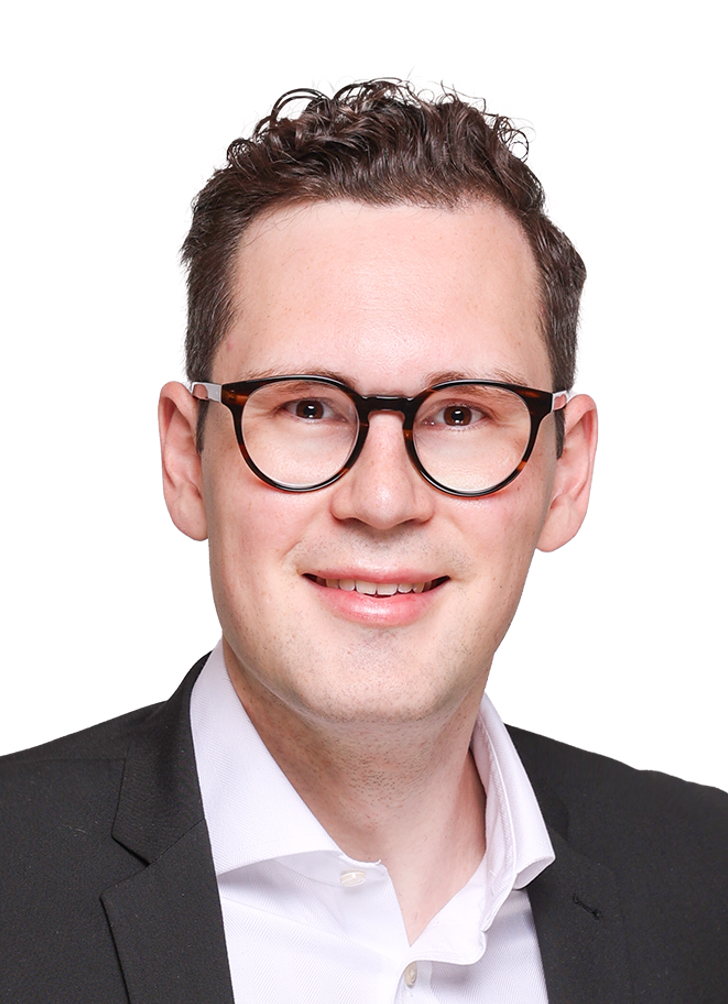}}]{Johannes Kübel}
is a Ph.D. candidate at the DRAGON Lab at the University of Tokyo under the supervision of Moju Zhao. His academic background includes studies in Stuttgart, Germany, as well as Gothenburg, Sweden. His research focuses on the intersection of optimal control and machine learning, specifically the integration of neural networks into model predictive control.
\end{IEEEbiography}

\biographysep

\begin{IEEEbiography}[{\includegraphics[width=1in,height=1.25in,clip,keepaspectratio]{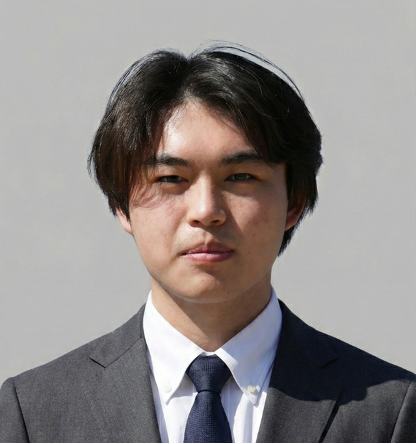}}]{Haokun Liu}
(Graduate Student Member, IEEE) is currently a Ph.D. student with the Department of Mechanical Engineering, The University of Tokyo, Tokyo, Japan. His research interests include vision-language and large language model-based robot control, heterogeneous multi-robot collaboration, and learning-enabled motion planning and state estimation.
\end{IEEEbiography}

\biographysep

\begin{IEEEbiography}[{\includegraphics[width=1in,height=1.25in,clip,keepaspectratio]{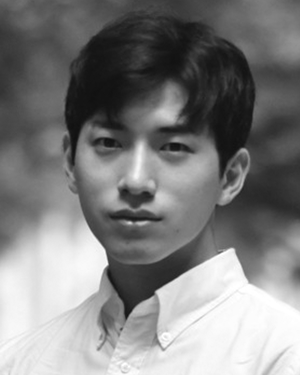}}]{Junichiro Sugihara}
received the B.E. degree in 2023 from the Department of Mechanical Engineering, The University of Tokyo, Tokyo, Japan, where he is currently working toward the Ph.D. degree with the Department of Mechano-Informatics, School of Information Science and Technology. His research interests include mechanical design, modeling and control, motion planning, and state estimation of aerial robots.
\end{IEEEbiography}

\biographysep

\begin{IEEEbiography}[{\includegraphics[width=1in,height=1.25in,clip,keepaspectratio]{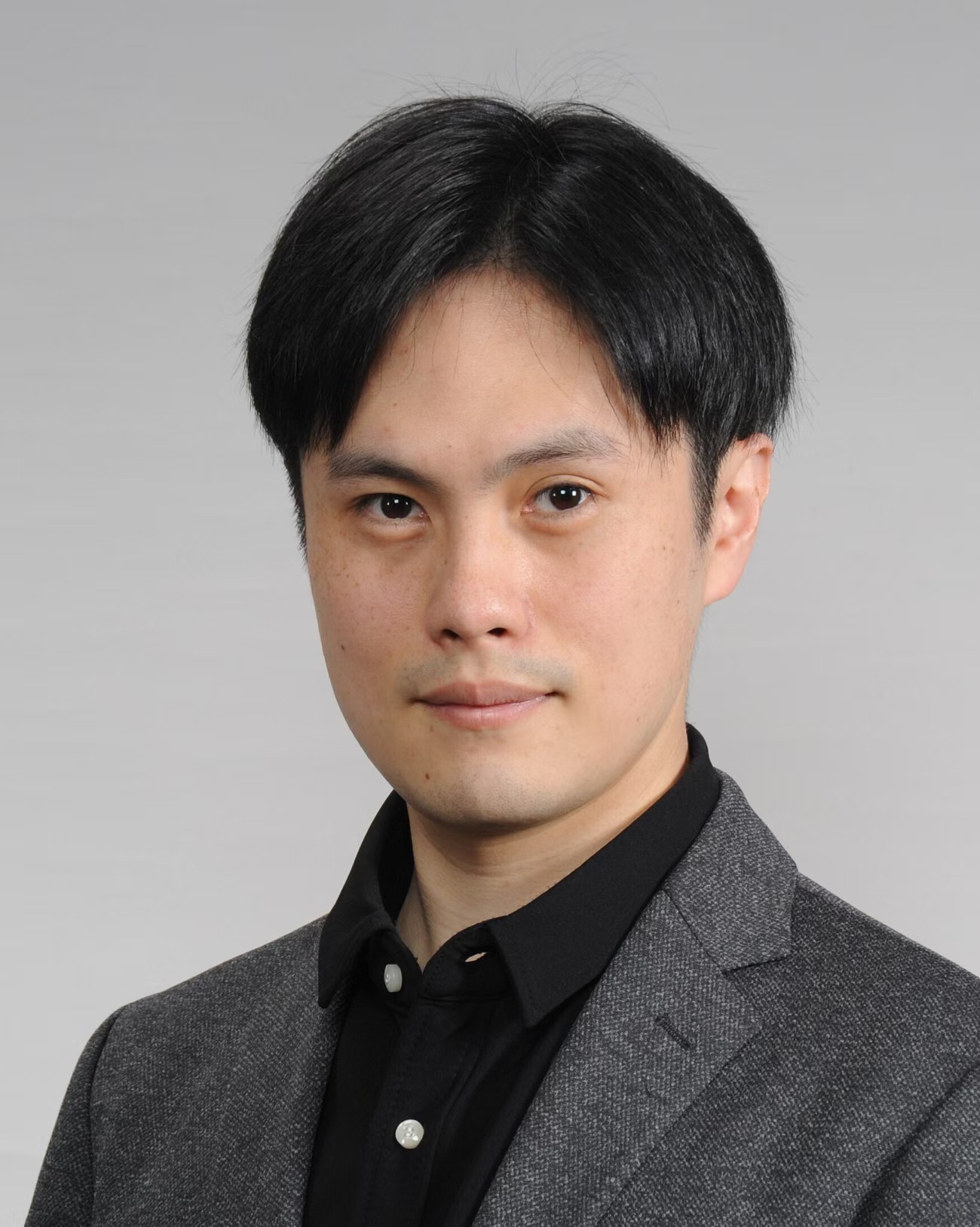}}]{Moju Zhao}
(Member, IEEE) received the Ph.D. degree in information science and technology from the Department of Mechano-Informatics, The University of Tokyo, Tokyo, Japan, in 2018. He is currently a Lecturer (a Junior Associate Professor) with the University of Tokyo. His research interests include mechanical design, modeling and control, motion planning, and vision-based recognition in aerial robotics. Dr.\ Zhao was the recipient of several awards from conferences and journals, including the Best Paper Award at the 2018 IEEE International Conference on Robotics and Automation. His main achievements include articulated aerial robots, such as DRAGON and SPIDAR. He has been an Associate Editor for IEEE Transactions on Robotics since 2026.
\end{IEEEbiography}



\newpage
\appendices

\section{Allocation Matrix} \label{app:allo_mtx}

The matrix $\boldsymbol{A}$ is provided in (\ref{eq:Q_general_nob}) for reproducibility.
\revised{For compactness, let $\rho_i=\lVert\boldsymbol{p}_{E_i,xy}\rVert$ and $\gamma_i=b_i k_q/k_t$.}

\begingroup
\makeatletter
\renewcommand{\fnum@figure}{}
\makeatother
\begin{figure*}[t] 
  \centering
  \begin{equation}
    \boldsymbol{A} \;=\;
    \begin{pmatrix}
      \frac{p_{E_1,y}}{\rho_1}
      &
      0
      &
      \cdots
      &
      \frac{p_{E_{N_p},y}}{\rho_{N_p}}
      &
      0
      \\[3pt]
      -\,\frac{p_{E_1,x}}{\rho_1}
      &
      0
      &
      \cdots
      &
      -\,\frac{p_{E_{N_p},x}}{\rho_{N_p}}
      &
      0
      \\[5pt]
      0
      &
      1
      &
      \cdots
      &
      0
      &
      1
      \\[2pt]
      -\,\frac{\gamma_1\,p_{E_1,y}}{\rho_1}
      \;+\;
      \frac{p_{E_1,x}\,p_{E_1,z}}{\rho_1}
      &
      p_{E_1,y}
      &
      \cdots
      &
      -\,\frac{\gamma_{N_p}\,p_{E_{N_p},y}}{\rho_{N_p}}
      \;+\;
      \frac{p_{E_{N_p},x}\,p_{E_{N_p},z}}{\rho_{N_p}}
      &
      p_{E_{N_p},y}
      \\[3pt]
      \frac{\gamma_1\,p_{E_1,x}}{\rho_1}
      \;+\;
      \frac{p_{E_1,y}\,p_{E_1,z}}{\rho_1}
      &
      -\,p_{E_1,x}
      &
      \cdots
      &
      \frac{\gamma_{N_p}\,p_{E_{N_p},x}}{\rho_{N_p}}
      \;+\;
      \frac{p_{E_{N_p},y}\,p_{E_{N_p},z}}{\rho_{N_p}}
      &
      -\,p_{E_{N_p},x}
      \\[3pt]
      -\,\rho_1
      &
      -\,\gamma_1
      &
      \cdots
      &
      -\,\rho_{N_p}
      &
      -\,\gamma_{N_p}
    \end{pmatrix}
    \label{eq:Q_general_nob}
  \end{equation}
\caption{} \vspace{-3mm}
\end{figure*}
\endgroup

\begin{algorithm}[t]
\caption{Calculate the Available Wrench Envelope in the Body Frame}
\label{alg:analyze_body_wrench}
\begin{algorithmic}[1]
\Require ${\rm mode}\in\{{\rm force},{\rm torque}\}$; angular resolution $r$; search bounds $s_{\rm min}$ and $s_{\rm max}$; search tolerance $s_{\rm tol}$; iteration limit $N_{\rm it,max}$; feasibility tolerance $w_{\rm tol}$; allocation matrix $\boldsymbol{A}$ in (\ref{eq:Q_general_nob}).
\Ensure $\boldsymbol{M}_{\rm result}$, the map of available force or torque magnitudes.
\For{each $\theta_i \in [0^\circ,180^\circ]$ sampled at resolution $r$}
    \For{each $\psi_j \in [-180^\circ,180^\circ]$ sampled at resolution $r$}
        \State ${}^B_N\boldsymbol{R} \gets \boldsymbol{R}_{\rm zyx}(\psi_j,\theta_i,0^\circ)^\top$;
        \State ${}^B\boldsymbol{d}_{ij} \gets {}^B_N\boldsymbol{R}[0,0,1]^\top$;
        \State $l \gets s_{\rm min}$; $h \gets s_{\rm max}$; $s_{\rm best} \gets s_{\rm min}$;
        \For{$n=1$ to $N_{\rm it,max}$}
            \State $s \gets (l+h)/2$;
            \If{${\rm mode} = {\rm ``force"}$}
                \State ${}^B\boldsymbol{w} \gets \left[s{}^B\boldsymbol{d}_{ij};\boldsymbol{0}\right]$;
            \Else
                \State ${}^B\boldsymbol{w} \gets \left[\boldsymbol{0};s{}^B\boldsymbol{d}_{ij}\right]$;
            \EndIf
            \State $\boldsymbol{z}^* \gets $ solve (\ref{eq:alloc_opt});
            \If{$\left\lVert\boldsymbol{A}\boldsymbol{z}^*-{}^B\boldsymbol{w}\right\rVert \leq w_{\rm tol}$}
                \State $s_{\rm best} \gets s$; $l \gets s$;
            \Else
                \State $h \gets s$;
            \EndIf
            \If{$h-l<s_{\rm tol}$}
                \State \textbf{break};
            \EndIf
        \EndFor
        \State $\boldsymbol{M}_{\rm result}(i,j) \gets s_{\rm best}$;
    \EndFor
\EndFor
\end{algorithmic}
\end{algorithm}

\begin{algorithm}[t]
\caption{Calculate the Available Wrench Envelope in the World Frame}
\label{alg:analyze_world_wrench}
\begin{algorithmic}[1]
\Require ${\rm mode}\in\{{\rm force},{\rm torque}\}$; angular resolution $r$; search bounds $s_{\rm min}$ and $s_{\rm max}$; search tolerance $s_{\rm tol}$; iteration limit $N_{\rm it,max}$; feasibility tolerance $w_{\rm tol}$; $\boldsymbol{A}$; gravity-support force ${}^W\boldsymbol{f}_g=[0,0,mg]^\top$; external wrench ${}^W\boldsymbol{f}_{\rm ext},{}^W\boldsymbol{\tau}_{\rm ext}$; mounting angle $\lambda$; fixed yaw angle $\psi_c$.
\Ensure $\boldsymbol{M}_{\rm result}$, the available force or torque magnitude for each sampled pitch angle.
\State ${}^B\boldsymbol{e}_T(\lambda) \gets [\cos\lambda,0,\sin\lambda]^\top$;
\For{each $\theta_i \in [-90^\circ+\lambda,90^\circ+\lambda]$ sampled at resolution $r$}
    \State ${}^W_B\boldsymbol{R} \gets \boldsymbol{R}_{\rm zyx}(\psi_c,\theta_i,0^\circ)$;
    \State ${}^B_W\boldsymbol{R} \gets ({}^W_B\boldsymbol{R})^\top$;
    \State ${}^B\boldsymbol{f}_0 \gets {}^B_W\boldsymbol{R}({}^W\boldsymbol{f}_g+{}^W\boldsymbol{f}_{\rm ext})$;
    \State ${}^B\boldsymbol{\tau}_0 \gets {}^B_W\boldsymbol{R}{}^W\boldsymbol{\tau}_{\rm ext}$;
    \State $l \gets s_{\rm min}$; $h \gets s_{\rm max}$; $s_{\rm best} \gets s_{\rm min}$;
    \For{$n=1$ to $N_{\rm it,max}$}
        \State $s \gets (l+h)/2$;
        \If{${\rm mode} = {\rm ``force"}$}
            \State ${}^B\boldsymbol{w} \gets \left[{}^B\boldsymbol{f}_0+s{}^B\boldsymbol{e}_T(\lambda);{}^B\boldsymbol{\tau}_0\right]$;
        \Else
            \State ${}^B\boldsymbol{w} \gets \left[{}^B\boldsymbol{f}_0;{}^B\boldsymbol{\tau}_0+s{}^B\boldsymbol{e}_T(\lambda)\right]$;
        \EndIf
        \State $\boldsymbol{z}^* \gets $ solve (\ref{eq:alloc_opt});
        \If{$\left\lVert\boldsymbol{A}\boldsymbol{z}^*-{}^B\boldsymbol{w}\right\rVert \leq w_{\rm tol}$}
            \State $s_{\rm best} \gets s$; $l \gets s$;
        \Else
            \State $h \gets s$;
        \EndIf
        \If{$h-l<s_{\rm tol}$}
            \State \textbf{break};
        \EndIf
    \EndFor
    \State $\boldsymbol{M}_{\rm result}(i) \gets s_{\rm best}$;
\EndFor
\end{algorithmic}
\end{algorithm}

\section{Available Wrench Envelope Calculation} \label{app:cal_wrench}

Here we present the algorithms used to calculate the available wrench envelopes in the body and world frames. In both cases, a bisection search is performed along each sampled direction. The initial lower bound $s_{\rm min}$ is assumed feasible, whereas the initial upper bound $s_{\rm max}$ is assumed infeasible.
\begin{equation}
\begin{aligned}
\min_{\boldsymbol{z}\in\mathbb{R}^{2{N_p}}}\quad &
\bigl\lVert\,\boldsymbol{A}\,\boldsymbol{z} - {^B\boldsymbol{w}}\bigr\rVert^{2}, \\
\text{s.t.}\quad
z_{2i-1}^{2} + z_{2i}^{2} & \ \le \ f_{i,\mathrm{max}}^{2},
\quad \forall i = 1,2,\cdots,{N_p}\, ,
\end{aligned}
\label{eq:alloc_opt}
\end{equation}
where $\boldsymbol{z}$ is defined in (\ref{eq:define_z}).

For the world-frame calculation, the implementation evaluates each mounting angle $\lambda$ over the pitch range $[-90^\circ+\lambda,90^\circ+\lambda]$. The yaw angle is fixed for each airframe configuration: $\psi_c=0^\circ$ represents the $\times$ configuration, whereas $\psi_c=45^\circ$ represents the $+$ configuration. The additional force or torque is applied along the end-effector axis ${}^B\boldsymbol{e}_T(\lambda)=[\cos\lambda,0,\sin\lambda]^\top$. The reported experiments set ${}^W\boldsymbol{f}_{\rm ext}={}^W\boldsymbol{\tau}_{\rm ext}=\boldsymbol{0}$ when generating the envelopes.

\section{\revised{Dynamics Derivation of the Prediction-Error Integral Term}} \label{app:integral term}

\begingroup
\setlength{\abovedisplayskip}{4pt}
\setlength{\belowdisplayskip}{4pt}
\setlength{\abovedisplayshortskip}{2pt}
\setlength{\belowdisplayshortskip}{2pt}
\revised{With all variables expressed in the world frame, the corresponding frame notation is omitted for brevity. The integral-term dynamics are given by}
\begin{equation}
    \revised{\dot{\boldsymbol{f}}_{dm} = \boldsymbol{K}_I \Delta \boldsymbol{p} = \boldsymbol{K}_I \left( \hat{\boldsymbol{p}}- \boldsymbol p_{\mathrm{pred}} \right).}
\end{equation}
\revised{At the beginning of each control cycle, $\boldsymbol{p}_0$ is reset to the measured position. Assuming negligible position and velocity measurement errors, the actual position at $t_s$ can be approximated from $\boldsymbol{p}_0$ using a second-order Taylor expansion:}
\begin{equation}
    \revised{\boldsymbol p_{\mathrm{real}}(t_s) \approx \boldsymbol p_0 + \boldsymbol v_0 t_s + \frac{1}{2} \boldsymbol a_{\mathrm{real}} t_s^2 .}
\end{equation}
\revised{Similarly, the one-step prediction is approximated by}
\begin{equation}
    \revised{\boldsymbol p_1 \approx \boldsymbol p_0 + \boldsymbol v_0 t_{\mathrm{step}} + \frac{1}{2} \boldsymbol a_{\mathrm{nom}} t_{\mathrm{step}}^2.}
\end{equation}
\revised{The linearly interpolated predicted position is therefore}
{\color{black}
\begin{align}
    \boldsymbol p_{\mathrm{pred}}(t_s)
    &= \boldsymbol p_0 + \frac{t_s}{t_{\mathrm{step}}} \left( \boldsymbol p_1 - \boldsymbol p_0 \right) \\
    &\approx \boldsymbol p_0 + \boldsymbol v_0 t_s + \frac{1}{2} \boldsymbol a_{\mathrm{nom}} t_s t_{\mathrm{step}} .
\end{align}
}\revised{With position measurement noise neglected, $\hat{\boldsymbol{p}} \approx \boldsymbol p_{\mathrm{real}}$. Thus, the input to the integral term can be approximated as}
{\color{black}
\begin{align}
    \Delta \boldsymbol p
    &\approx \boldsymbol p_{\mathrm{real}} - \boldsymbol p_{\mathrm{pred}}
    \approx \frac{1}{2} \boldsymbol a_{\mathrm{real}} t_s^2 - \frac{1}{2} \boldsymbol a_{\mathrm{nom}} t_s t_{\mathrm{step}}\\
    &= \frac{1}{2} \left( \boldsymbol a_{\mathrm{real}} - \boldsymbol a_{\mathrm{nom}} \right) t_s^2 + \frac{1}{2} \boldsymbol a_{\mathrm{nom}} \left( t_s^2 - t_s t_{\mathrm{step}} \right).
\end{align}
}\revised{For the near-static conditions considered here, $\boldsymbol a_{\mathrm{nom}} \approx \boldsymbol{0}$. Consequently, the integral-term dynamics reduce to}
{\color{black}
\begin{align}
    \dot{\boldsymbol{f}}_{dm}
    &\approx \frac{1}{2}\boldsymbol{K}_I \left( \boldsymbol a_{\mathrm{real}} - \boldsymbol a_{\mathrm{nom}} \right) t_s^2 \\
    &\approx \frac{\boldsymbol{K}_I}{2m} \left( \boldsymbol f_{de} + \boldsymbol f_{\mathrm{other}} - \hat{\boldsymbol f}_{de} - \boldsymbol f_{dm} \right) t_s^2.
\end{align}
}
\endgroup



\end{document}